%% file: main.tex
\pdfoutput=1
\documentclass[10pt, logo, onecolumn, copyright]{nv}
\usepackage{graphicx}
\definecolor{nvidiagreen}{HTML}{76B900}
\definecolor{nvgreen}{HTML}{D3E9AD}

\usepackage{mdframed}
\usepackage{color}
\usepackage{xcolor}
\usepackage[table]{xcolor}
\usepackage[utf8]{inputenc}
\usepackage[T1]{fontenc}

\usepackage{amsfonts}
\usepackage{nicefrac}
\usepackage{microtype}
\usepackage{multirow}
\usepackage{multicol}
\usepackage{tabto}
\usepackage{xspace}
\usepackage{amsmath}
\usepackage{adjustbox}
\usepackage{enumitem}
\usepackage{wrapfig}
\usepackage{dblfloatfix}
\usepackage{times}
\usepackage{inconsolata}
\usepackage{verbatim}
\usepackage{amssymb}
\usepackage{mathtools}
\usepackage{caption}
\usepackage{subcaption}
\usepackage{array}
\usepackage{colortbl}
\usepackage{booktabs}
\usepackage{bbm}
\usepackage{makecell}
\usepackage{float}
\usepackage{placeins}
\usepackage{siunitx}
\usepackage{pifont}
\usepackage{marvosym}
\usepackage{listings}
\usepackage{pdflscape}
\usepackage{footmisc}
\usepackage{url}
\usepackage{tabularx}
\usepackage{arydshln}
\usepackage{hhline}
\usepackage{diagbox}
\usepackage{tcolorbox}
\usepackage[nameinlink]{cleveref}
\usepackage{hyperref}
\usepackage[square,sort,comma,numbers]{natbib}
\usepackage{fp}
\usepackage{authblk}
\usepackage{xspace}

\usepackage{tikz}
\usetikzlibrary{calc}

\crefname{section}{Sec.}{Sec.}
\crefname{equation}{Eq.}{Eqs.}
\crefname{figure}{Figure}{Figures}
\crefname{table}{Table}{Tables}
\Crefname{figure}{Figure}{Figures}
\Crefname{table}{Table}{Tables}
\crefname{appendix}{Appendix}{Appendices}
\newcommand{\ourmethod}{Sol-Attn}

\title{\ourmethod{}: Accelerating Video Generation Inference via On-the-Fly Attention Sparsification}

\author{
\parbox{\linewidth}{
\centering
\vspace{-5pt}
\fontsize{9.6pt}{18pt}\selectfont
Haopeng Li, ~ Yitong Li, ~ Junsong Chen, ~ Tian Ye, ~ Haozhe Liu, ~ Jincheng Yu, ~ Duomin Wang, ~ Ruihua Zhang
\\
\vspace{0.4em}
\textbf{\fontsize{9.6pt}{18pt}\selectfont
Zeke Xie, ~ Enze Xie, ~ Song Han
}
\\
\vspace{2.5mm}
{\normalsize NVIDIA}
\\
\vspace{2pt}
}
}

\begin{abstract}
\vspace{-10pt}
Diffusion transformers are essential for high-fidelity video generation, but long token sequences make attention a dominant inference bottleneck. Training-free dynamic sparse attention alleviates this bottleneck by computing only selected key-value blocks, yet existing methods struggle to sparsify attention both efficiently and accurately for two reasons: (1) Rigid, unpredictable, and costly routing: 
selecting a fixed fraction of top-ranked blocks by proxy score imposes fixed budgets, whereas retaining blocks to reach a target cumulative proxy probability mass yields dynamic but potentially imbalanced budgets;
both incur non-negligible overhead from computing and materializing proxy scores. (2) Lossy keep-or-drop sparsification: unselected blocks are discarded entirely, degrading accuracy under aggressive sparsity. These limitations motivate cheaper dynamic-budget routing while limiting accuracy degradation.
In this paper, we introduce training-free \textbf{\textcolor{nvidiagreen}{\textit{\ourmethod{}}}} (\textit{\underline{S}parsifying \underline{o}n\underline{l}ine attention}) that unifies dynamic routing, sparse computation, and approximation correction in a single online-softmax pass, achieving a better accuracy--efficiency trade-off in sparse attention. The core of \ourmethod{} is \textbf{on-the-fly block thresholding} with \textbf{proxy-score reuse}, which selects critical blocks by comparing block proxy scores against a threshold during online softmax. This design enables dynamic yet controllable block budgets without materializing the proxy map, while directly reusing the proxy scores of unselected blocks to approximate their contribution.
Experiments across image and video generation tasks show that \ourmethod{} advances the quality--efficiency frontier of training-free sparse attention, delivering $2.1\times$ and $2.3\times$ end-to-end speedups for video generation and editing, respectively, while preserving visual quality. Furthermore, for video generation, integrating \ourmethod{} into Sol-Engine with complementary acceleration techniques delivers up to a $5\times$ end-to-end speedup.
    \newline
    \textbf{Links:}
    {\hypersetup{urlcolor=nvidiagreen}
    \href{https://github.com/NVlabs/Sana/tree/sol-engine}{GitHub Code} |
    \href{https://nvlabs.github.io/Sana/Sol-Attn/}{Project Page}
    }
\end{abstract}

\begin{document}

\maketitle

\vspace{1pt}

\begin{figure}[h]
    \centering
    \input{figures_tex/titlepage}
    \vspace{-12pt}
    \caption{Overview of generated samples comparing dense attention and Sol-Attn. Across diverse generation tasks, Sol-Attn substantially accelerates generation without noticeable quality degradation.}
    \label{fig:title}
\end{figure}
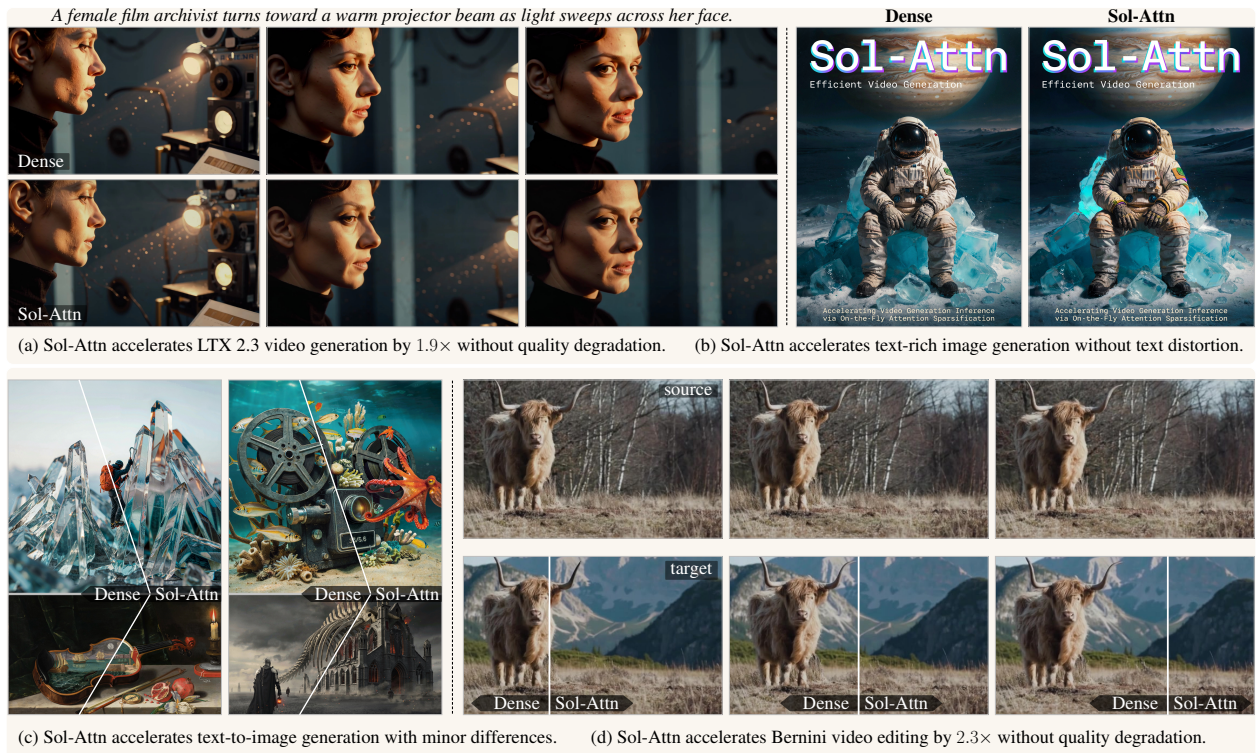

\input{sections/1_intro}
\input{sections/2_preliminary}
\input{sections/3_method}
\input{sections/4_exps}
\input{sections/5_related}

\newpage
\appendix
\input{sections/x_appendix}

\clearpage
{
  \small
  \bibliographystyle{unsrtnat}
  \bibliography{ref}
}

\end{document}

%% file: figures_tex/titlepage.tex
\usetikzlibrary{fadings}
\tikzfading[
  name=sol label fade,
  left color=transparent!100,
  right color=transparent!0
]

\definecolor{panel_color}{RGB}{245,238,225}
\definecolor{highlight_color}{RGB}{245,194,66}
\definecolor{method_a_color}{HTML}{ED220D}
\definecolor{method_b_color}{HTML}{76B900}

\newcommand{\placeholderimg}[3][140pt]{%
  \begin{tikzpicture}
    \node[
      draw=black!35,
      fill=black!5,
      minimum width=#1,
      minimum height=#2,
      inner sep=0pt,
      align=center,
      font=\large\bfseries,
      text=black!55
    ] {#3};
  \end{tikzpicture}%
}

\newcommand{\titleimg}[3]{%
  \begin{tikzpicture}
    \node[
      draw=black!25,
      line width=0.35pt,
      minimum width=#1,
      minimum height=#2,
      inner sep=0pt
    ] {\includegraphics[width=#1,height=#2]{#3}};
  \end{tikzpicture}%
}

\newcommand{\titlefitimg}[3]{%
  \begin{tikzpicture}
    \coordinate (posterSW) at (0pt,0pt);
    \coordinate (posterNE) at (#1,#2);
    \path[use as bounding box] (posterSW) rectangle (posterNE);
    \begin{scope}
      \clip (posterSW) rectangle (posterNE);
      \node[anchor=center, inner sep=0pt] at ($(posterSW)!0.5!(posterNE)$)
        {\includegraphics[height=#2]{#3}};
    \end{scope}
    \draw[black!25, line width=0.35pt] (posterSW) rectangle (posterNE);
  \end{tikzpicture}%
}

\newcommand{\titlecropimg}[3]{%
  \begin{tikzpicture}
    \coordinate (cropSW) at (0pt,0pt);
    \coordinate (cropNE) at (#1,#2);
    \path[use as bounding box] (cropSW) rectangle (cropNE);
    \begin{scope}
      \clip (cropSW) rectangle (cropNE);
      \node[anchor=center, inner sep=0pt] at ($(cropSW)!0.5!(cropNE)$)
        {\includegraphics[height=#2]{#3}};
    \end{scope}
    \draw[black!25, line width=0.35pt] (cropSW) rectangle (cropNE);
  \end{tikzpicture}%
}

\newcommand{\titlewidecropimg}[3]{%
  \begin{tikzpicture}
    \coordinate (wideSW) at (0pt,0pt);
    \coordinate (wideNE) at (#1,#2);
    \path[use as bounding box] (wideSW) rectangle (wideNE);
    \begin{scope}
      \clip (wideSW) rectangle (wideNE);
      \node[anchor=center, inner sep=0pt] at ($(wideSW)!0.5!(wideNE)$)
        {\includegraphics[width=#1]{#3}};
    \end{scope}
    \draw[black!25, line width=0.35pt] (wideSW) rectangle (wideNE);
  \end{tikzpicture}%
}

\newcommand{\titlesplitimg}[5]{%
  \begin{tikzpicture}
    \coordinate (splitSW) at (0pt,0pt);
    \coordinate (splitNE) at (#1,#2);
    \coordinate (splitS) at (#3,0pt);
    \coordinate (splitN) at (#3,#2);
    \path[use as bounding box] (splitSW) rectangle (splitNE);
    \begin{scope}
      \clip (splitSW) rectangle (splitN);
      \node[anchor=center, inner sep=0pt] at ($(splitSW)!0.5!(splitNE)$)
        {\includegraphics[height=#2]{#4}};
    \end{scope}
    \begin{scope}
      \clip (splitS) rectangle (splitNE);
      \node[anchor=center, inner sep=0pt] at ($(splitSW)!0.5!(splitNE)$)
        {\includegraphics[height=#2]{#5}};
    \end{scope}
    \draw[white, line width=2pt] (splitS) -- (splitN);
    \draw[black!25, line width=0.35pt] (splitSW) rectangle (splitNE);
  \end{tikzpicture}%
}

\newcommand{\titlesplitimgnolines}[5]{%
  \begin{tikzpicture}
    \coordinate (splitSW) at (0pt,0pt);
    \coordinate (splitNE) at (#1,#2);
    \coordinate (splitS) at (#3,0pt);
    \coordinate (splitN) at (#3,#2);
    \path[use as bounding box] (splitSW) rectangle (splitNE);
    \begin{scope}
      \clip (splitSW) rectangle (splitN);
      \node[anchor=center, inner sep=0pt] at ($(splitSW)!0.5!(splitNE)$)
        {\includegraphics[height=#2]{#4}};
    \end{scope}
    \begin{scope}
      \clip (splitS) rectangle (splitNE);
      \node[anchor=center, inner sep=0pt] at ($(splitSW)!0.5!(splitNE)$)
        {\includegraphics[height=#2]{#5}};
    \end{scope}
    \draw[black!25, line width=0.35pt] (splitSW) rectangle (splitNE);
  \end{tikzpicture}%
}

\newcommand{\titlediagsplitimg}[7]{%
  \begin{tikzpicture}
    \coordinate (diagSW) at (0pt,0pt);
    \coordinate (diagSE) at (#1,0pt);
    \coordinate (diagNW) at (0pt,#2);
    \coordinate (diagNE) at (#1,#2);
    \coordinate (diagTop) at (#3,#2);
    \coordinate (diagBottom) at (#4,0pt);
    \path[use as bounding box] (diagSW) rectangle (diagNE);
    \begin{scope}
      \clip (diagSW) -- (diagBottom) -- (diagTop) -- (diagNW) -- cycle;
      \node[anchor=center, inner sep=0pt] at ($(diagSW)!0.5!(diagNE)$)
        {\includegraphics[#7]{#5}};
    \end{scope}
    \begin{scope}
      \clip (diagBottom) -- (diagSE) -- (diagNE) -- (diagTop) -- cycle;
      \node[anchor=center, inner sep=0pt] at ($(diagSW)!0.5!(diagNE)$)
        {\includegraphics[#7]{#6}};
    \end{scope}
    \draw[white, line width=1.5pt] (diagTop) -- (diagBottom);
    \draw[black!25, line width=0.35pt] (diagSW) rectangle (diagNE);
  \end{tikzpicture}%
}

\newcommand{\titlediagsplitimgnolines}[7]{%
  \begin{tikzpicture}
    \coordinate (diagSW) at (0pt,0pt);
    \coordinate (diagSE) at (#1,0pt);
    \coordinate (diagNW) at (0pt,#2);
    \coordinate (diagNE) at (#1,#2);
    \coordinate (diagTop) at (#3,#2);
    \coordinate (diagBottom) at (#4,0pt);
    \path[use as bounding box] (diagSW) rectangle (diagNE);
    \begin{scope}
      \clip (diagSW) -- (diagBottom) -- (diagTop) -- (diagNW) -- cycle;
      \node[anchor=center, inner sep=0pt] at ($(diagSW)!0.5!(diagNE)$)
        {\includegraphics[#7]{#5}};
    \end{scope}
    \begin{scope}
      \clip (diagBottom) -- (diagSE) -- (diagNE) -- (diagTop) -- cycle;
      \node[anchor=center, inner sep=0pt] at ($(diagSW)!0.5!(diagNE)$)
        {\includegraphics[#7]{#6}};
    \end{scope}
    \draw[black!25, line width=0.35pt] (diagSW) rectangle (diagNE);
  \end{tikzpicture}%
}

\newcommand{\titlehdiagimageyshift}{0pt}
\newcommand{\titlehdiagsplitimg}[7]{%
  \begin{tikzpicture}
    \coordinate (diagSW) at (0pt,0pt);
    \coordinate (diagSE) at (#1,0pt);
    \coordinate (diagNW) at (0pt,#2);
    \coordinate (diagNE) at (#1,#2);
    \coordinate (diagLeft) at (0pt,#3);
    \coordinate (diagRight) at (#1,#4);
    \path[use as bounding box] (diagSW) rectangle (diagNE);
    \begin{scope}
      \clip (diagSW) -- (diagSE) -- (diagRight) -- (diagLeft) -- cycle;
      \node[anchor=center, inner sep=0pt] at ([yshift=\titlehdiagimageyshift]$(diagSW)!0.5!(diagNE)$)
        {\includegraphics[#7]{#5}};
    \end{scope}
    \begin{scope}
      \clip (diagLeft) -- (diagRight) -- (diagNE) -- (diagNW) -- cycle;
      \node[anchor=center, inner sep=0pt] at ([yshift=\titlehdiagimageyshift]$(diagSW)!0.5!(diagNE)$)
        {\includegraphics[#7]{#6}};
    \end{scope}
    \draw[white, line width=1.5pt] (diagLeft) -- (diagRight);
    \draw[black!25, line width=0.35pt] (diagSW) rectangle (diagNE);
  \end{tikzpicture}%
}

\tikzset{
  modulelink/.style={
    very thick,
    densely dashed,
    shorten >=1pt,
    shorten <=1pt,
    rounded corners=3pt
  },
  panelbox/.style={
    draw=black!0,
    very thick,
    fill=panel_color!50,
    minimum width=118pt,
    rounded corners=8pt,
  },
  titlelabel/.style={
    fill=black,
    fill opacity=0.48,
    text opacity=1,
    text=white,
    font=\huge,
    inner xsep=6pt,
    inner ysep=2.5pt,
    rounded corners=1pt,
  },
}

\resizebox{\linewidth}{!}{
\begin{tikzpicture}
\centering

\node[panelbox, minimum height=400pt, minimum width=1400pt] (panel_top) at (0,0) {};

\node[
  panelbox,
  minimum height=443.75pt,
  minimum width=1400pt,
  anchor=north
] (panel_bottom) at ([xshift=0pt, yshift=-2pt]panel_top.south) {};

\node[font=\huge, anchor=center] (top_prompt)
  at ([xshift=-267pt, yshift=-11pt]panel_top.north)
  {\textit{A female film archivist turns toward a warm projector beam as light sweeps across her face.}};

\node[anchor=west, inner sep=0pt] (seq_a_1)
  at ([xshift=-698pt, yshift=95.421875pt]panel_top.center)
  {\titlecropimg{282pt}{164.84375pt}{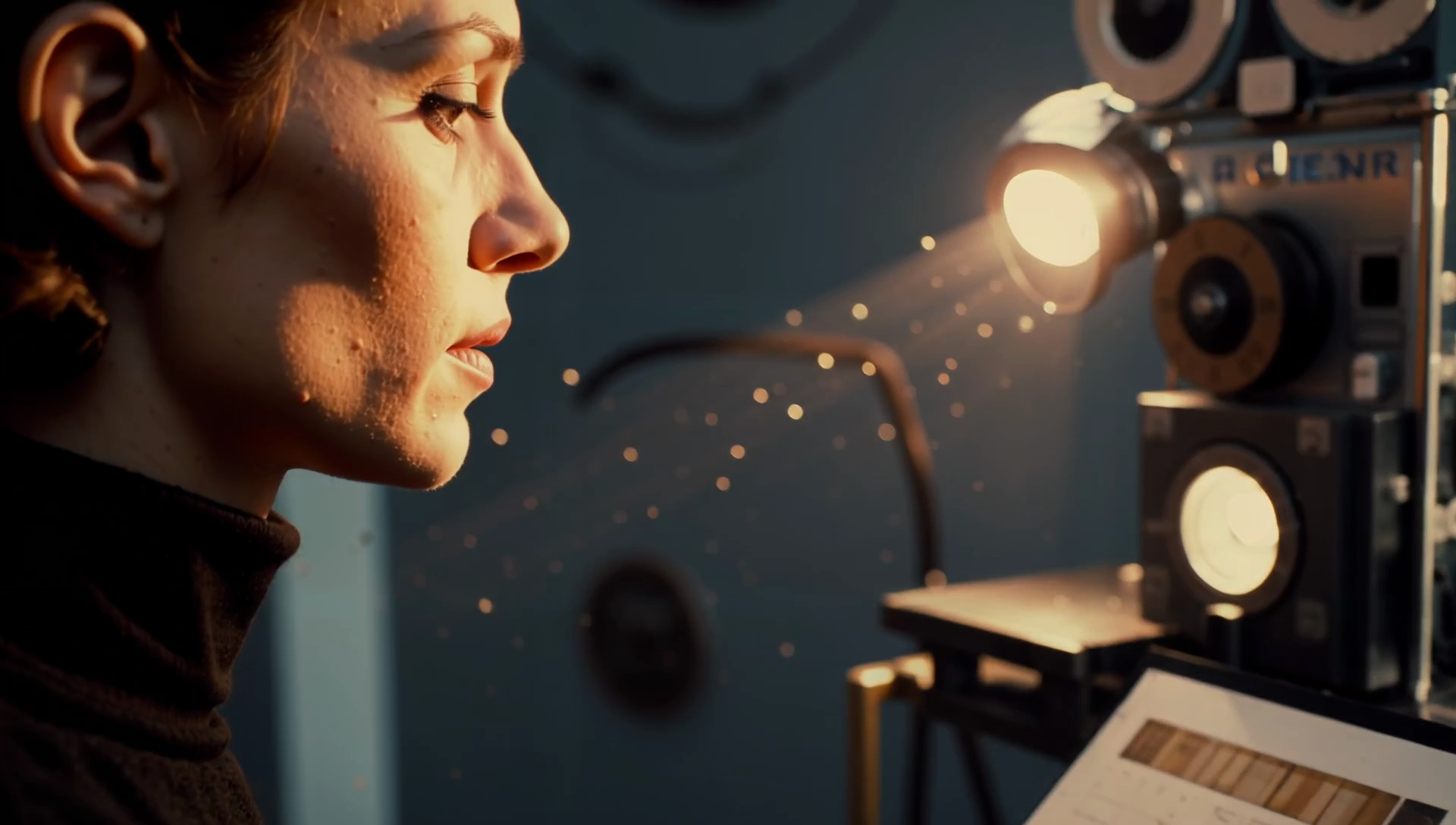}};

\node[anchor=west, inner sep=0pt] (seq_a_2)
  at ([xshift=-408pt, yshift=95.421875pt]panel_top.center)
  {\titlecropimg{282pt}{164.84375pt}{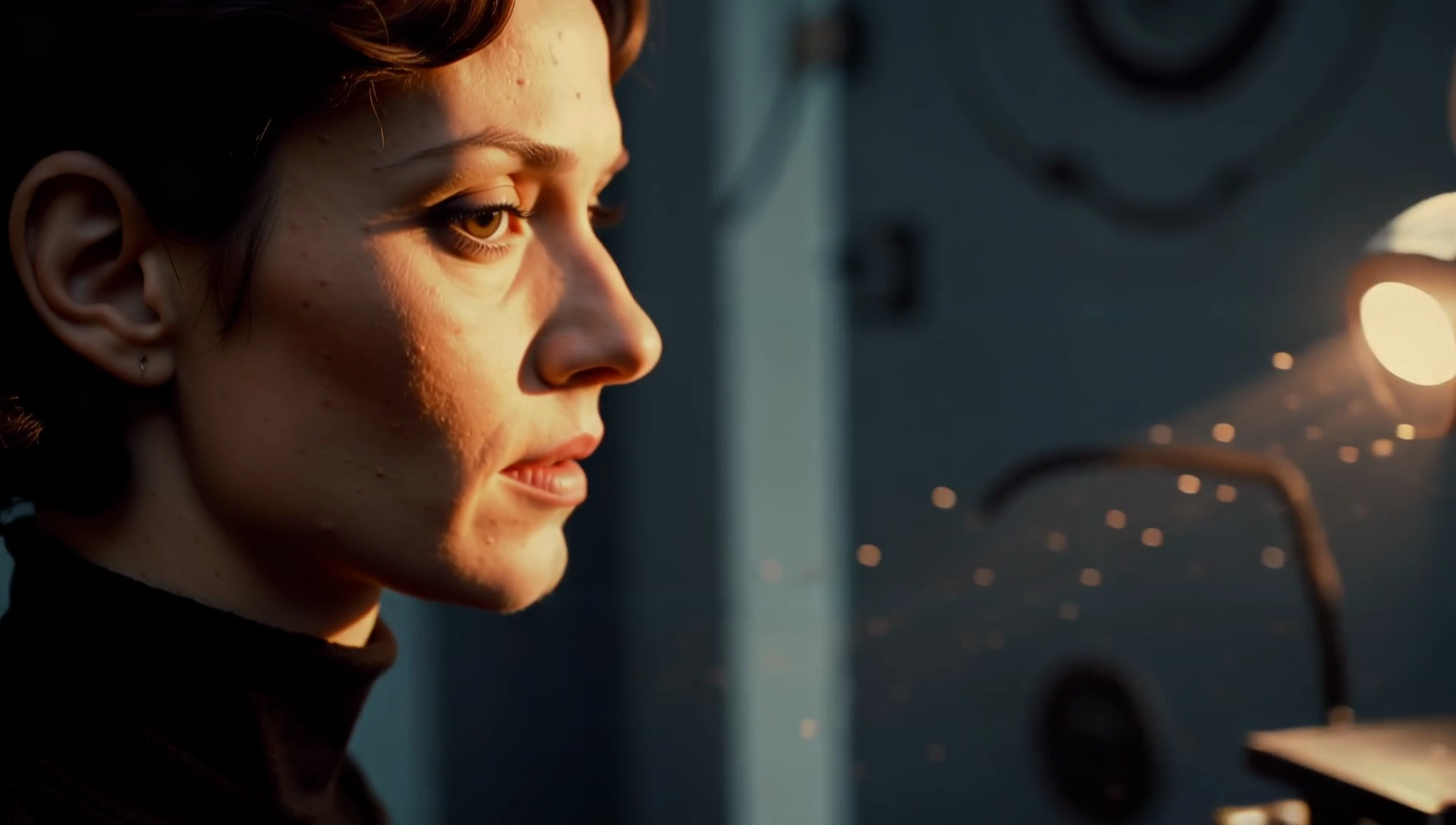}};

\node[anchor=west, inner sep=0pt] (seq_a_3)
  at ([xshift=-118pt, yshift=95.421875pt]panel_top.center)
  {\titlecropimg{282pt}{164.84375pt}{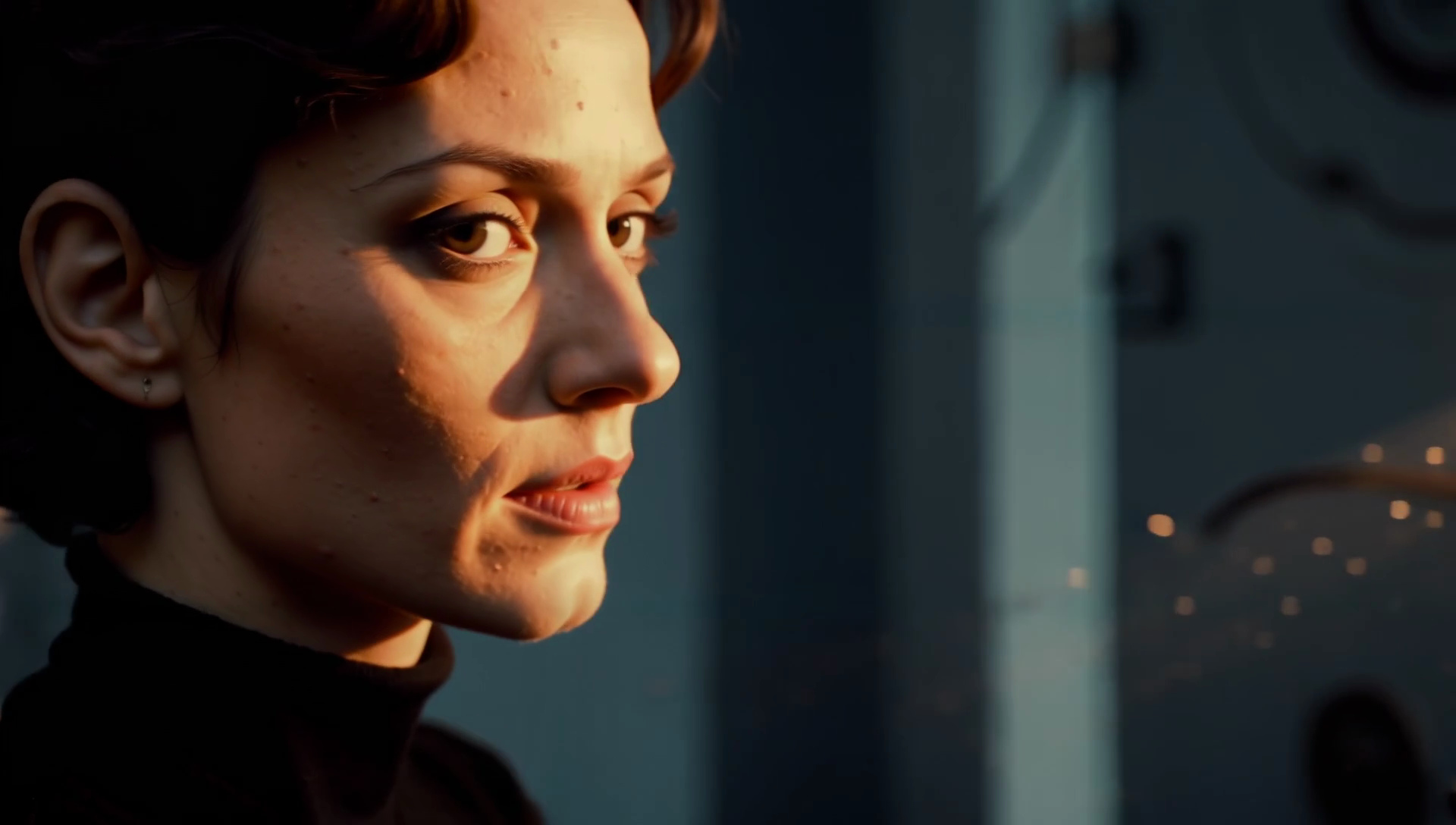}};

\node[anchor=west, inner sep=0pt] (seq_b_1)
  at ([xshift=-698pt, yshift=-75.421875pt]panel_top.center)
  {\titlecropimg{282pt}{164.84375pt}{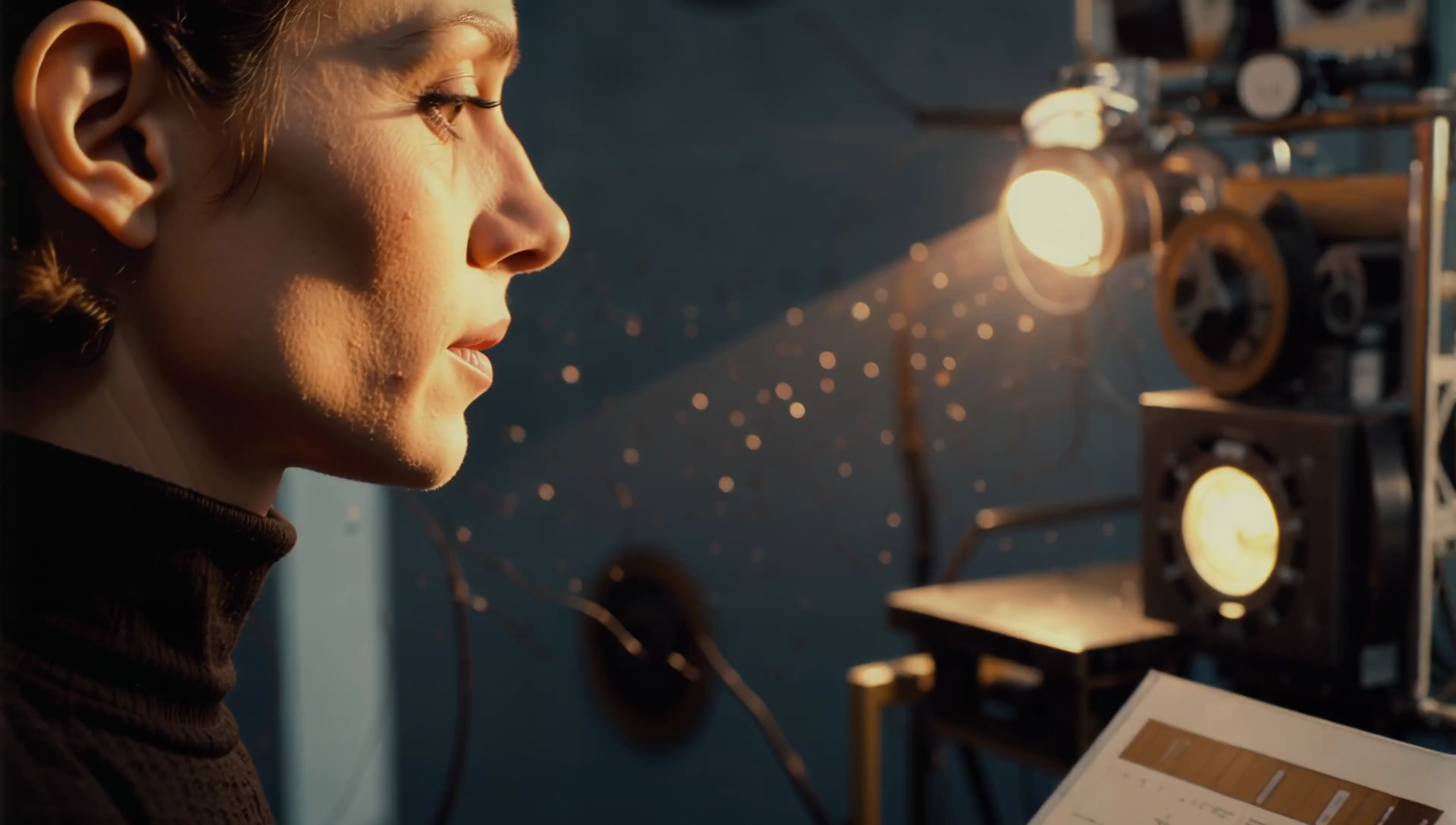}};

\node[anchor=west, inner sep=0pt] (seq_b_2)
  at ([xshift=-408pt, yshift=-75.421875pt]panel_top.center)
  {\titlecropimg{282pt}{164.84375pt}{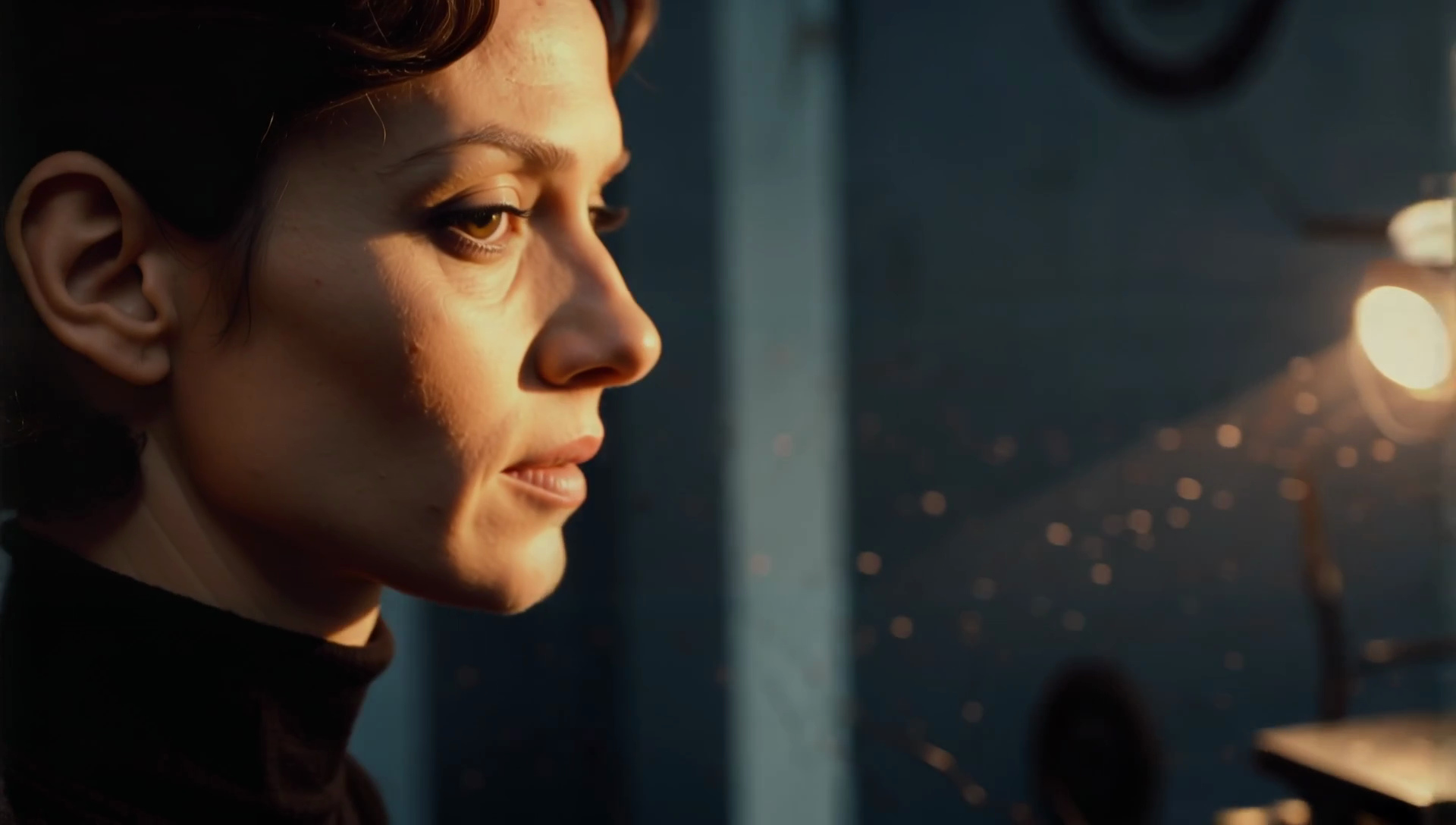}};

\node[anchor=west, inner sep=0pt] (seq_b_3)
  at ([xshift=-118pt, yshift=-75.421875pt]panel_top.center)
  {\titlecropimg{282pt}{164.84375pt}{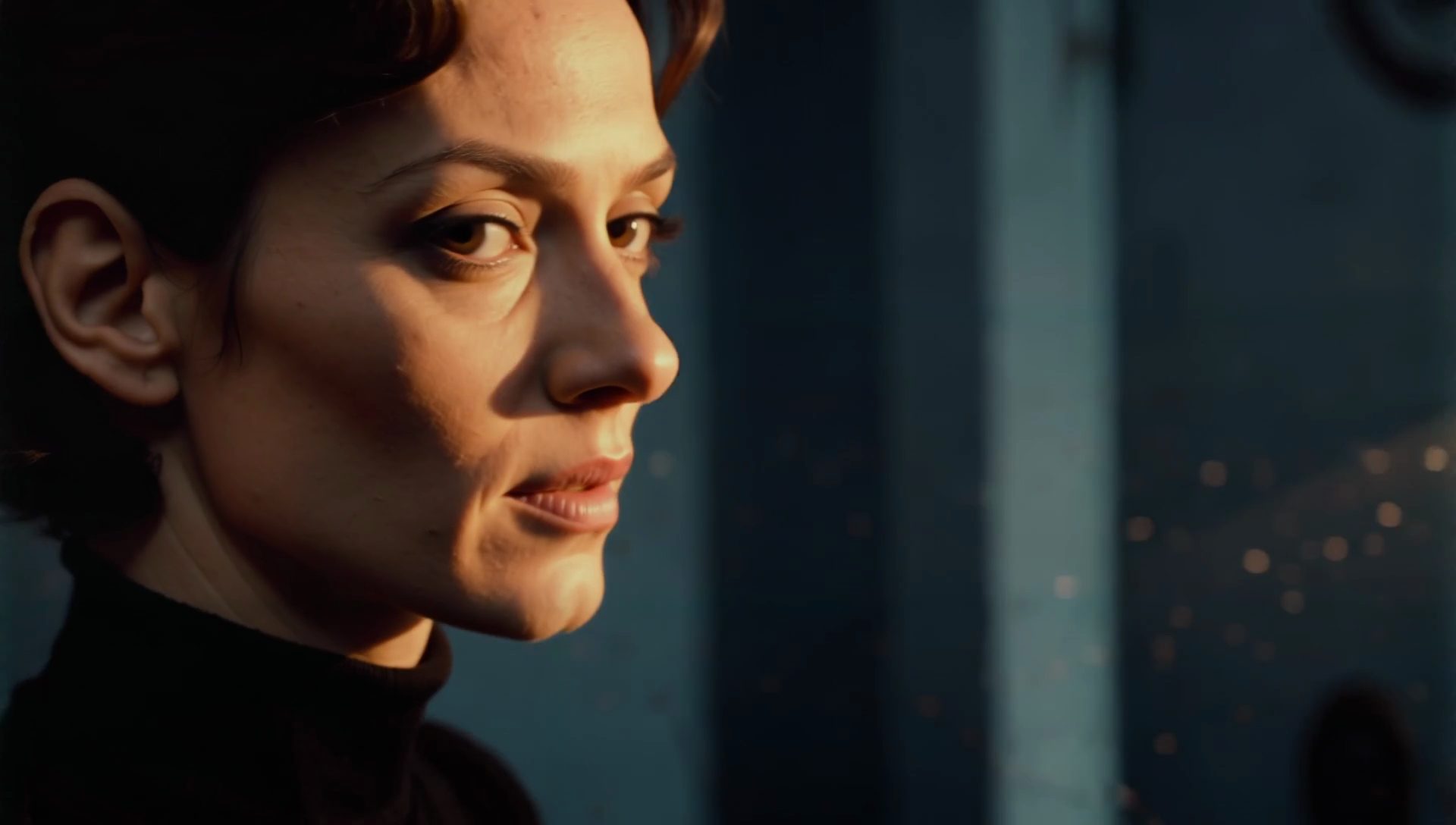}};

\node[titlelabel, anchor=south west]
  at ([xshift=5pt, yshift=5pt]seq_a_1.south west) {Dense};

\node[titlelabel, anchor=south west]
  at ([xshift=5pt, yshift=5pt]seq_b_1.south west) {Sol-Attn};

\node[anchor=west, inner sep=0pt] (poster_1)
  at ([xshift=185pt, yshift=10pt]panel_top.center)
  {\titlewidecropimg{252.5pt}{335.6875pt}{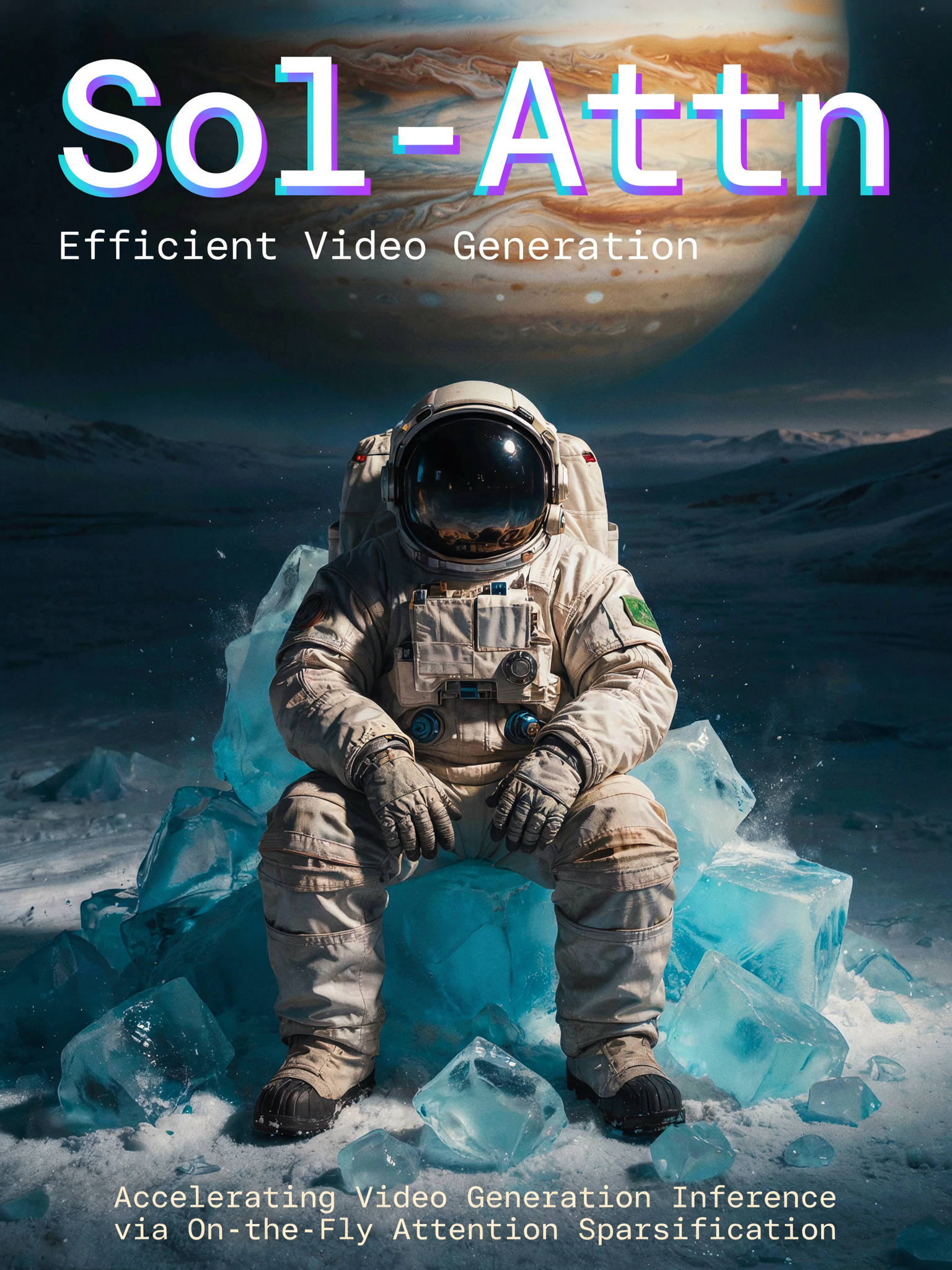}};

\node[anchor=west, inner sep=0pt] (poster_2)
  at ([xshift=445.5pt, yshift=10pt]panel_top.center)
  {\titlewidecropimg{252.5pt}{335.6875pt}{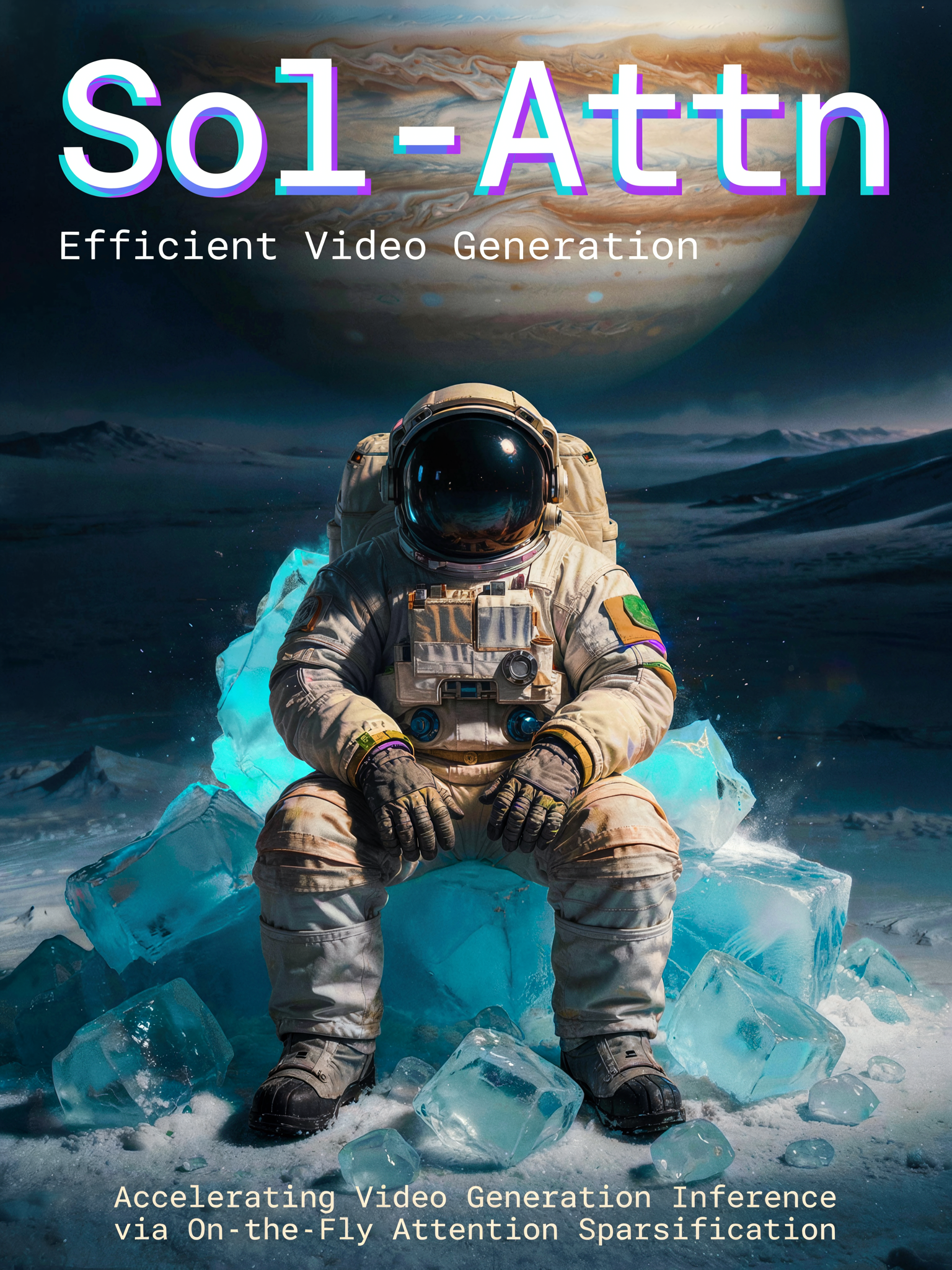}};

\node[font=\huge\bfseries, anchor=south]
  at ([yshift=1.5pt]poster_1.north) {Dense};

\node[font=\huge\bfseries, anchor=south]
  at ([yshift=1.5pt]poster_2.north) {Sol-Attn};

\node[font=\huge, anchor=west] (seq_caption)
  at ([xshift=-690pt, yshift=20pt]panel_top.south)
  {(a) Sol-Attn accelerates LTX 2.3 video generation by $1.9\times$ without quality degradation.};

\node[font=\huge, anchor=east] (poster_caption)
  at ([xshift=690pt, yshift=20pt]panel_top.south)
  {(b) Sol-Attn accelerates text-rich image generation without text distortion.};

\draw[modulelink, draw=black]
  ([xshift=-10.5pt]poster_1.north west)
  -- ([xshift=-10.5pt]poster_1.south west);

\node[anchor=west, inner sep=0pt] (copy_a_1)
  at ([xshift=-188pt, yshift=119.078125pt]panel_bottom.center)
  {\titlecropimg{290pt}{177.28125pt}{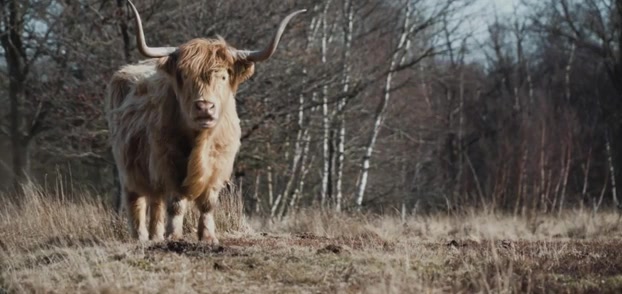}};

\node[anchor=west, inner sep=0pt] (copy_a_2)
  at ([xshift=110pt, yshift=119.078125pt]panel_bottom.center)
  {\titlecropimg{290pt}{177.28125pt}{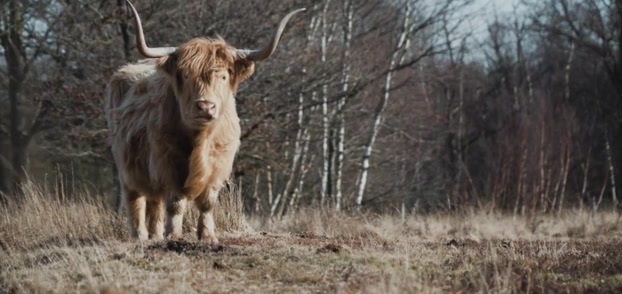}};

\node[anchor=west, inner sep=0pt] (copy_a_3)
  at ([xshift=408pt, yshift=119.078125pt]panel_bottom.center)
  {\titlecropimg{290pt}{177.28125pt}{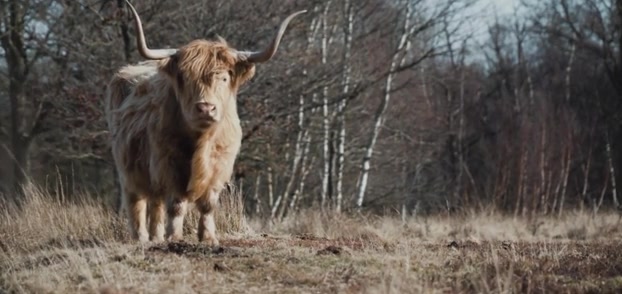}};

\node[anchor=west, inner sep=0pt] (copy_b_1)
  at ([xshift=-188pt, yshift=-78.578125pt]panel_bottom.center)
  {\titlesplitimgnolines{290pt}{177.28125pt}{96.666667pt}
    {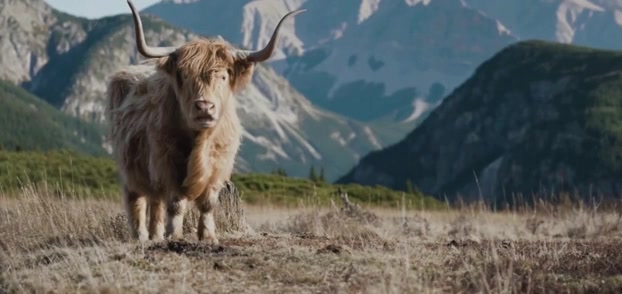}{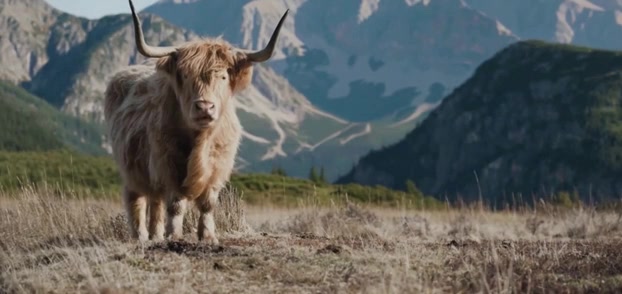}};

\node[anchor=west, inner sep=0pt] (copy_b_2)
  at ([xshift=110pt, yshift=-78.578125pt]panel_bottom.center)
  {\titlesplitimgnolines{290pt}{177.28125pt}{145pt}
    {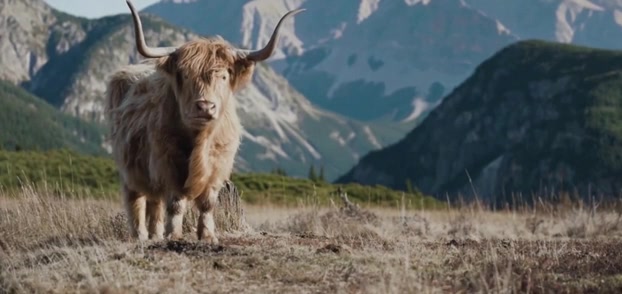}{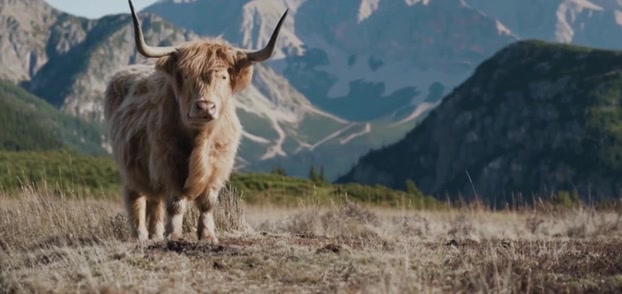}};

\node[anchor=west, inner sep=0pt] (copy_b_3)
  at ([xshift=408pt, yshift=-78.578125pt]panel_bottom.center)
  {\titlesplitimgnolines{290pt}{177.28125pt}{193.333333pt}
    {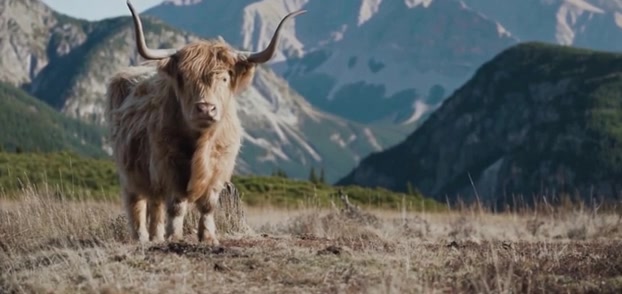}{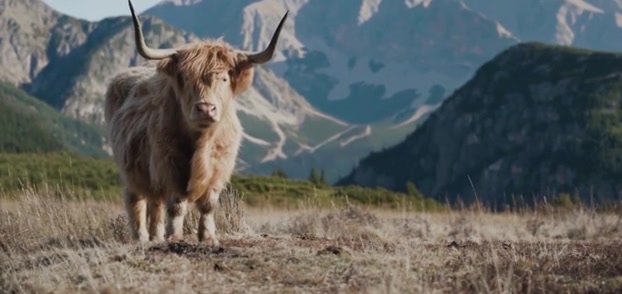}};

\node[titlelabel, anchor=north east]
  at ([xshift=-5pt, yshift=-5pt]copy_a_1.north east) {source};

\node[titlelabel, anchor=north east]
  at ([xshift=-5pt, yshift=-5pt]copy_b_1.north east) {target};
\coordinate (copy_b_1_joint) at ([xshift=96.666667pt, yshift=13pt]copy_b_1.south west);
\coordinate (copy_b_2_joint) at ([xshift=145pt, yshift=13pt]copy_b_2.south west);
\coordinate (copy_b_3_joint) at ([xshift=193.333333pt, yshift=13pt]copy_b_3.south west);
\foreach \joint in {copy_b_1_joint,copy_b_2_joint,copy_b_3_joint} {
  \path[fill=black, fill opacity=0.66, rounded corners=1pt]
    ([xshift=-77pt, yshift=9pt]\joint) --
    ([xshift=-88pt, yshift=0pt]\joint) --
    ([xshift=-77pt, yshift=-9pt]\joint) --
    ([xshift=83pt, yshift=-9pt]\joint) --
    ([xshift=94pt, yshift=0pt]\joint) --
    ([xshift=83pt, yshift=9pt]\joint) -- cycle;
  \node[font=\huge, text=white, anchor=east]
    at ([xshift=-7pt]\joint) {Dense};
  \node[font=\huge, text=white, anchor=west]
    at ([xshift=3pt]\joint) {Sol-Attn};
}
\foreach \img/\splitx in {copy_b_1/96.666667pt,copy_b_2/145pt,copy_b_3/193.333333pt} {
  \draw[white, line width=2pt]
    ([xshift=\splitx]\img.south west) -- ([xshift=\splitx]\img.north west);
}

\node[anchor=west, inner sep=0pt] (t2i_103)
  at ([xshift=-698pt, yshift=88.21875pt]panel_bottom.center)
  {\titlediagsplitimgnolines{239pt}{239pt}{79.666667pt}{159.333333pt}
    {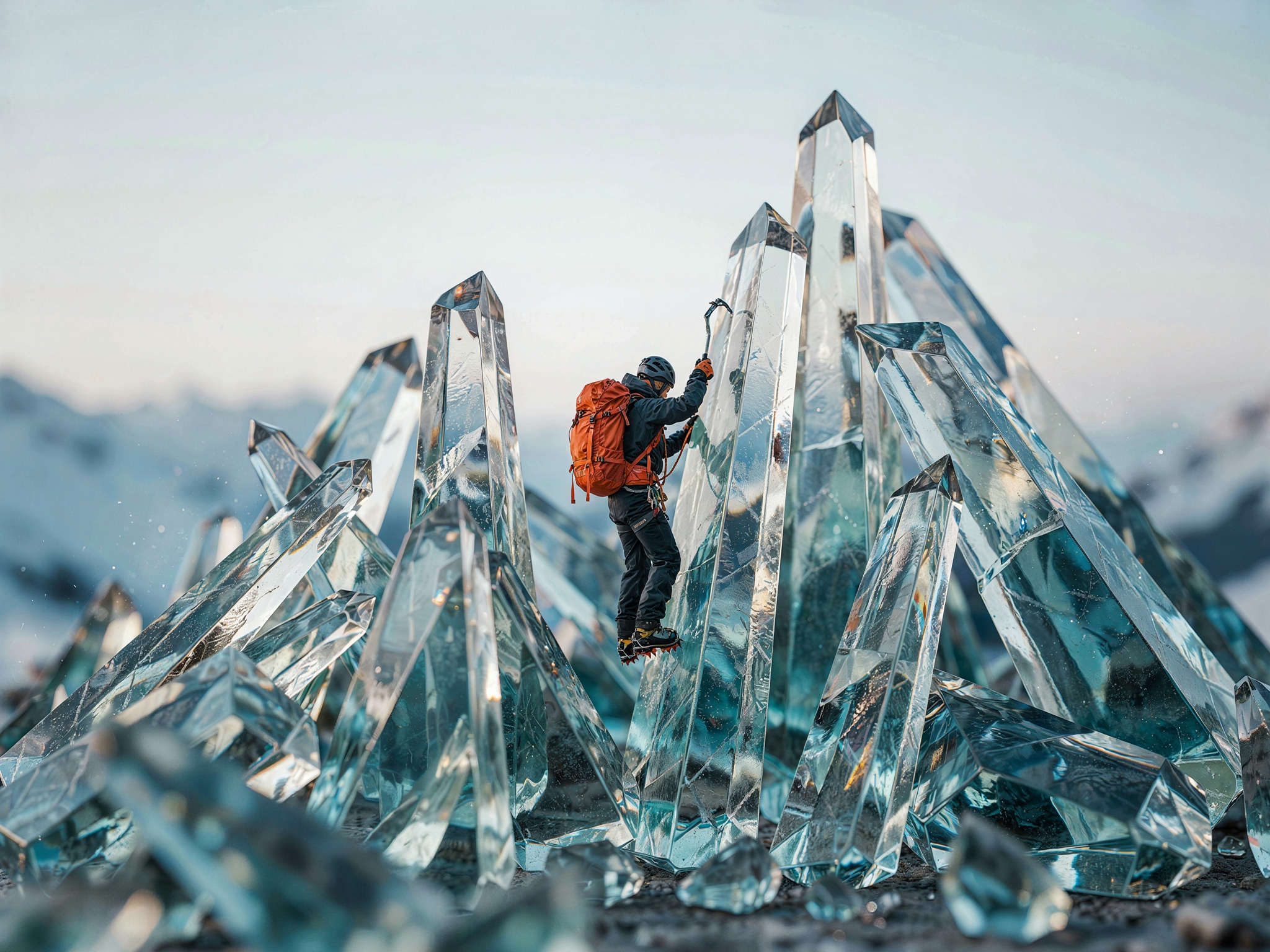}
    {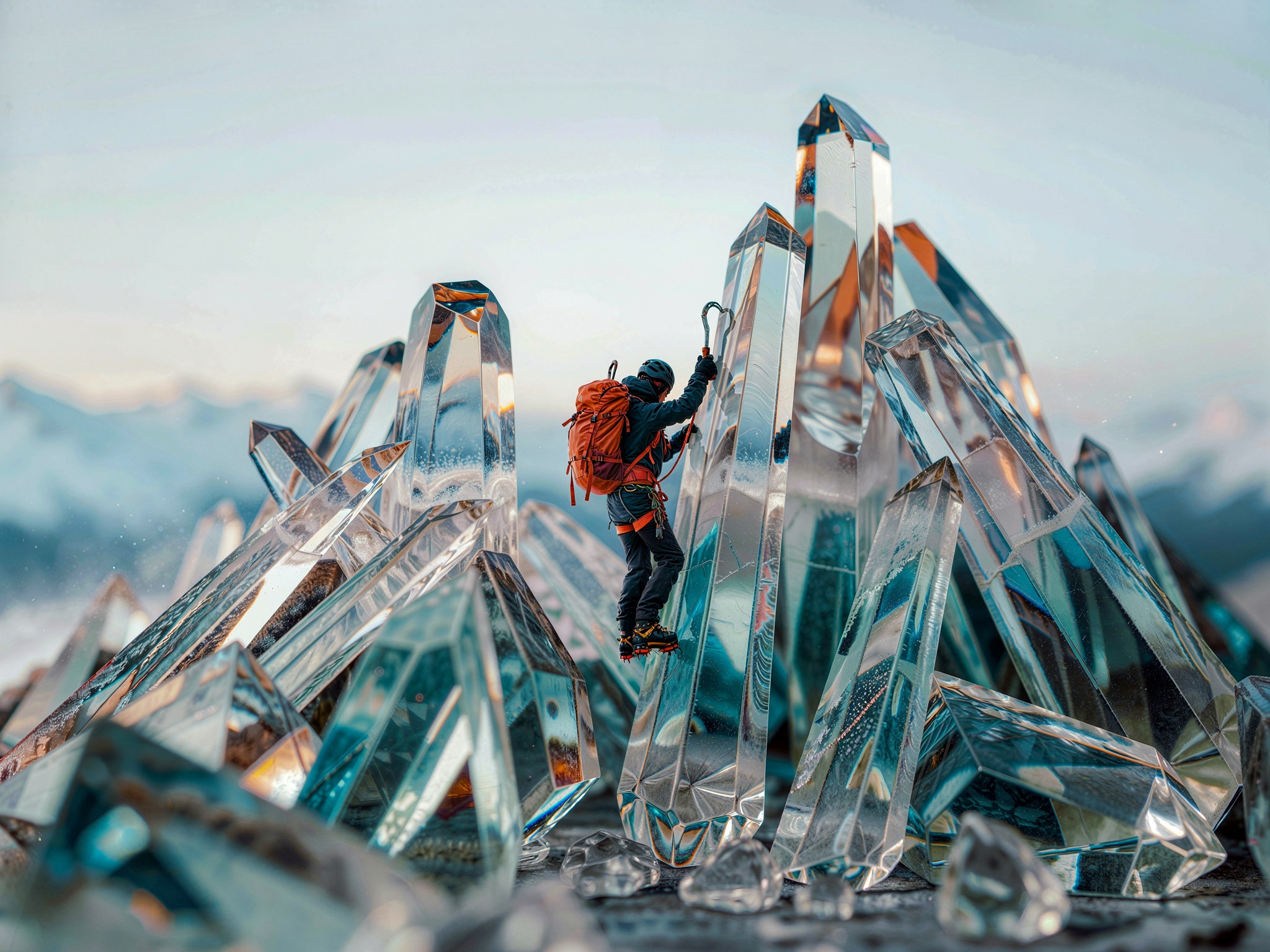}{height=239pt}};

\node[anchor=west, inner sep=0pt] (t2i_104)
  at ([xshift=-451pt, yshift=88.21875pt]panel_bottom.center)
  {\titlediagsplitimgnolines{239pt}{239pt}{79.666667pt}{159.333333pt}
    {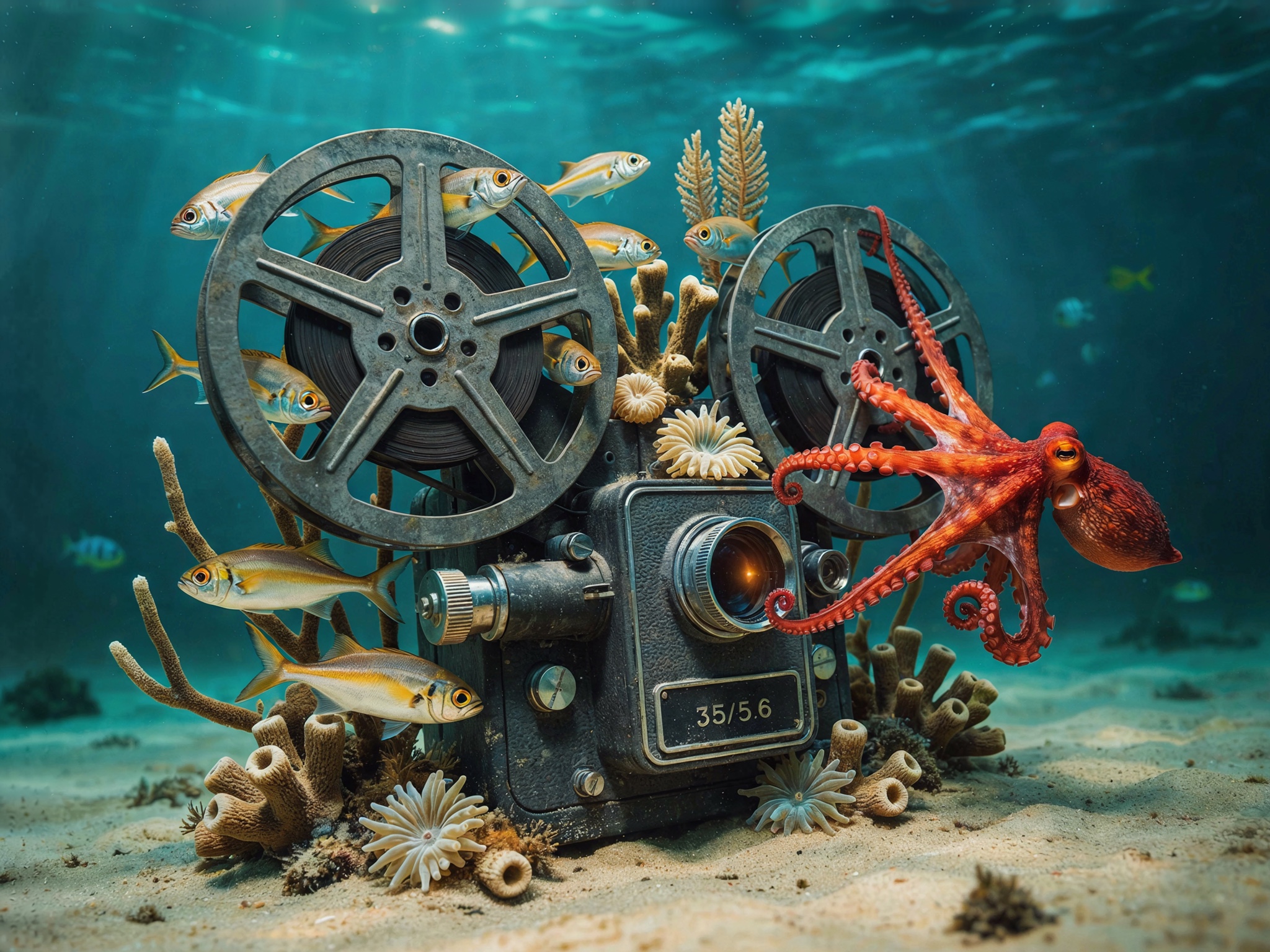}
    {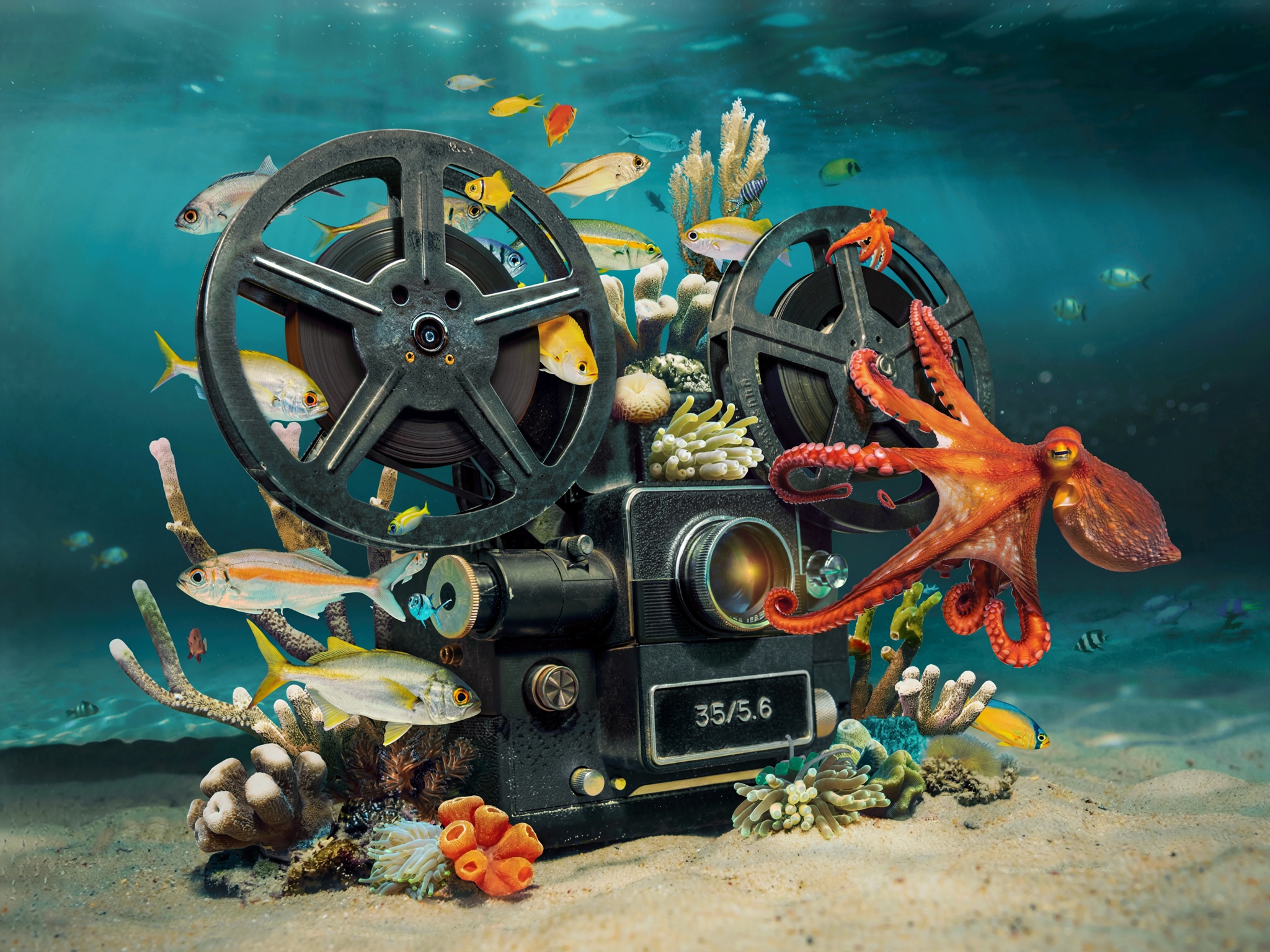}{height=239pt}};

\node[anchor=west, inner sep=0pt] (t2i_105)
  at ([xshift=-698pt, yshift=-100pt]panel_bottom.center)
  {\titlediagsplitimgnolines{239pt}{134.4375pt}{159.333333pt}{79.666667pt}
    {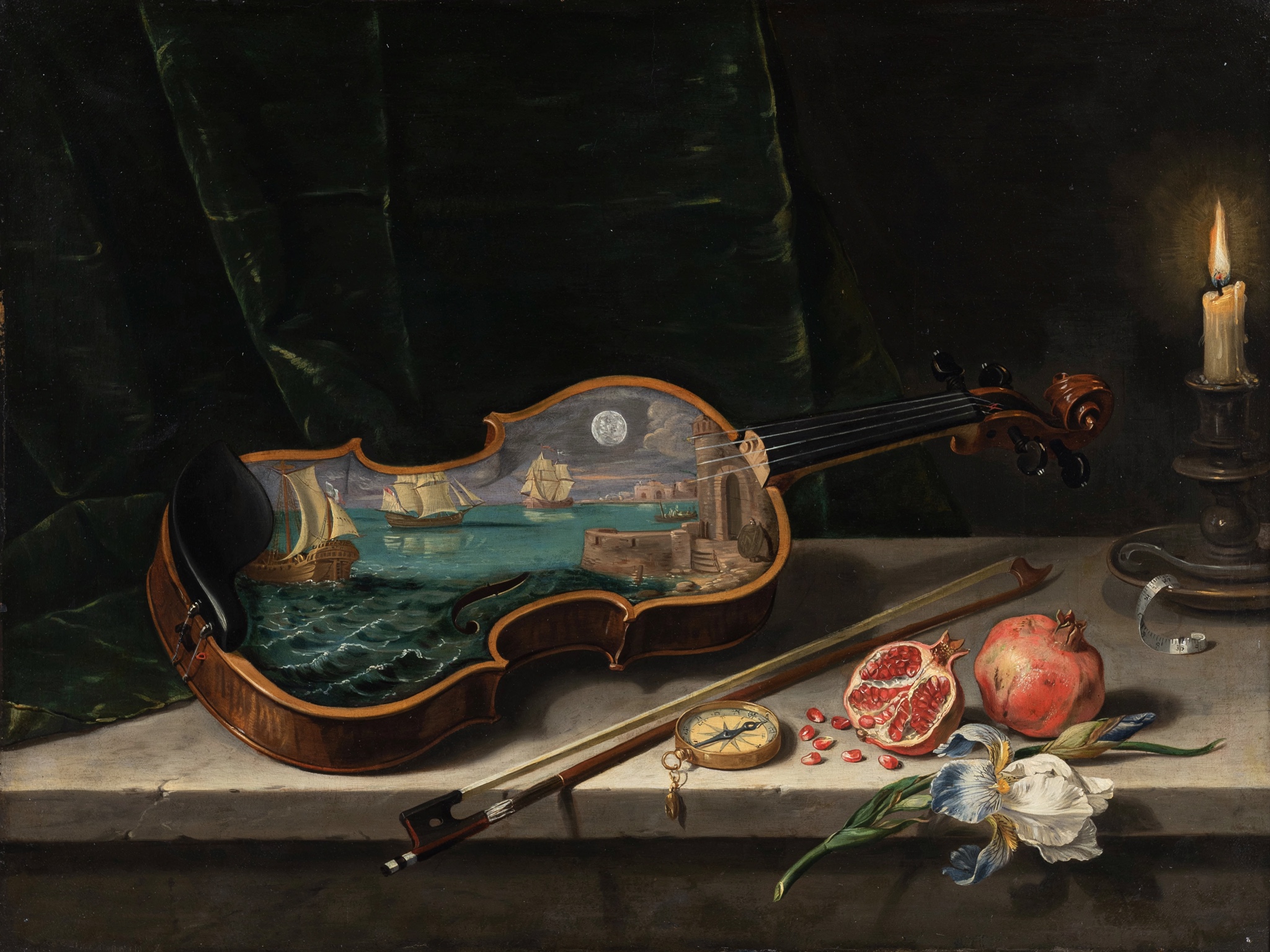}
    {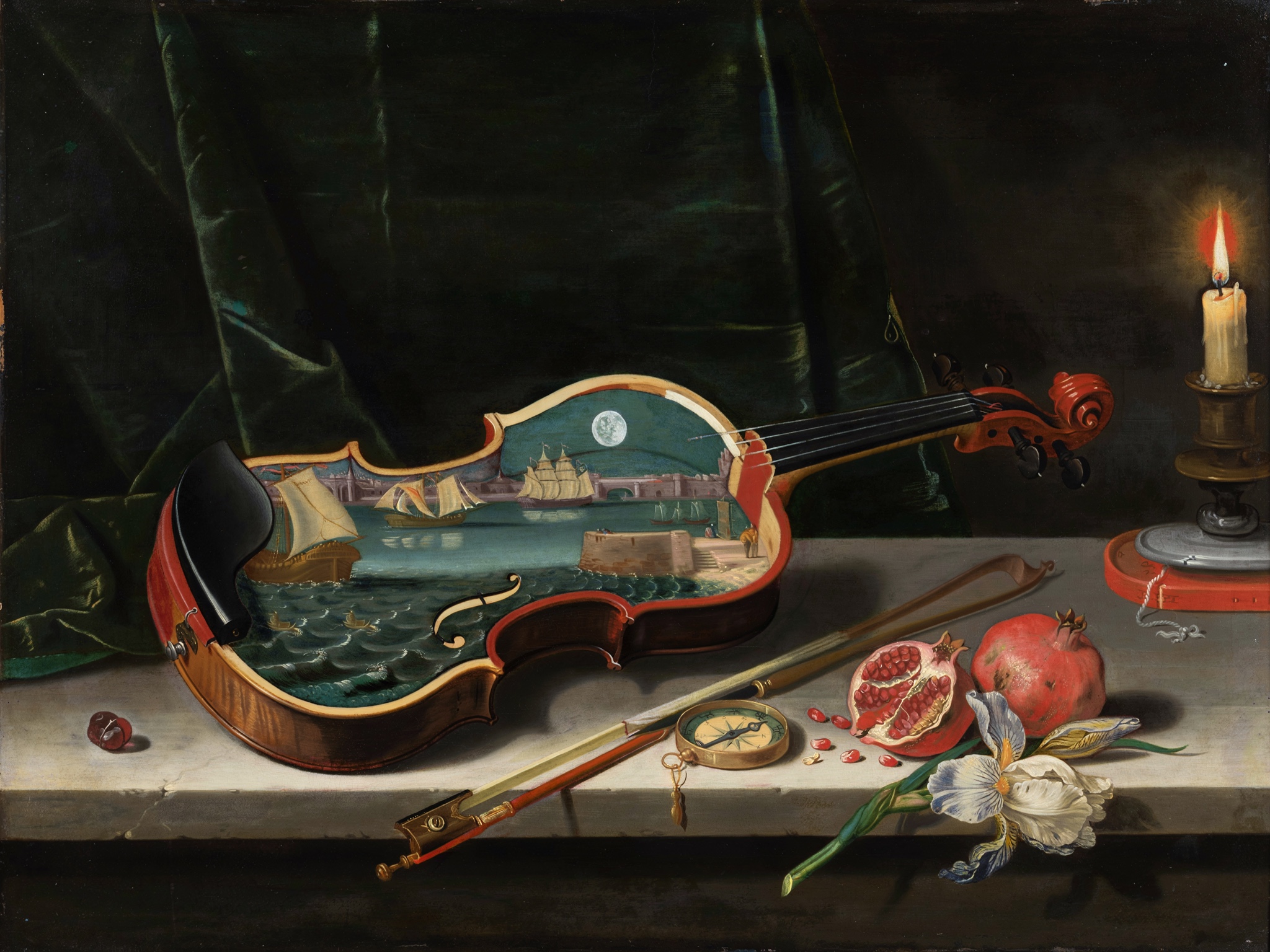}{width=239pt}};

\node[anchor=west, inner sep=0pt] (t2i_106)
  at ([xshift=-451pt, yshift=-100pt]panel_bottom.center)
  {\titlediagsplitimgnolines{239pt}{134.4375pt}{159.333333pt}{79.666667pt}
    {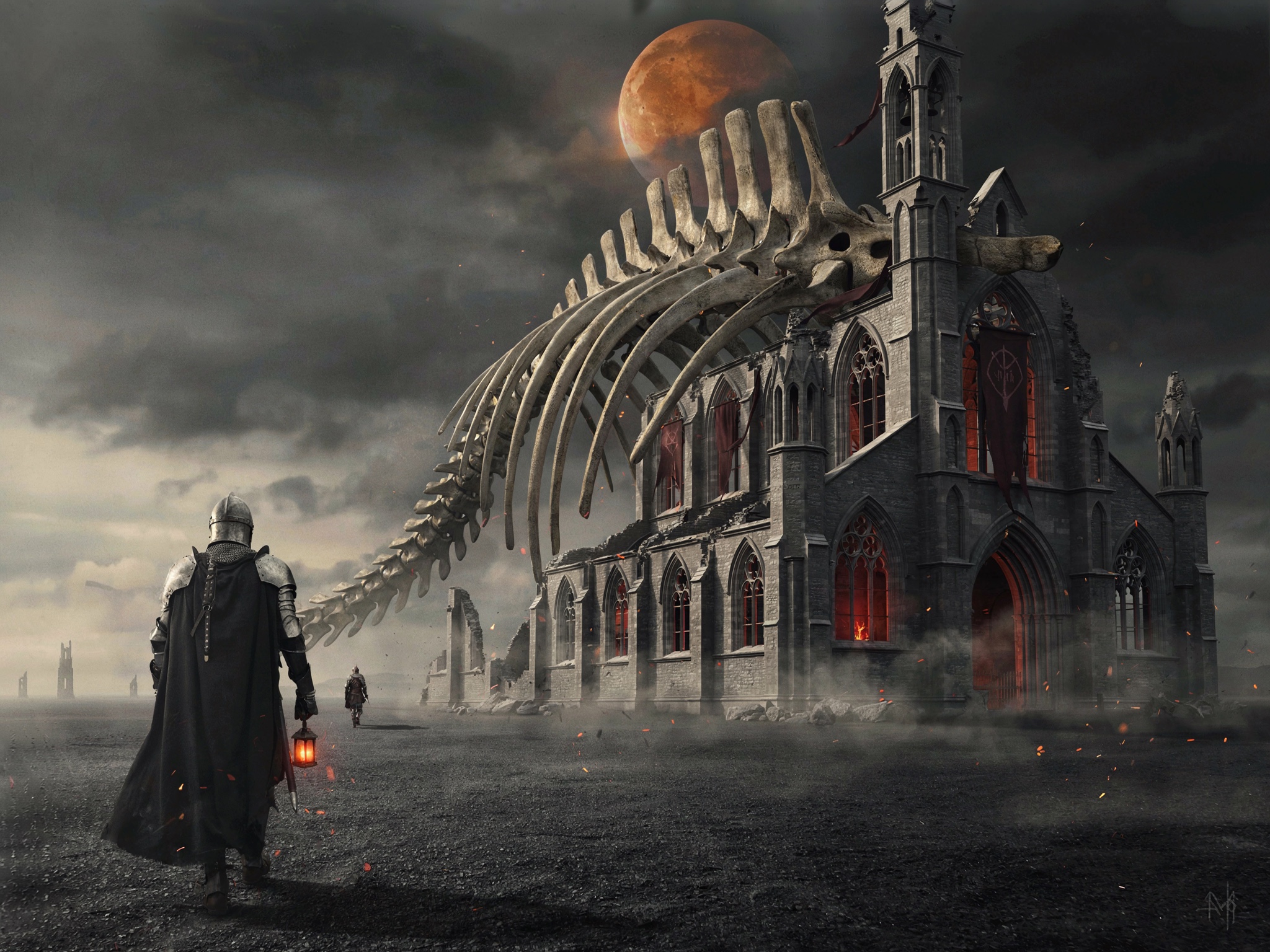}
    {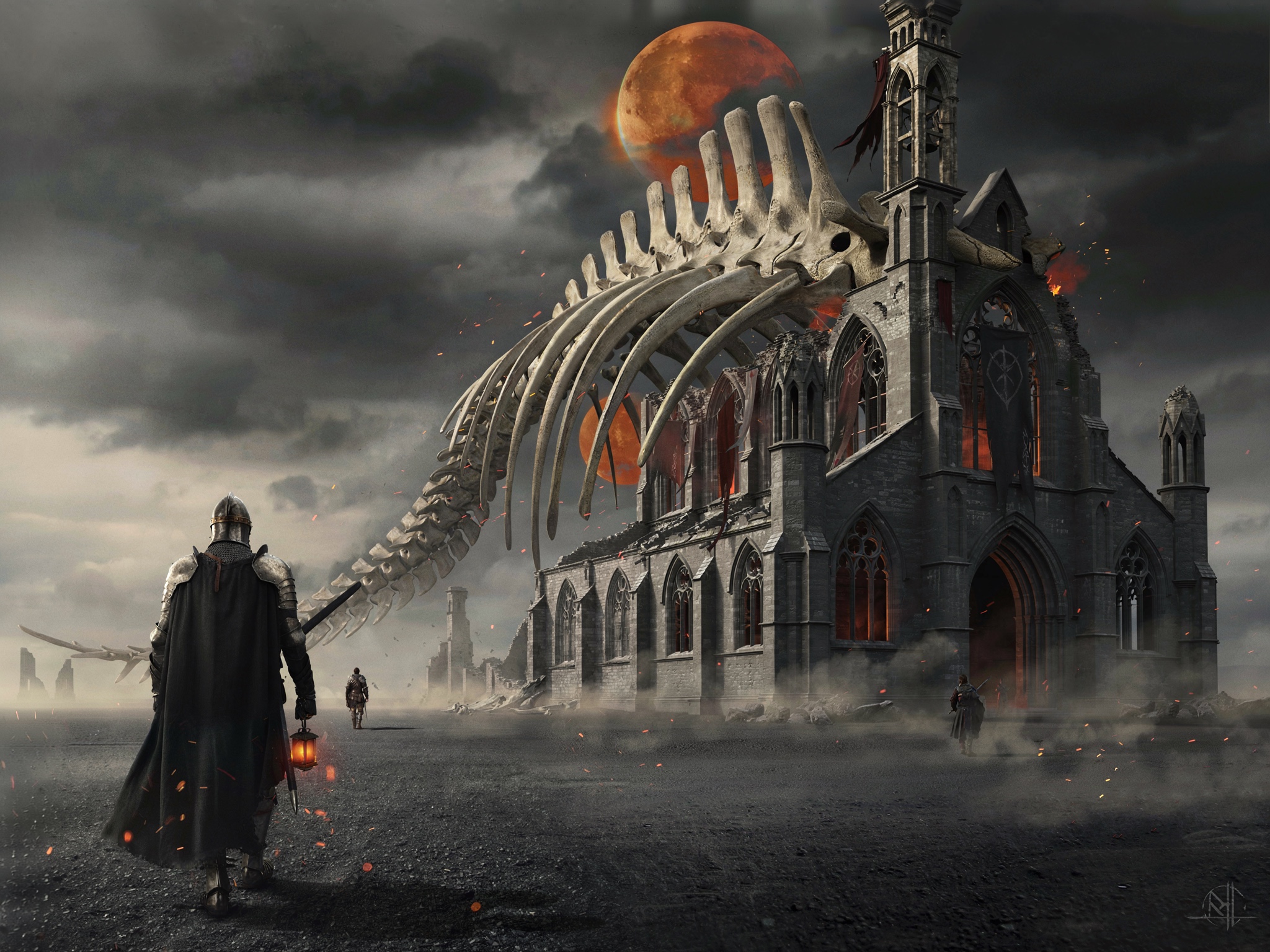}{width=239pt}};

\coordinate (t2i_103_joint) at ([xshift=159.333333pt]$(t2i_103.south west)!0.5!(t2i_105.north west)$);
\path[fill=black, fill opacity=0.66, rounded corners=1pt]
  ([xshift=-67pt, yshift=9pt]t2i_103_joint) --
  ([xshift=-78pt, yshift=0pt]t2i_103_joint) --
  ([xshift=-67pt, yshift=-9pt]t2i_103_joint) --
  ([xshift=79.666667pt, yshift=-9pt]t2i_103_joint) --
  ([xshift=79.666667pt, yshift=9pt]t2i_103_joint) -- cycle;
\node[font=\huge, text=white, anchor=east]
  at ([xshift=-7pt]t2i_103_joint) {Dense};
\node[font=\huge, text=white, anchor=west]
  at ([xshift=2pt]t2i_103_joint) {Sol-Attn};

\coordinate (t2i_104_joint) at ([xshift=159.333333pt]$(t2i_104.south west)!0.5!(t2i_106.north west)$);
\path[fill=black, fill opacity=0.66, rounded corners=1pt]
  ([xshift=-67pt, yshift=9pt]t2i_104_joint) --
  ([xshift=-78pt, yshift=0pt]t2i_104_joint) --
  ([xshift=-67pt, yshift=-9pt]t2i_104_joint) --
  ([xshift=79.666667pt, yshift=-9pt]t2i_104_joint) --
  ([xshift=79.666667pt, yshift=9pt]t2i_104_joint) -- cycle;
\node[font=\huge, text=white, anchor=east]
  at ([xshift=-7pt]t2i_104_joint) {Dense};
\node[font=\huge, text=white, anchor=west]
  at ([xshift=2pt]t2i_104_joint) {Sol-Attn};

\draw[white, line width=1.5pt]
  ([xshift=79.666667pt]t2i_103.north west) -- ([xshift=159.333333pt]t2i_103.south west);
\draw[white, line width=1.5pt]
  ([xshift=159.333333pt]t2i_105.north west) -- ([xshift=79.666667pt]t2i_105.south west);
\draw[white, line width=1.5pt]
  ([xshift=79.666667pt]t2i_104.north west) -- ([xshift=159.333333pt]t2i_104.south west);
\draw[white, line width=1.5pt]
  ([xshift=159.333333pt]t2i_106.north west) -- ([xshift=79.666667pt]t2i_106.south west);

\draw[modulelink, draw=black]
  ([xshift=-12pt]copy_a_1.north west)
  -- ([xshift=-12pt]copy_b_1.south west);

\node[font=\huge, anchor=west] (t2i_caption)
  at ([xshift=-690pt, yshift=28pt]panel_bottom.south)
  {(c) Sol-Attn accelerates text-to-image generation with minor differences.};

\node[font=\huge, anchor=center] (bottom_caption)
  at ([xshift=300pt, yshift=28pt]panel_bottom.south)
  {(d) Sol-Attn accelerates Bernini video editing by $2.3\times$ without quality degradation.};

\end{tikzpicture}
}

%% file: sections/1_intro.tex
\section{Introduction}
\label{sec:intro}

Diffusion transformers (DiTs)~\cite{peebles2023dit} have become the foundation of video generation~\cite{wan2025,kong2024hunyuanvideo,HaCohen2024LTXVideo,lightricks2026ltx23,bai2026mainecoon}. However, the pursuit of higher-resolution and longer-duration generation results in increasingly long token sequences, making the quadratic complexity of self-attention~\cite{vaswani2017attention} a dominant inference bottleneck. Training-free dynamic sparse attention is an attractive remedy because it can accelerate pretrained models without changing weights~\cite{yang2025sparse,zhang2025spargeattn,xu2025xattention,li2026pisa}. Yet current block-sparse methods still rely heavily on offline routing from compressed attention-score proxies and keep-or-drop sparsification, which expose two sharp limitations. \textbf{(L1) Block routing is rigid, hard to control, and costly.} 
For each query block, top-$k$ selects the $k$ key-value blocks with the highest proxy scores, thus enforcing a uniform budget across query blocks; top-$p$ retains blocks until their cumulative probability mass reaches $p$, risking severe budget imbalance when attention patterns are diffuse or sink-dominated.
Both strategies materialize the proxy score map and routing indices in HBM before sparse attention begins, incurring non-negligible memory traffic at long sequence lengths. \textbf{(L2) Keep-or-drop sparsification is lossy.} Unselected blocks are discarded entirely, even when they carry non-negligible attention mass, degrading accuracy under aggressive sparsity. These limitations point to a tension between routing efficiency and sparse-attention accuracy:
\textit{Can we achieve cheaper dynamic-budget routing while limiting accuracy degradation?}

We propose \ourmethod{}, a training-free sparse attention that fuses threshold-based dynamic sparsification into online softmax and reuses proxy scores to approximate unselected blocks, all within a single kernel, thereby achieving a better accuracy--efficiency trade-off. This design is realized through three steps:

\textbf{Query-dependent thresholding.} We observe that attention scores in pretrained video models are often near-symmetric and normal-like, suggesting that block selection can be performed by threshold-based filtering. Based on this insight, we propose a query-dependent threshold estimated from lightweight score statistics such as mean and standard deviation. This threshold selects blocks under a dynamic yet controllable budget, without materializing the full proxy map.

\textbf{On-the-fly sparsification.} During online softmax, each query block scans the pooled-key sequence tile by tile to produce proxy scores. These scores are compared on chip with the query-dependent threshold, and blocks above the threshold are immediately dispatched to a nested sparse attention loop over the original key-value tiles. Since each score is consumed as it is generated, no full proxy score map or routing indices need to be stored in HBM.

\textbf{Proxy-score reuse.} Instead of dropping proxy scores after routing, \ourmethod{} keeps them in registers during online softmax and reuses the below-threshold scores to approximate unselected blocks. This recovers part of the contribution from dropped blocks and reduces approximation error relative to dense attention. The exact and approximate branches share the same online-softmax state, unifying routing, sparse computation, and correction in one streaming operator.

Experiments across image generation, video generation, editing, and refinement show that \ourmethod{} advances the quality--efficiency frontier of training-free sparse attention while preserving generation quality. It delivers $2.1\times$--$3.0\times$ end-to-end speedups across video tasks and reaches $5.1\times$ when integrated with complementary optimizations in Sol-Engine.

%% file: sections/2_preliminary.tex
\section{Preliminaries}
\label{sec:preliminary}

\paragraph{Attention and Online Softmax.}
\label{sec:prelim_dense}

Given query, key, and value matrices
$\boldsymbol{Q},\boldsymbol{K},\boldsymbol{V}\in\mathbb{R}^{L\times d}$, where
$L$ is the sequence length and $d$ is the feature dimension, softmax attention~\cite{vaswani2017attention}
computes the output $\boldsymbol{O}\in\mathbb{R}^{L\times d}$ (omitting the scale factor):
\begin{equation}
    \boldsymbol{O}
    =
    \operatorname{Softmax}\left(
        \boldsymbol{Q}\boldsymbol{K}^{\top}
    \right)\boldsymbol{V}.
    \label{eq:dense_attn}
\end{equation}
FlashAttention~\cite{dao2022flashattention}
introduces online softmax to avoid materializing the full attention matrix
$\operatorname{Softmax}(\boldsymbol{Q}\boldsymbol{K}^{\top})\in\mathbb{R}^{L\times L}$
in HBM via tiled on-chip computation, thereby reducing I/O overhead and
eliminating the $\mathcal{O}(L^2)$ memory footprint. However, it still evaluates
all query-key interactions and therefore keeps the quadratic compute cost.

\vspace{-0.8em}
\paragraph{Block-Sparse Attention.}
\label{sec:prelim_sparse}

To reduce the $\mathcal{O}(L^2)$ computational complexity, sparse attention~\cite{xu2025xattention,zhang2025spargeattn,shao2026liveditor,li2026pisa,yang2025sparse,jiang2024minference,xiao2025optimizing} computes only a
subset of critical key-value pairs for each query. Mathematically, it can be
formulated as:
\begin{equation}
    \boldsymbol{O}
    =
    \operatorname{Softmax}\left(
        \boldsymbol{Q}\boldsymbol{K}^{\top}
        + \boldsymbol{M}
    \right)\boldsymbol{V},
    \label{eq:block_sparse_attn}
\end{equation}
where $\boldsymbol{M}\in\{0,-\infty\}^{L\times L}$ is the attention mask, with
$-\infty$ entries ignored by the softmax.
To align sparse attention with hardware-friendly tiled online softmax, the
sparse pattern is organized at block granularity, allowing masked key-value
tiles to be skipped entirely and translating sparsity into compute savings.

\begin{figure}[t]
    \centering
    \includegraphics[width=\linewidth]{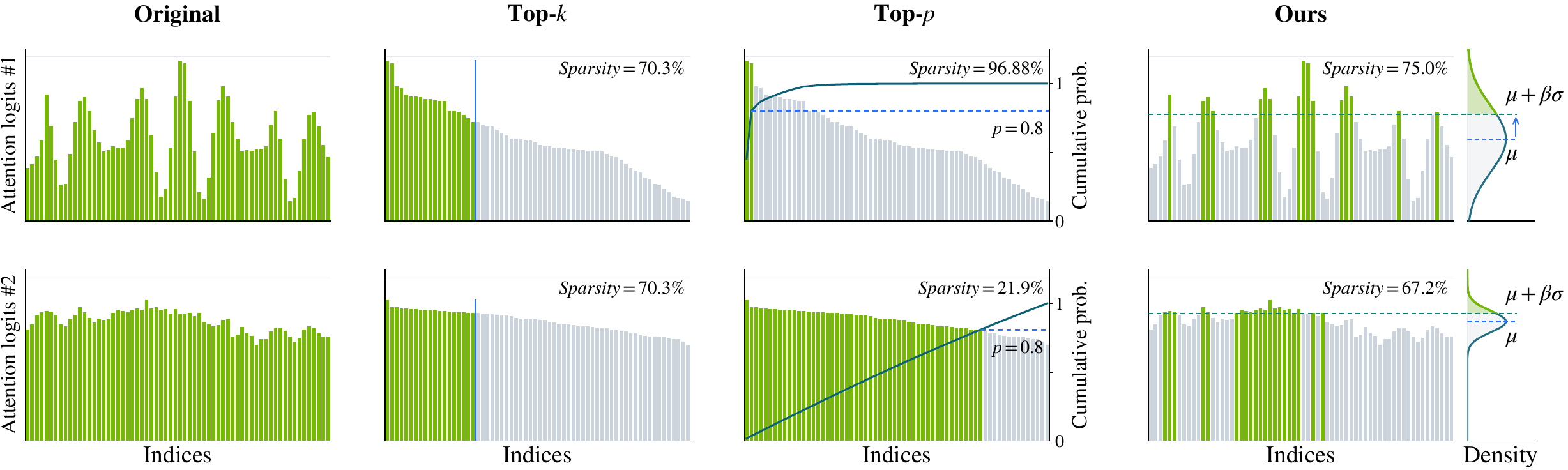}
    \caption{\textbf{Comparison of routing strategies.}
    Bars represent pre-softmax block-proxy logits, with selected blocks highlighted in green.
    Top-$k$ selects the highest-scoring blocks under a fixed budget.
    Top-$p$ accumulates the softmax probabilities to a target $p$, yielding a dynamic budget that is sensitive to the probability distribution and thus hard to balance.
    \ourmethod{} instead derives a threshold from the mean $\mu$ and standard deviation $\sigma$ of the block-proxy logits, with $\beta$ setting the standardized offset, yielding a dynamic yet controllable budget without materializing the block-proxy map.
    }
    \label{fig:routing_comparison}
\end{figure}

To dynamically construct $\boldsymbol{M}$ under this block structure, standard
methods typically rely on block-wise proxy attention scores. Let the sequence be
partitioned into $N$ blocks with block size $B$, and denote the $i$-th query
block as $\boldsymbol{Q}_i\in\mathbb{R}^{B\times d}$ and the $j$-th key-value
block as $\boldsymbol{K}_j,\boldsymbol{V}_j\in\mathbb{R}^{B\times d}$.
The block scores are proxied by:
\begin{equation}
    \widehat{s}_{ij}
    =
    \bar{\boldsymbol{Q}}_i
    \bar{\boldsymbol{K}}_j^{\top},
    \qquad
    \widehat{\boldsymbol{p}}_i
    =
    \operatorname{Softmax}(\widehat{\boldsymbol{s}}_i),
    \label{eq:block_proxy}
\end{equation}
where $\bar{\boldsymbol{Q}}_i:=\operatorname{Mean}(\boldsymbol{Q}_i)$,
$\bar{\boldsymbol{K}}_j:=\operatorname{Mean}(\boldsymbol{K}_j) \in \mathbb{R}^{1\times d}$ denote means along the token dimension;
$\widehat{\boldsymbol{s}}_i=[\widehat{s}_{i1},\ldots,\widehat{s}_{iN}]$
collects the proxy scores between query block $i$ and all key blocks. Standard routing applies top-$k$ to $\widehat{\boldsymbol{s}}_i$ or top-$p$ to $\widehat{\boldsymbol{p}}_i$ to select key-value blocks, which are then marked as active in the block-sparse mask $\boldsymbol{M}$. As discussed in
Sec.~\ref{sec:intro} and illustrated in Figure~\ref{fig:routing_comparison},
top-$k$ enforces a rigid fixed budget, whereas top-$p$ produces a dynamic but
hard-to-control budget; both additionally traverse the complete proxy row
before sparse computation, incurring non-negligible routing overhead.

%% file: sections/3_method.tex
\section{Methodology}
\label{sec:method}

To enable dynamic sparse attention to adaptively allocate its compute budget across varying softmax-score distributions without costly routing, 
we propose \ourmethod{}, a training-free sparse attention method for video diffusion transformers that unifies proxy
routing, sparse computation, and approximate correction within a single online-softmax pass, thereby reducing routing
overhead while improving approximation accuracy. We first derive query-dependent thresholds, then
integrate routing into tile-wise online softmax, and finally reuse the tile-wise routing
score for approximate correction.

\subsection{Query-Dependent Thresholding}
\label{sec:method_threshold}

\begin{wrapfigure}[12]{r}{0.50\textwidth}
    \vspace{-\intextsep}
    \centering
    \includegraphics[width=0.96\linewidth]{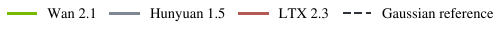}
    \begin{minipage}[t]{0.60\linewidth}
        \raggedright
        \includegraphics[height=80pt]{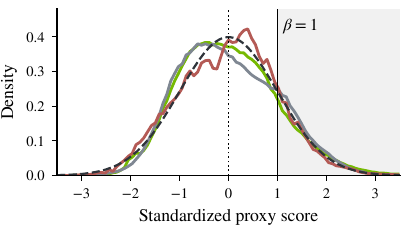}
    \end{minipage}\hfill
    \begin{minipage}[t]{0.38\linewidth}
        \raggedleft
        \includegraphics[height=80pt]{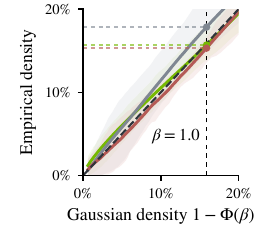}
    \end{minipage}
    \vspace{\baselineskip}
    \caption{\textbf{Gaussian-calibrated density control.}
    Standardized block-proxy logits are near-Gaussian (left); empirical
    densities match the predicted tail at cutoff $\beta$ (right).
    }
    \label{fig:threshold_density}
\end{wrapfigure}

\newcommand{\routingoverheadwrap}{%
\begin{wrapfigure}[14]{r}{0.5\textwidth}
    \vspace{-32pt}
    \centering
    \definecolor{figFourQuery}{HTML}{F3CE7F}
    \definecolor{figFourKey}{HTML}{8EBDEC}
    \definecolor{figFourThreshold}{HTML}{BD8083}
    \resizebox{\linewidth}{!}{%
    \begin{tikzpicture}[x=1pt,y=1pt,font=\scriptsize,
        line cap=round,line join=round,>=latex]
        \useasboundingbox (10,27) rectangle (202,92);
        \node[font=\fontsize{6.46}{7.0}\selectfont,inner sep=0pt,anchor=base]
            at (60,83.90) {Top-$k/p$ routing};
        \node[font=\fontsize{6.46}{7.0}\selectfont,inner sep=0pt,anchor=base]
            at (158,83.90) {Thresholding};
        \def\figFourCell{9.333333}
        \def\figFourMapBottom{33}

        \foreach \c/\idx in {0/1,1/2,2/3,3/4,4/5,5/6} {
            \fill[figFourKey] ({32+\figFourCell*\c},66) rectangle ({32+\figFourCell*(\c+1)},{66+\figFourCell});
            \draw[black!55,line width=0.35pt] ({32+\figFourCell*\c},66) rectangle ({32+\figFourCell*(\c+1)},{66+\figFourCell});
            \node[font=\fontsize{3.15}{3.5}\selectfont,inner sep=0pt]
                at ({32+\figFourCell*(\c+0.5)},{66+0.5*\figFourCell}) {$\bar{K}_{\idx}$};
        }
        \draw[black!80,line width=0.65pt] (32,66) rectangle (88,{66+\figFourCell});
        \foreach \r/\idx in {0/3,1/2,2/1} {
            \fill[figFourQuery] ({27-\figFourCell},{\figFourMapBottom+\figFourCell*\r}) rectangle (27,{\figFourMapBottom+\figFourCell*(\r+1)});
            \draw[black!55,line width=0.35pt] ({27-\figFourCell},{\figFourMapBottom+\figFourCell*\r}) rectangle (27,{\figFourMapBottom+\figFourCell*(\r+1)});
            \node[font=\fontsize{3.15}{3.5}\selectfont,inner sep=0pt]
                at ({27-0.5*\figFourCell},{\figFourMapBottom+\figFourCell*(\r+0.5)}) {$\bar{Q}_{\idx}$};
        }
        \draw[black!80,line width=0.65pt] ({27-\figFourCell},\figFourMapBottom) rectangle (27,61);
        \fill[black!4] (32,\figFourMapBottom) rectangle (88,61);
        \foreach \g in {0,...,6} {
            \draw[black!22,line width=0.3pt] ({32+\figFourCell*\g},\figFourMapBottom) -- ({32+\figFourCell*\g},61);
        }
        \foreach \g in {0,...,3} {
            \draw[black!22,line width=0.3pt] (32,{\figFourMapBottom+\figFourCell*\g}) -- (88,{\figFourMapBottom+\figFourCell*\g});
        }
        \draw[black!80,line width=0.65pt] (32,\figFourMapBottom) rectangle (88,61);
        \foreach \c/\idx in {0/1,1/2,2/3,3/4,4/5,5/6} {
            \fill[figFourKey] ({130+\figFourCell*\c},66) rectangle ({130+\figFourCell*(\c+1)},{66+\figFourCell});
            \draw[black!55,line width=0.35pt] ({130+\figFourCell*\c},66) rectangle ({130+\figFourCell*(\c+1)},{66+\figFourCell});
            \node[font=\fontsize{3.15}{3.5}\selectfont,inner sep=0pt]
                at ({130+\figFourCell*(\c+0.5)},{66+0.5*\figFourCell}) {$\bar{K}_{\idx}$};
        }
        \draw[black!80,line width=0.65pt] (130,66) rectangle (186,{66+\figFourCell});
        \foreach \r/\idx in {0/3,1/2,2/1} {
            \fill[figFourQuery] ({125-\figFourCell},{\figFourMapBottom+\figFourCell*\r}) rectangle (125,{\figFourMapBottom+\figFourCell*(\r+1)});
            \draw[black!55,line width=0.35pt] ({125-\figFourCell},{\figFourMapBottom+\figFourCell*\r}) rectangle (125,{\figFourMapBottom+\figFourCell*(\r+1)});
            \node[font=\fontsize{3.15}{3.5}\selectfont,inner sep=0pt]
                at ({125-0.5*\figFourCell},{\figFourMapBottom+\figFourCell*(\r+0.5)}) {$\bar{Q}_{\idx}$};
            \fill[figFourThreshold] (191,{\figFourMapBottom+\figFourCell*\r}) rectangle ({191+\figFourCell},{\figFourMapBottom+\figFourCell*(\r+1)});
            \draw[black!55,line width=0.35pt] (191,{\figFourMapBottom+\figFourCell*\r}) rectangle ({191+\figFourCell},{\figFourMapBottom+\figFourCell*(\r+1)});
            \node[font=\fontsize{3.15}{3.5}\selectfont,inner sep=0pt]
                at ({191+0.5*\figFourCell},{\figFourMapBottom+\figFourCell*(\r+0.5)}) {$\tau_{\idx}$};
        }
        \draw[black!80,line width=0.65pt] ({125-\figFourCell},\figFourMapBottom) rectangle (125,61);
        \draw[black!80,line width=0.65pt] (191,\figFourMapBottom) rectangle ({191+\figFourCell},61);
        \draw[black!30,densely dashed,line width=0.45pt] (130,\figFourMapBottom) rectangle (186,61);
        \fill[nvidiagreen!38]
            ({130+\figFourCell},{\figFourMapBottom+\figFourCell})
            rectangle ({130+2*\figFourCell},{\figFourMapBottom+2*\figFourCell});
        \draw[black!80,line width=0.55pt]
            ({130+\figFourCell},{\figFourMapBottom+\figFourCell})
            rectangle ({130+2*\figFourCell},{\figFourMapBottom+2*\figFourCell});
        \fill[figFourThreshold]
            (158,{\figFourMapBottom+\figFourCell})
            rectangle ({158+\figFourCell},{\figFourMapBottom+2*\figFourCell});
        \draw[black!80,line width=0.55pt]
            (158,{\figFourMapBottom+\figFourCell})
            rectangle ({158+\figFourCell},{\figFourMapBottom+2*\figFourCell});
        \node[font=\normalsize]
            at (153.33,{\figFourMapBottom+1.5*\figFourCell}) {$>$};
        \draw[->,black!75,line width=0.65pt]
            ({130+1.5*\figFourCell},66)
            -- ({130+1.5*\figFourCell},{\figFourMapBottom+2*\figFourCell});
        \draw[->,black!75,line width=0.65pt]
            (125,{\figFourMapBottom+1.5*\figFourCell})
            -- ({130+\figFourCell},{\figFourMapBottom+1.5*\figFourCell});
        \draw[->,black!75,line width=0.65pt]
            (191,{\figFourMapBottom+1.5*\figFourCell})
            -- ({158+\figFourCell},{\figFourMapBottom+1.5*\figFourCell});
    \end{tikzpicture}
    }
    \captionsetup{skip=0pt}
    \caption{
    Top-$k/p$ materializes proxy map in HBM, whereas
    thresholding streams scores on chip.}
    \vspace{-32pt}
    \label{fig:routing_overhead}
\end{wrapfigure}%
}

\paragraph{Structured Attention Logits.}
Across models, we aggregate the row-wise pre-softmax block-proxy logits $\hat{s}_{ij}$ in
Eq.~\eqref{eq:block_proxy} over denoising steps, layers, attention heads, and
query blocks. The resulting distribution is consistently near-Gaussian within
each model, as shown in Figure~\ref{fig:threshold_density}. This aggregate
structure provides a natural routing coordinate: for a Gaussian variable, a
cutoff $\beta$ standard deviations above the mean corresponds to a unique
upper-tail density. A shared $\beta$ can therefore control the overall
sparsity level of a model, while each row maps this common standardized cutoff
back to its own raw-logit scale through its row-wise mean and standard
deviation rather than sharing a fixed raw-score cutoff.

\paragraph{Threshold-Based Routing.}
Motivated by this observation, we propose query-dependent threshold routing. For
query block $i$, let $\mu_i$ and $\sigma_i$ denote the mean and standard
deviation of its proxy row $\widehat{\boldsymbol{s}}_i$. Given a shared
standardized cutoff $\beta>0$, we define its routing threshold
$\tau_i$ and the set of selected key-block indices $\mathcal{S}_i$ as
\begin{equation}
    \tau_i=\mu_i+\beta\sigma_i,
    \qquad
    \mathcal{S}_i
    =\{j:\widehat{s}_{ij}>\tau_i\}.
    \label{eq:threshold_routing}
\end{equation}

Let $\Phi$ denote the standard Gaussian CDF. Under the Gaussian reference,
$\beta$ corresponds to the selected density
$\rho_{\mathrm G}(\beta):=1-\Phi(\beta)$, while $\mu_i$ and $\sigma_i$ map the
shared standardized cutoff to the raw-score scale of query $i$. Figure~\ref{fig:threshold_density} shows that the empirical mean densities
remain close to $\rho_{\mathrm G}(\beta)$ across the evaluated models, while
the shaded bands span the 10th--90th percentiles of the per-row densities.
Thus, $\beta$ calibrates the model-level mean density, whereas
$|\mathcal{S}_i|$ remains query-dependent rather than fixed as in
top-$k$. Top-$p$ also permits variable counts, but $p$ controls cumulative
mass rather than block density: depending on the concentration of the softmax-score distribution, the same $p$ may retain between one and
$\lceil pN\rceil$ blocks. Threshold routing therefore combines a dynamic
budget with an empirically calibrated density scale.

\vspace{-4pt}
\paragraph{Efficient Threshold Computation.}
The remaining question is how to efficiently obtain $\tau_i$ without materializing the full proxy-score row $\widehat{\boldsymbol{s}}_i$. Since $\widehat{\boldsymbol{s}}_i$ is obtained by linearly projecting the pooled keys with $\bar{\boldsymbol{Q}}_i$, its mean and variance can be computed directly from the first and second moments of the pooled-keys (see Appendix~\ref{sec:threshold_variance_derivation} for the derivation):
\begin{equation}
    \mu_i
    = \bar{\boldsymbol{Q}}_i
      \left(\frac{1}{N}\sum\nolimits_{j=1}^{N}\bar{\boldsymbol{K}}_j\right)^{\top},
    \qquad
    \sigma_i^2
    =
      \bar{\boldsymbol{Q}}_i
      \left(
      \frac{1}{N}\sum\nolimits_{j=1}^{N}
      \bar{\boldsymbol{K}}_j^{\top}\bar{\boldsymbol{K}}_j
      \right)
      \bar{\boldsymbol{Q}}_i^{\top}
      - \mu_i^2.
    \label{eq:threshold_stats_exact}
\end{equation}
All row-wise means $\mu_i$ and variances $\sigma_i^2$, and hence the
query-dependent thresholds $\tau_i$, can therefore be computed in
$\mathcal{O}(Ld+Nd^2)$ time using $\mathcal{O}(d^2)$ auxiliary storage, without
materializing the $N\times N$ proxy score map.

\subsection{\ourmethod{}: Unified On-the-Fly Sparsification}
\label{sec:method_sparsification}

\paragraph{Chunk-wise On-the-Fly Thresholding.}
Once the threshold \(\tau_i\) for the \(i\)-th query block is available, each key-value block can be selected as soon as its proxy score is produced, without materializing the full proxy-score row. This property allows block selection to proceed directly within online softmax. 
We regard the block-wise pooled keys $\{\bar{\boldsymbol{K}}_j\}_{j=1}^{N}$ as
a new key sequence of length $N$ and partition it into
$T$ chunks of size $C$. 
For chunk $t$, let
$\bar{\boldsymbol{K}}^{{(t)}}=[\bar{\boldsymbol{K}}_{tC+1};\ldots;
\bar{\boldsymbol{K}}_{(t+1)C}]\in\mathbb{R}^{C\times d}$. Each block $\boldsymbol{Q}_i$ scans these
chunks sequentially to produce chunk-wise proxy scores:
\begin{equation}
    \widehat{\boldsymbol{s}}_i^{(t)}
    = \bar{\boldsymbol{Q}}_i (\bar{\boldsymbol{K}}^{(t)})^{\top}
    \in \mathbb{R}^{1\times C}.
    \label{eq:stream_proxy}
\end{equation}
Applying Eq.~\eqref{eq:threshold_routing} to each chunk yields
\begin{equation}
\mathcal{S}_i^{(t)}
=
\big\{tC+r
\mid
r\in\{1,\ldots,C\},
[\widehat{\boldsymbol{s}}_i^{(t)}]_r>\tau_i\big\}.
\end{equation}
Since
$\mathcal{S}_i=\bigcup_{t=0}^{T-1}\mathcal{S}_i^{(t)}$,
chunk-wise streaming thresholding recovers the same block set as full-row thresholding.
The selected blocks in each $\mathcal{S}_i^{(t)}$ are immediately consumed by sparse
attention, naturally yielding a \textit{nested-loop structure}:
\begin{itemize}[leftmargin=1.1em,itemsep=0.15em,topsep=0.2em,parsep=0pt]
    \item \textbf{Outer loop} densely scans the pooled-key sequence chunk by chunk to determine the chunk-wise selected set.
    \item \textbf{Inner loop} sparsely traverses the key-value blocks selected within the current chunk to perform attention computation.
\end{itemize}
This nested-loop structure enables on-the-fly attention sparsification that is mathematically equivalent to offline global thresholding, without materializing the full proxy-score map or routing indices as intermediate tensors.

\paragraph{Proxy-Score Reuse as Approximate Correction.}

Blocks whose proxy scores fall below the threshold would otherwise be discarded by sparse attention. To retain their contributions at low cost, we approximate their block-wise exponential score matrices. Specifically, we apply the Taylor expansion around the row-wise mean score within each key block:
\begin{equation}
    \exp(\boldsymbol{Q}_i\boldsymbol{K}_j^{\top})
    =
    \exp(\boldsymbol{Q}_i\bar{\boldsymbol{K}}_j^{\top})
    \odot
    \left[
        \boldsymbol{1}
        +
        \boldsymbol{Q}_i
        (\boldsymbol{K}_j-\bar{\boldsymbol{K}}_j)^{\top}
        +
        \mathcal{O}\!\left(
            \left(
                \boldsymbol{Q}_i
                (\boldsymbol{K}_j-\bar{\boldsymbol{K}}_j)^{\top}
            \right)^{\odot 2}
        \right)
    \right],
    \label{eq:zeroth_order_expansion}
\end{equation}
where $\odot$ denotes element-wise multiplication and the $B\times1$ center is broadcast along the key dimension. Retaining only the zeroth-order term approximates the entire block-wise exponential score matrix using its pooled key, thereby preserving approximate contributions from skipped blocks to both the softmax numerator and denominator.

\pagebreak
For each block $j$, let
$\widehat{\boldsymbol{V}}_j\in\mathbb{R}^{1\times d}$ denote the sum of
$\boldsymbol{V}_j$ along the token dimension, and let
$\mathcal{U}_i:=\{1,\ldots,N\}\setminus\mathcal{S}_i$ denote the set of
unselected blocks. Replacing the exponential scores of blocks in
$\mathcal{U}_i$ with their zeroth-order approximations, while retaining exact
computation for blocks in $\mathcal{S}_i$, yields the combined softmax
denominator and numerator:
\definecolor{figpooledkeyblue}{HTML}{B5B0F5}
\definecolor{figkeyblue}{HTML}{84BAE9}
\definecolor{figqueryyellow}{HTML}{FFD965}
\newcommand{\figMathBoxStrut}{\vphantom{\bar{\boldsymbol{K}}_j^{\top}}}
\newcommand{\figQueryMath}[1]{%
    \begingroup
    \setlength{\fboxsep}{0.5pt}%
    \colorbox{figqueryyellow}{\(\color{black}\figMathBoxStrut#1\)}%
    \endgroup
}
\newcommand{\figPooledKeyMath}[1]{%
    \begingroup
    \setlength{\fboxsep}{0.5pt}%
    \colorbox{figpooledkeyblue!75}{\(\color{black}\figMathBoxStrut#1\)}%
    \endgroup
}
\newcommand{\figKeyMath}[1]{%
    \begingroup
    \setlength{\fboxsep}{0.5pt}%
    \colorbox{figkeyblue!60}{\(\color{black}\figMathBoxStrut#1\)}%
    \endgroup
}
\begin{equation}
    \boldsymbol{D}_i
    :=
    \underbrace{
        \sum\nolimits_{j \in \mathcal{U}_i}
        B\cdot\exp\!\left(
            \figQueryMath{\boldsymbol{Q}_i}
            \hspace{0.8pt}
            \figPooledKeyMath{\bar{\boldsymbol{K}}_j^{\top}}
        \right)
    }_{\text{Approx.}}
    +
    \underbrace{
        \sum\nolimits_{j \in \mathcal{S}_i}
        \operatorname{RowSum}\!\left(
            \exp\!\left(
                \figQueryMath{\boldsymbol{Q}_i}
                \hspace{0.8pt}
                \figKeyMath{\boldsymbol{K}_j^{\top}}
            \right)
        \right)
    }_{\text{Exact}}
    \;\in\mathbb{R}^{B\times1}.
    \label{eq:solattn_denominator}
\end{equation}
\begin{equation}
    \boldsymbol{N}_i
    :=
    \underbrace{
        \sum\nolimits_{j \in \mathcal{U}_i}
        \exp\!\left(
            \figQueryMath{\boldsymbol{Q}_i}
            \hspace{0.8pt}
            \figPooledKeyMath{\bar{\boldsymbol{K}}_j^{\top}}
        \right)\widehat{\boldsymbol{V}}_j
    }_{\text{Approx.}}
    +
    \underbrace{
        \sum\nolimits_{j \in \mathcal{S}_i}
        \exp\!\left(
            \figQueryMath{\boldsymbol{Q}_i}
            \hspace{0.8pt}
            \figKeyMath{\boldsymbol{K}_j^{\top}}
        \right)\boldsymbol{V}_j
    }_{\text{Exact}}
    \;\in\mathbb{R}^{B\times d}.
    \label{eq:solattn_numerator}
\end{equation}
Normalizing each row of $\boldsymbol{N}_i$ by the corresponding entry of
$\boldsymbol{D}_i$ yields the output block
$\boldsymbol{O}_i\in\mathbb{R}^{B\times d}$.

In practice, the per-block matrix--vector multiplication (GEMV) in the approximate terms of Eqs.~\eqref{eq:solattn_denominator} and \eqref{eq:solattn_numerator} can be batched over a chunk of $C$ pooled keys and evaluated as an efficient matrix--matrix multiplication (GEMM), producing a $B\times C$ token-to-block score tile. This tiled computation also exposes its direct computational overlap with routing: averaging each column over query tokens exactly recovers the chunk-wise routing scores in Eq.~\eqref{eq:stream_proxy}:
\begin{equation}
    \widetilde{\boldsymbol{S}}_i^{(t)}
    :=
    \boldsymbol{Q}_i(\bar{\boldsymbol{K}}^{(t)})^\top,
    \qquad
    \operatorname{Mean}\!\big(
        \widetilde{\boldsymbol{S}}_i^{(t)}
    \big)
    =
    \bar{\boldsymbol{Q}}_i(\bar{\boldsymbol{K}}^{(t)})^\top
    =
    \widehat{\boldsymbol{s}}_i^{(t)}.
    \label{eq:proxy_reformulation}
\end{equation}
As illustrated in Figure~\ref{fig:pipeline}, block-wise approximation and chunk-wise routing are naturally unified in the outer loop. For each pooled-key chunk, the column means of the token-to-block score tile determine block selection; unselected columns provide approximate correction, while selected blocks are dispatched to the exact inner loop. Reusing the same score tile in this way fuses routing, sparse computation, and approximate correction within a single online-softmax pass.

\newlength{\figglyphheight}
\DeclareRobustCommand{\figPooledKeyIcon}{%
    \begingroup
    \settoheight{\figglyphheight}{t}%
    \tikz[baseline=0pt,x=0.80ex,y=\figglyphheight,line width=0.6pt]{%
        \path[draw=black,fill=figpooledkeyblue] (0,0) rectangle (1,1);
    }%
    \endgroup
}
\DeclareRobustCommand{\figPooledKeyTileIcon}{%
    \begingroup
    \settoheight{\figglyphheight}{t}%
    \tikz[baseline=0pt,x=0.80ex,y=\figglyphheight,line width=0.6pt]{%
        \path[draw=black,fill=figpooledkeyblue] (0,0) rectangle (1,1);
        \path[draw=black,fill=figpooledkeyblue] (1,0) rectangle (2,1);
        \path[draw=black,fill=figpooledkeyblue] (2,0) rectangle (3,1);
    }%
    \endgroup
}
\DeclareRobustCommand{\figKeyIcon}{%
    \begingroup
    \settoheight{\figglyphheight}{t}%
    \tikz[baseline=0pt,x=0.80ex,y=\figglyphheight,line width=0.6pt]{%
        \path[draw=black,fill=figkeyblue] (0,0) rectangle (1,1);
        \path[draw=black,fill=figkeyblue] (1,0) rectangle (2,1);
        \path[draw=black,fill=figkeyblue] (2,0) rectangle (3,1);
        \path[draw=black,fill=figkeyblue] (3,0) rectangle (4,1);
    }%
    \endgroup
}
\DeclareRobustCommand{\figProxyScoreIcon}{%
    \tikz[baseline=0pt,x=1ex,y=1ex,line width=0.6pt]{%
        \path[draw=black,fill=nvidiagreen!78!white] (0,0) rectangle (1,1);
        \path[draw=black,fill=nvidiagreen!58!white] (1,0) rectangle (2,1);
        \path[draw=black,fill=nvgreen] (2,0) rectangle (3,1);
    }%
}

\begin{figure}[t]
    \centering
    \includegraphics[width=\linewidth]{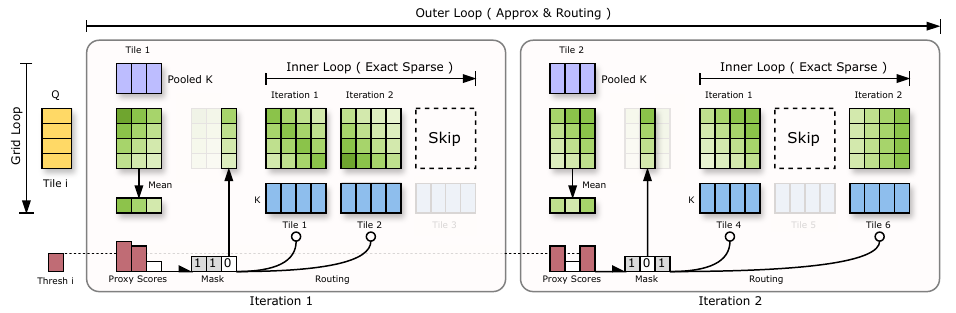}
    \captionsetup{skip=4pt}
    \caption[\ourmethod{} execution pipeline.]{\textbf{\ourmethod{} execution pipeline}.
    At each outer-loop iteration, a query block multiplies a pooled-key
    tile~\figPooledKeyTileIcon{} (each pooled key~\figPooledKeyIcon{} is pooled
    from an original key tile~\figKeyIcon{}) to produce token-to-block scores;
    averaging each column over query tokens yields the block-proxy
    scores~\figProxyScoreIcon{}.
    The selected blocks are dispatched to the exact inner loop, while the token-to-block scores are reused for approximation, with the selected columns masked to avoid double counting.}
    \vspace{-4pt}
    \label{fig:pipeline}
\end{figure}

\definecolor{algcomment}{rgb}{0,0.6,0}
\definecolor{algstring}{rgb}{0.58,0,0.82}
\definecolor{algfunction}{HTML}{2468A2}
\definecolor{algback}{HTML}{F3F3F0}
\newcommand{\solattnindentmark}[1]{%
    \tikz[remember picture,overlay]\coordinate (#1);%
}
\lstdefinestyle{solattnalgorithm}{
    language=Python,
    basicstyle=\ttfamily\fontsize{6.75}{9}\selectfont,
    keywordstyle=\color{magenta},
    commentstyle=\color{algcomment},
    stringstyle=\color{algstring},
    emph={QK_GEMM,Softmax,PV_GEMM},
    emphstyle=\color{algfunction},
    morekeywords={parallel},
    numbers=none,
    tabsize=4,
    keepspaces=true,
    columns=fullflexible,
    breaklines=true,
    escapeinside={(*@}{@*)},
    showstringspaces=false,
    frame=single,
    framerule=0pt,
    framesep=0pt,
    backgroundcolor=\color{algback},
    xleftmargin=5pt,
    xrightmargin=5pt,
    framexleftmargin=5pt,
    framexrightmargin=5pt,
    framextopmargin=0pt,
    framexbottommargin=0pt,
    aboveskip=0pt,
    belowskip=0pt
}

\newcounter{solalgorithm}
\vspace{-4pt}
\begin{wrapfigure}[14]{r}{0.45\textwidth}
    \vspace{-\intextsep}\vspace{-0.75pt}
    \refstepcounter{solalgorithm}\label{alg:solattn}%
    \textbf{Algorithm \thesolalgorithm: Sol-Attn Forward Pass}\par\vspace{0.25em}
\begin{lstlisting}[style=solattnalgorithm]
def sol_attn(Q, K, V, O, KC, VC, tau, N, B, C):
(*@\solattnindentmark{alg-g1-top}@*)    for i in range(N):
    (*@\solattnindentmark{alg-g2-top}@*)    Q_i = Q[i*B:(i+1)*B]
        acc, l_i, m_i = 0, 0, -inf
        # outer loop: routing and approximation
        for j in range(0, N, C):
        (*@\solattnindentmark{alg-g3-top}@*)    s_ij = QK_GEMM(Q_i, KC[j:j+C])
            mask = s_ij.mean(axis=0) > tau[i]
            s_ap = where(mask, -inf, s_ij)
            p_ap = Softmax(s_ap, acc, l_i, m_i)
            acc += PV_GEMM(p_ap, VC[j:j+C])
            # inner loop: exact sparse attention
            for t in nonzero(mask):
            (*@\solattnindentmark{alg-g4-top}@*)    s_ex = QK_GEMM(Q_i, K[(j+t)*B:(j+t+1)*B])
                p_ex = Softmax(s_ex, acc, l_i, m_i)
        (*@\solattnindentmark{alg-g3-bottom}@*)    (*@\solattnindentmark{alg-g4-bottom}@*)    acc += PV_GEMM(p_ex, V[(j+t)*B:(j+t+1)*B])
    (*@\solattnindentmark{alg-g2-bottom}@*)    O[i*B:(i+1)*B] = acc / l_i[:, None]
(*@\solattnindentmark{alg-g1-bottom}@*)    return O
\end{lstlisting}
    \begin{tikzpicture}[remember picture,overlay]
        \draw[black!20,line width=0.3pt]
            ([yshift=5pt]alg-g1-top) -- ([yshift=-3pt]alg-g1-bottom);
        \draw[black!20,line width=0.3pt]
            ([yshift=5pt]alg-g2-top) -- ([yshift=-3pt]alg-g2-bottom);
        \draw[black!20,line width=0.3pt]
            ([yshift=5pt]alg-g3-top) -- ([yshift=-3pt]alg-g3-bottom);
        \draw[black!20,line width=0.3pt]
            ([yshift=5pt]alg-g4-top) -- ([yshift=-3pt]alg-g4-bottom);
    \end{tikzpicture}
    \vspace{-\intextsep}
\end{wrapfigure}

\paragraph{Hardware-Aligned Implementation.}
We implement \ourmethod{} using the nested-loop kernel shown in
Figure~\ref{fig:pipeline} and summarized in
Algorithm~\ref{alg:solattn}. Here, \texttt{KC[j]} and \texttt{VC[j]} correspond respectively to 
the pooled key $\bar{\boldsymbol{K}}_j$ and summed value
$\widehat{\boldsymbol{V}}_j$ used by the approximate terms in
Eqs.~\eqref{eq:solattn_denominator} and \eqref{eq:solattn_numerator}.

Each kernel program stages $\boldsymbol{Q}_i$ and $\tau_i$ in shared memory
and maintains a single online-softmax state in registers. The outer loop
streams chunks of \texttt{KC} and \texttt{VC}, while the inner loop loads
the original $\langle\boldsymbol{K}_j,\boldsymbol{V}_j\rangle$ blocks only
for routed indices. Both the approximate and exact paths update the same
online-softmax statistics. Since routing is derived from the score tile already computed for approximation, its overhead is largely absorbed into the approximate online-softmax pipeline, without materializing proxy scores or routing indices in HBM.

\clearpage

%% file: sections/4_exps.tex
\section{Experiments}
\label{sec:exps}

\subsection{Experimental Setup}
\label{sec:exp_setup}

\paragraph{Models.}
We evaluate text-to-video generation with Wan2.1-14B~\cite{wan2025}, HunyuanVideo-13B~\cite{kong2024hunyuanvideo}, and LTX 2.3-22B~\cite{HaCohen2024LTXVideo,lightricks2026ltx23}, spanning increasingly long spatiotemporal sequences. Video-to-video evaluation covers editing with Bernini-14B~\cite{bernini} and refinement with SANA-WM Refiner~\cite{zhu2026sanawm}. We also evaluate native 2K text-to-image generation with Ideogram 4~\cite{ideogram-4-2026} to assess the effectiveness of \ourmethod{} at moderate sequence lengths.

\vspace{-4pt}
\paragraph{Metrics.}
We measure task quality using VBench~\cite{huang2024vbench} for text-to-video generation, pose accuracy for SANA-WM refinement, Bernini-Bench~\cite{bernini} for Bernini editing, and Qwen-Image-Bench~\cite{li2026qwenimagebench} for high-resolution text-to-image generation. Across tasks, PSNR, SSIM, and LPIPS quantify fidelity to dense attention outputs. Because theoretical FLOPs do not necessarily reflect practical efficiency, we report end-to-end wall-clock speedup relative to dense attention.

\vspace{-4pt}
\paragraph{Baselines.}
We use FlashAttention-3 (FA3)~\cite{shah2024flashattention3} as the dense attention baseline and compare with recent state-of-the-art training-free sparse attention methods: XAttention (XAttn)~\cite{xu2025xattention}, Sparse-VideoGen2 (SVG2)~\cite{yang2025sparse}, and Piecewise Sparse Attention (PISA)~\cite{li2026pisa}. For XAttn and SVG2, we adapt their officially recommended configurations to models not covered by the original releases, such as LTX, and match sparsity across methods for fair comparison.

\vspace{-4pt}
\paragraph{Implementation Details.}
We implement \ourmethod{} with custom GPU kernels and profile on NVIDIA H100 GPUs. Exact attention uses fixed $64{\times}64$ physical blocks. Since the routing and approximation chunk size \(C\) affects efficiency but not the mathematical output, we do not treat it as an ablation variable. Following prior work, for Wan2.1, HunyuanVideo, Bernini, and Ideogram 4, we use dense attention during the first 20\% of denoising steps and in the first layer as a warm-up. For the two-stage LTX pipeline, stage-1 uses an 8-step LoRA with dense attention throughout, whereas stage-2 uses \ourmethod{} for all steps without warm-up. The SANA-WM refiner likewise uses \ourmethod{} throughout.

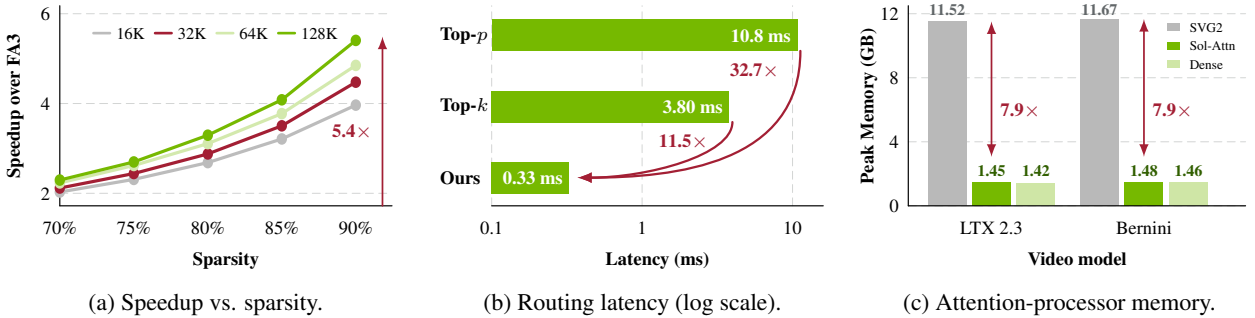
\begin{figure}[!b]
    \vspace{-16pt}
    \captionsetup{skip=8pt}
    \captionsetup[subfigure]{skip=6pt}
    \centering
    \begin{subfigure}[b]{0.32\linewidth}
        \centering
        \input{figures_tex/kernel_speedup_tikz}\par
        \caption{Speedup vs. sparsity.}
        \label{fig:kernel_speedup_scaling}
    \end{subfigure}\hfill
    \begin{subfigure}[b]{0.32\linewidth}
        \centering
        \input{figures_tex/routing_latency_tikz}\par
        \caption{Routing latency (log scale).%
        }
        \label{fig:routing_latency}
    \end{subfigure}\hfill
    \begin{subfigure}[b]{0.32\linewidth}
        \centering
        \input{figures_tex/routing_memory_tikz}\par
        \caption{Attention-processor memory.}
        \label{fig:routing_memory}
    \end{subfigure}
    \caption{Kernel efficiency profiled on an NVIDIA H100 GPU. (a) Speedup over FA3 increases with sparsity and sequence length. (b) Threshold routing is faster than top-$k$ and top-$p$. (c) Incremental peak memory of the attention processor. \ourmethod{} remains near Dense, while SVG2 requires about $8\times$ more memory on both workloads.}
    \label{fig:real_qkv_profile}
\end{figure}

\subsection{Kernel Efficiency Evaluation}
\label{sec:kernel_efficiency}

\paragraph{Kernel Efficiency Evaluation.}
Figure~\ref{fig:kernel_speedup_scaling} reports the kernel-level speedup of \ourmethod{} over FlashAttention-3 across different sequence lengths and sparsity levels. The speedup increases consistently with both sequence length and sparsity, reaching $5.41\times$ at 128K tokens and 90\% sparsity. This consistent scaling trend demonstrates that the efficiency advantage of \ourmethod{} persists throughout the 16K--128K sequence-length range.

\vspace{-4pt}
\paragraph{Efficient On-the-Fly Sparsification.}
Figures~\ref{fig:routing_latency} and~\ref{fig:routing_memory} isolate the two system costs of sparsification: selection latency and intermediate memory. Top-$k$ and top-$p$ materialize and globally traverse the proxy map, with top-$p$ further requiring cumulative selection over sorted scores. \ourmethod{} instead consumes each score on chip as it is produced, making routing $11.5\times$ and $32.7\times$ faster, respectively. Tile-local routing state also keeps \ourmethod{} near dense attention in processor memory, whereas SVG2's routing structures and permutation buffers require about $8\times$ more memory. The consistent separation on LTX 2.3 and Bernini attributes both gains to on-the-fly sparsification rather than model-specific integration.

\clearpage
\subsection{Visual Generation Evaluation}
\label{sec:generation_evaluation}

\input{tables/main_results}

\begin{table}[!t]
    \vspace{0.4cm}
    \centering

    \begin{minipage}[t]{0.49\linewidth}
        \vspace{0pt}
        \begin{minipage}[t][1.1cm][t]{\linewidth}
            \captionof{table}{Comparison of one-minute video refinement with the SANA-WM refiner across different attention methods.}
            \label{tab:sana_wm_results}
        \end{minipage}
        \input{tables/sana_wm}
    \end{minipage}\hfill
    \begin{minipage}[t]{0.49\linewidth}
        \vspace{0pt}
        \begin{minipage}[t][1.1cm][t]{\linewidth}
            \captionof{table}{Comparison of video-to-video editing with the Bernini-14B across different attention methods.}
            \label{tab:bernini_results}
        \end{minipage}
        \input{tables/bernini_v2v}
    \end{minipage}
\end{table}

\paragraph{Text-to-Video Generation.}
We evaluate text-to-video generation across three models, spanning short 720p videos to long 1080p videos. Table~\ref{tab:video_results} compares \ourmethod{} with recent training-free sparse attention methods in terms of VBench quality and similarity to dense-attention outputs, while speedup is measured end-to-end relative to dense attention with the FA3 backend. At matched sparsity, \ourmethod{} achieves state-of-the-art efficiency across all models, reaching up to $2.12\times$ end-to-end speedup while overall outperforming other methods in VBench quality and dense-reference similarity.

\paragraph{Video-to-Video Generation.}
We further evaluate \ourmethod{} on video-to-video generation, a more demanding
regime with substantially longer sequences. We consider two 720p settings:
refining a one-minute draft video with SANA-WM Refiner and editing videos with
Bernini. Tables~\ref{tab:sana_wm_results} and~\ref{tab:bernini_results} report
task-specific quality, dense-reference similarity, and end-to-end speedup at
matched sparsity. On SANA-WM, \ourmethod{} achieves the best pose accuracy and
dense-reference similarity among sparse methods while delivering a
$3.04\times$ speedup. On Bernini-Bench, it achieves the strongest aggregate
generation quality and dense-reference similarity among sparse methods,
together with the highest speedup.

\paragraph{Text-to-Image Generation.}
Beyond long-sequence video generation, we evaluate \ourmethod{} under moderate sequence lengths in native 2K text-to-image generation. Table~\ref{tab:mjhq_quality} shows that, on Ideogram 4, \ourmethod{} outperforms other sparse attention methods in both Qwen-Image-Bench scores and dense-reference similarity while achieving a $1.56\times$ speedup.

\paragraph{Video Generation on Consumer GPUs.}
We further evaluate \ourmethod{} on the consumer-grade NVIDIA RTX 5090 GPU. Figure~\ref{fig:gpu_portability} reports the kernel speedup over FlashAttention-4~\cite{zadouri2026flashattention}, while Table~\ref{tab:wan13b_5090_e2e} compares different sparse attention methods with Wan2.1-1.3B for 480p video generation. \ourmethod{} still achieves the best overall quality and efficiency.

\FloatBarrier
\begin{table}[H]
    \centering
    \input{tables/ideogram}\hfill
    \begin{minipage}[t]{0.335\linewidth}
        \vspace{0pt}
        \begin{minipage}[t][1.15cm][t]{\linewidth}
            \captionof{table}{Comparison of Wan2.1-1.3B 480p video generation on the RTX 5090.}
            \label{tab:wan13b_5090_e2e}
        \end{minipage}  
        \centering
        \scriptsize
        \setlength{\tabcolsep}{0.8pt}
        \renewcommand{\arraystretch}{1.0}
        \newcommand{\tabfivehighlight}{%
            \makebox[0pt][l]{\hspace{-\tabcolsep}\color{nvgreen}%
                \rule[-\dp\strutbox]{1000pt}{\dimexpr\ht\strutbox+\dp\strutbox\relax}}}
        \begin{adjustbox}{trim=0pt 0pt 0pt 0pt,clip}
            \begin{tabular*}{\linewidth}{@{\hspace{3pt}\extracolsep{\fill}}lccccc@{}}
                \toprule
                \multirow{2}{*}{\textbf{Method}}
                & \multicolumn{3}{c}{\textbf{Similarity}}
                & \multicolumn{2}{c}{\textbf{Efficiency}} \\
                \cmidrule(r{0.25em}){2-4}\cmidrule(l{0.25em}){5-6}
                & \textbf{PSNR} & \textbf{SSIM} & \textbf{LPIPS}
                & \textbf{Sparsity} & \textbf{Speedup} \\
                \midrule
                FA4 & -- & -- & -- & -- & 1.00$\times$ \\
                XAttn & 15.16 & 0.41 & 0.554 & 86.09\% & 1.36$\times$ \\
                SVG2 & 24.08 & 0.77 & 0.195 & 86.00\% & 1.41$\times$ \\
                \tabfivehighlight \ourmethod{} & 24.70 & 0.794 & 0.162 & 84.90\% & 1.71$\times$ \\
                \bottomrule
            \end{tabular*}
        \end{adjustbox}
    \end{minipage}
\end{table}

\FloatBarrier
\subsection{Integration with Sol-Engine}
\label{sec:sol_engine_integration}

Since \ourmethod{} sparsifies attention itself, it is compatible with complementary training-free acceleration techniques, including diffusion-step caching~\cite{zhou2025easycache,liu2024timestep,liu2025taylorseer} and kernel fusion. We therefore integrate it into the Sol-Engine~\cite{li2026sol}, a full-stack inference framework for composing such techniques, and evaluate the resulting system on NVIDIA B200 GPUs. Figure~\ref{fig:orthogonality_latency} shows that the combined system achieves end-to-end speedups of $3.48\times$ on Wan2.1-14B and $5.08\times$ on HunyuanVideo, demonstrating that \ourmethod{} is a practical drop-in component for optimized inference engines.

\begin{figure}[H]
    \vspace{-8pt}
    \centering
    \begin{subfigure}[b]{0.32\linewidth}
        \centering
        \input{figures_tex/rtx5090_kernel_speedup_tikz}\par
    \end{subfigure}\hfill
    \begin{subfigure}[b]{0.32\linewidth}
        \centering
        \input{figures_tex/sol_engine_hunyuan_latency_tikz}\par
    \end{subfigure}\hfill
    \begin{subfigure}[b]{0.32\linewidth}
        \centering
        \input{figures_tex/sol_engine_wan_latency_tikz}\par
    \end{subfigure}
    \addtocounter{figure}{-1}
    \par
    \begin{minipage}[t]{0.32\linewidth}
        \captionsetup{skip=0pt}
        \captionof{figure}{Sparsity versus speedup over FA4 on an NVIDIA RTX 5090.}
        \label{fig:gpu_portability}
    \end{minipage}\hfill
    \begin{minipage}[t]{0.66\linewidth}
        \captionsetup{skip=-4pt}
        \captionof{figure}{Integrating \ourmethod{} into Sol-Engine. End-to-end latency as acceleration techniques are progressively incorporated.}
        \label{fig:orthogonality_latency}
    \end{minipage}
\end{figure}

\subsection{Ablation Study}
\label{sec:ablation}

\begin{figure}[!b]
    \vspace{-8pt}
    \centering
    \captionsetup[subfigure]{skip=2pt}
    \begin{subfigure}[b]{0.32\linewidth}
        \centering
        \input{figures_tex/ablation_routing_budget_tikz}\par
    \end{subfigure}\hfill
    \begin{subfigure}[b]{0.32\linewidth}
        \centering
        \input{figures_tex/ablation_error_sparsity_tikz}\par
    \end{subfigure}\hfill
    \begin{subfigure}[b]{0.32\linewidth}
        \centering
        \input{figures_tex/ablation_error_length_tikz}\par
    \end{subfigure}
    \addtocounter{figure}{-1}
    \par
    \begin{minipage}[t]{0.32\linewidth}
        \captionsetup{skip=-4pt,justification=raggedright,singlelinecheck=false}
        \captionof{figure}{Routing-density distributions across different models.}
        \label{fig:routing_budget_ablation}
    \end{minipage}\hfill
    \begin{minipage}[t]{0.66\linewidth}
        \captionsetup{skip=-4pt,justification=raggedright,singlelinecheck=false}
        \captionof{figure}{Approximate-correction ablation at a sequence length of 32K across sparsity levels: relative $\ell_2$ error (left) and mean row cosine similarity (right).}
        \label{fig:core_ablation}
    \end{minipage}
\end{figure}

\paragraph{Effect of Query-Dependent Thresholding.}
Figure~\ref{fig:routing_budget_ablation} compares the per-query-block density distributions of different routing strategies at the same mean density of 15\%. Top-$k$ fixes every query block at the target density by construction, whereas Top-$p$ produces broadly varying densities. In contrast, our method remains tightly concentrated across models, demonstrating that it enables dynamic block budgets while maintaining stable density control.

\vspace{-4pt}
\paragraph{Effect of Approximate Correction.}
Figure~\ref{fig:core_ablation} evaluates approximate correction by comparing standard sparse attention (exact-only) and \ourmethod{} (exact-or-approx) under identical selected block indices as sparsity increases, measuring relative $\ell_2$ error and cosine similarity against dense attention outputs. Standard sparse attention accumulates rapidly growing error, whereas approximate correction consistently reduces relative $\ell_2$ error and preserves higher cosine similarity, with a widening advantage at higher sparsity. This confirms that retaining an approximate contribution from unselected blocks substantially improves the accuracy of sparse attention.

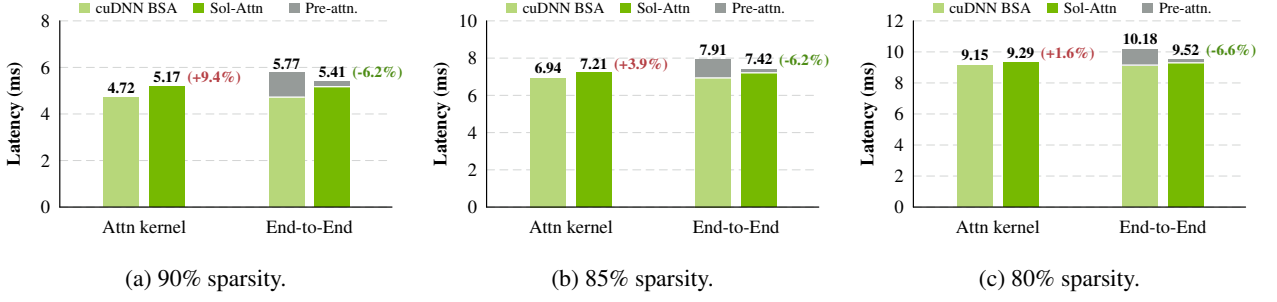
\begin{figure}[!t]
    \centering
    \begin{subfigure}[b]{0.32\linewidth}
        \centering
        {\def\smPanelDensity{10}\input{figures_tex/ablation_sm90_oracle_gap_tikz}}\par\vspace{-8pt}
        \caption{90\% sparsity.}
    \end{subfigure}\hfill
    \begin{subfigure}[b]{0.32\linewidth}
        \centering
        {\def\smPanelDensity{15}\input{figures_tex/ablation_sm90_oracle_gap_tikz}}\par\vspace{-8pt}
        \caption{85\% sparsity.}
    \end{subfigure}\hfill
    \begin{subfigure}[b]{0.32\linewidth}
        \centering
        {\def\smPanelDensity{20}\input{figures_tex/ablation_sm90_oracle_gap_tikz}}\par\vspace{-8pt}
        \caption{80\% sparsity.}
    \end{subfigure}
    \caption{Latency breakdown at a sequence length of 32K. Attn Kernel measures the attention operator alone, whereas End-to-End measures the full QKV-to-output path. The approximate correction in \ourmethod{} introduces only minor kernel overhead. At the end-to-end level, \ourmethod{} is faster because it computes only lightweight thresholds and hides routing overhead within the online softmax pipeline, whereas BSA materializes routing indices externally.}
    \label{fig:sm90_bsa_latency}
\end{figure}

\paragraph{Efficient Unified Execution Pipeline.}
To break down the latency of our unified kernel, we compare \ourmethod{} against cuDNN BSA, the state-of-the-art block-sparse attention implementation on H100 GPUs, using identical block indices at 10\%, 15\%, and 20\% exact density. Since \ourmethod{} fuses routing and approximate correction into the online-softmax pipeline of sparse attention, its kernel performs more work than BSA. Nevertheless, Figure~\ref{fig:sm90_bsa_latency} shows only 9.4\%, 3.9\%, and 1.6\% kernel-level overhead, respectively, confirming that approximate correction is lightweight. The comparison reverses end to end once BSA's separate routing stage is included: \ourmethod{} is 6.2\%, 6.2\%, and 6.6\% faster because only threshold computation remains outside its attention kernel, while the routing work is largely hidden by other operations in the online-softmax pipeline. Unified execution thus makes on-the-fly routing nearly free in the end-to-end pipeline.

%% file: figures_tex/kernel_speedup_tikz.tex
\begingroup
\usetikzlibrary{plotmarks}
\definecolor{solGrid}{HTML}{D3D5D4}
\definecolor{solMuted}{HTML}{BCBCBC}
\definecolor{solRed}{HTML}{A31F34}
\definecolor{solDarkGreen}{HTML}{315F00}
\definecolor{solGreen}{HTML}{76B900}
\begin{tikzpicture}[x=0.84cm,y=0.98cm,baseline=0pt,line cap=round,line join=round,>=latex]
  \def\axisbaseline{0.92}
  \def\xtickbaseline{0.54}
  \def\xlabelbaseline{0.12}
  \def\plotleft{0.82}
  \def\plotright{6.05}
  \def\plottop{3.62}
  \def\legendy{3.25}
  \path[use as bounding box] (0,0) rectangle (6.25,4.20);
  \begin{pgfinterruptboundingbox}

  \draw[solGrid,densely dashed,line width=0.25pt] (\plotleft,1.088750) -- (\plotright,1.088750);
  \node[anchor=east,font=\scriptsize,inner sep=1pt] at (0.72,1.088750) {2};
  \draw[solGrid,densely dashed,line width=0.25pt] (\plotleft,2.300375) -- (\plotright,2.300375);
  \node[anchor=east,font=\scriptsize,inner sep=1pt] at (0.72,2.300375) {4};
  \draw[solGrid,densely dashed,line width=0.25pt] (\plotleft,3.512000) -- (\plotright,3.512000);
  \node[anchor=east,font=\scriptsize,inner sep=1pt] at (0.72,3.512000) {6};

  \foreach \value/\x in {70\%/0.82,75\%/1.985,80\%/3.15,85\%/4.315,90\%/5.48}{
    \node[anchor=base,font=\scriptsize,inner sep=1pt] at (\x,\xtickbaseline) {\value};
  }

  \draw[line width=0.45pt] (\plotleft,\axisbaseline) -- (\plotright,\axisbaseline);
  \draw[line width=0.45pt] (\plotleft,\axisbaseline) -- (\plotleft,\plottop);

  \begin{scope}[cm={1,0,0,1.16862,(0,-0.592193)}]
  \foreach \x/\lo/\hi in {0.820000/1.452331/1.453795,1.985000/1.597584/1.601281,3.150000/1.790457/1.794923,4.315000/2.062062/2.066743,5.480000/2.453518/2.457573}{
    \draw[solMuted,line width=0.45pt,opacity=0.7] (\x,\lo) -- (\x,\hi);
    \draw[solMuted,line width=0.45pt,opacity=0.7] (\x,\lo) ++(-0.025,0) -- ++(0.05,0);
    \draw[solMuted,line width=0.45pt,opacity=0.7] (\x,\hi) ++(-0.025,0) -- ++(0.05,0);
  }
  \foreach \x/\lo/\hi in {0.820000/1.493682/1.503767,1.985000/1.664754/1.668970,3.150000/1.882129/1.896392,4.315000/2.213799/2.219410,5.480000/2.719901/2.723470}{
    \draw[solRed,line width=0.45pt,opacity=0.7] (\x,\lo) -- (\x,\hi);
    \draw[solRed,line width=0.45pt,opacity=0.7] (\x,\lo) ++(-0.025,0) -- ++(0.05,0);
    \draw[solRed,line width=0.45pt,opacity=0.7] (\x,\hi) ++(-0.025,0) -- ++(0.05,0);
  }
  \foreach \x/\lo/\hi in {0.820000/1.557652/1.565093,1.985000/1.755196/1.763355,3.150000/2.005040/2.017213,4.315000/2.355610/2.366880,5.480000/2.909400/2.923770}{
    \draw[nvgreen,line width=0.45pt,opacity=0.7] (\x,\lo) -- (\x,\hi);
    \draw[nvgreen,line width=0.45pt,opacity=0.7] (\x,\lo) ++(-0.025,0) -- ++(0.05,0);
    \draw[nvgreen,line width=0.45pt,opacity=0.7] (\x,\hi) ++(-0.025,0) -- ++(0.05,0);
  }
  \foreach \x/\lo/\hi in {0.820000/1.590991/1.591848,1.985000/1.799315/1.802085,3.150000/2.105130/2.112839,4.315000/2.505999/2.551020,5.480000/3.189996/3.230283}{
    \draw[nvidiagreen,line width=0.45pt,opacity=0.7] (\x,\lo) -- (\x,\hi);
    \draw[nvidiagreen,line width=0.45pt,opacity=0.7] (\x,\lo) ++(-0.025,0) -- ++(0.05,0);
    \draw[nvidiagreen,line width=0.45pt,opacity=0.7] (\x,\hi) ++(-0.025,0) -- ++(0.05,0);
  }

  \draw[solMuted,solid,line width=1.20pt,mark=*,mark size=1.35pt] plot coordinates {(0.820000,1.453192) (1.985000,1.598886) (3.150000,1.792862) (4.315000,2.066069) (5.480000,2.456526)};
  \draw[solRed,solid,line width=1.20pt,mark=*,mark size=1.35pt] plot coordinates {(0.820000,1.499158) (1.985000,1.666705) (3.150000,1.893044) (4.315000,2.217591) (5.480000,2.721628)};
  \draw[nvgreen,solid,line width=1.20pt,mark=*,mark size=1.35pt] plot coordinates {(0.820000,1.558115) (1.985000,1.759431) (3.150000,2.012128) (4.315000,2.358366) (5.480000,2.915278)};
  \draw[nvidiagreen,solid,line width=1.20pt,mark=*,mark size=1.35pt] plot coordinates {(0.820000,1.591498) (1.985000,1.800431) (3.150000,2.109305) (4.315000,2.518096) (5.480000,3.203981)};
  \end{scope}

  \draw[->,solRed,line width=0.65pt] (5.90,\axisbaseline) -- (5.90,3.204065);
  \node[anchor=east,font=\scriptsize\bfseries,text=solRed,fill=white,inner sep=0.8pt]
    at (5.78,1.916020) {5.4$\times$};

  \draw[solMuted,solid,line width=1.0pt,mark=*,mark size=1.15pt] (1.35,\legendy) -- (1.61,\legendy);
  \node[anchor=west,font=\fontsize{6.2}{6.8}\selectfont,inner sep=1pt] at (1.66,\legendy) {16K};
  \draw[solRed,solid,line width=1.0pt,mark=*,mark size=1.15pt] (2.32,\legendy) -- (2.58,\legendy);
  \node[anchor=west,font=\fontsize{6.2}{6.8}\selectfont,inner sep=1pt] at (2.63,\legendy) {32K};
  \draw[nvgreen,solid,line width=1.0pt,mark=*,mark size=1.15pt] (3.29,\legendy) -- (3.55,\legendy);
  \node[anchor=west,font=\fontsize{6.2}{6.8}\selectfont,inner sep=1pt] at (3.60,\legendy) {64K};
  \draw[nvidiagreen,solid,line width=1.0pt,mark=*,mark size=1.15pt] (4.26,\legendy) -- (4.52,\legendy);
  \node[anchor=west,font=\fontsize{6.2}{6.8}\selectfont,inner sep=1pt] at (4.57,\legendy) {128K};

  \node[anchor=base,font=\scriptsize\bfseries] at (3.435,\xlabelbaseline) {Sparsity};
  \node[font=\scriptsize\bfseries,rotate=90,inner sep=0pt] at (0.12,2.27) {Speedup over FA3};
  \end{pgfinterruptboundingbox}
\end{tikzpicture}
\endgroup

%% file: figures_tex/routing_latency_tikz.tex
\begingroup
\definecolor{solGrid}{HTML}{D3D5D4}
\definecolor{solRed}{HTML}{A31F34}
\definecolor{solGreen}{HTML}{76B900}
\begin{tikzpicture}[x=0.84cm,y=0.98cm,baseline=0pt,line cap=round,line join=round,>=latex]
  \def\axisbaseline{0.92}
  \def\xtickbaseline{0.54}
  \def\xlabelbaseline{0.12}
  \def\plotleft{0.90}
  \def\plotright{6.12}
  \def\ylabelleft{0.04}
  \def\threshbarmidline{1.30}
  \def\arrowendx{2.25}
  \def\topparrowstartx{5.76}
  \def\topparrowstarty{3.01}
  \def\topparrowcontroloney{1.48}
  \def\topparrowcontroltwox{3.78}
  \def\topkarrowstartx{4.69}
  \def\topkarrowstarty{2.05}
  \def\topkarrowcontroloney{1.60}
  \def\topkarrowcontroltwox{3.34}
  \path[use as bounding box] (0,0) rectangle (6.25,4.20);
  \begin{pgfinterruptboundingbox}

  \foreach \value/\x in {0.1/0.90,1/3.2683,10/5.6366}{
    \draw[solGrid,densely dashed,line width=0.25pt] (\x,\axisbaseline) -- (\x,3.62);
    \node[anchor=base,font=\scriptsize,inner sep=1pt] at (\x,\xtickbaseline) {\value};
  }
  \draw[line width=0.45pt] (\plotleft,\axisbaseline) -- (\plotright,\axisbaseline);

  \node[anchor=west,font=\scriptsize\bfseries,inner sep=1pt] at (\ylabelleft,3.22) {Top-$p$};
  \node[anchor=west,font=\scriptsize\bfseries,inner sep=1pt] at (\ylabelleft,2.26) {Top-$k$};
  \node[anchor=west,font=\scriptsize\bfseries,inner sep=1pt] at (\ylabelleft,1.30) {Ours};

  \fill[solGreen] (\plotleft,3.01) rectangle (5.7157,3.43);
  \fill[solGreen] (\plotleft,2.05) rectangle (4.6414,2.47);
  \fill[solGreen] (\plotleft,1.09) rectangle (2.1217,1.51);
  \node[anchor=east,font=\scriptsize\bfseries,text=white,inner sep=1pt] at (5.66,3.22) {10.8 ms};
  \node[anchor=east,font=\scriptsize\bfseries,text=white,inner sep=1pt] at (4.59,2.26) {3.80 ms};
  \node[anchor=east,font=\scriptsize\bfseries,text=white,inner sep=1pt] at (2.07,1.30) {0.33 ms};

  \draw[->,solRed,line width=0.85pt]
    (\topparrowstartx,\topparrowstarty) .. controls (\topparrowstartx,\topparrowcontroloney) and (\topparrowcontroltwox,\threshbarmidline) .. (\arrowendx,\threshbarmidline);
  \draw[->,solRed,line width=0.85pt]
    (\topkarrowstartx,\topkarrowstarty) .. controls (\topkarrowstartx,\topkarrowcontroloney) and (\topkarrowcontroltwox,\threshbarmidline) .. (\arrowendx,\threshbarmidline);
  \node[font=\scriptsize\bfseries,text=solRed,inner sep=1pt] at (5.04,2.74) {32.7$\times$};
  \node[font=\scriptsize\bfseries,text=solRed,inner sep=1pt] at (3.92,1.78) {11.5$\times$};

  \node[anchor=base,font=\scriptsize\bfseries] at (3.51,\xlabelbaseline) {Latency (ms)};
  \end{pgfinterruptboundingbox}
\end{tikzpicture}
\endgroup

%% file: figures_tex/routing_memory_tikz.tex
\begingroup
\definecolor{solGray}{HTML}{B9B9B9}
\definecolor{solMuted}{HTML}{626769}
\definecolor{solGrid}{HTML}{D3D5D4}
\definecolor{solRed}{HTML}{A31F34}
\definecolor{solGreen}{HTML}{76B900}
\definecolor{solDarkGreen}{HTML}{315F00}
\colorlet{solLightGreen}{solGreen!35}
\begin{tikzpicture}[x=0.84cm,y=0.98cm,baseline=0pt,line cap=round,line join=round,>=latex]
  \def\axisbaseline{0.92}
  \def\xtickbaseline{0.54}
  \def\xlabelbaseline{0.12}
  \def\plotleft{0.74}
  \def\plotright{6.05}
  \def\plottop{3.62}
  \path[use as bounding box] (0,0) rectangle (6.25,4.20);
  \begin{pgfinterruptboundingbox}

  \foreach \value/\y in {0/0.920000,4/1.784000,8/2.648000,12/3.512000}{
    \draw[solGrid,densely dashed,line width=0.25pt] (\plotleft,\y) -- (\plotright,\y);
    \node[anchor=east,font=\scriptsize,inner sep=1pt] at (0.64,\y) {\value};
  }

  \fill[solGray] (1.05,\axisbaseline) rectangle (1.65,3.409309);
  \fill[solGreen] (1.75,\axisbaseline) rectangle (2.35,1.233402);
  \fill[solLightGreen] (2.45,\axisbaseline) rectangle (3.05,1.225926);
  \node[font=\fontsize{6}{6.8}\selectfont\bfseries,text=solMuted] at (1.35,3.57) {11.52};
  \node[font=\fontsize{6}{6.8}\selectfont\bfseries,text=solDarkGreen] at (2.05,1.40) {1.45};
  \node[font=\fontsize{6}{6.8}\selectfont\bfseries,text=solDarkGreen] at (2.75,1.40) {1.42};

  \fill[solGray] (3.45,\axisbaseline) rectangle (4.05,3.441321);
  \fill[solGreen] (4.15,\axisbaseline) rectangle (4.75,1.239077);
  \fill[solLightGreen] (4.85,\axisbaseline) rectangle (5.45,1.236329);
  \node[font=\fontsize{6}{6.8}\selectfont\bfseries,text=solMuted] at (3.75,3.57) {11.67};
  \node[font=\fontsize{6}{6.8}\selectfont\bfseries,text=solDarkGreen] at (4.45,1.40) {1.48};
  \node[font=\fontsize{6}{6.8}\selectfont\bfseries,text=solDarkGreen] at (5.15,1.40) {1.46};

  \draw[<->,solRed,line width=0.65pt] (2.05,1.58) -- (2.05,3.409309);
  \node[anchor=west,font=\scriptsize\bfseries,text=solRed,fill=white,inner sep=0.8pt]
    at (2.15,2.20) {7.9$\times$};
  \draw[<->,solRed,line width=0.65pt] (4.45,1.58) -- (4.45,3.441321);
  \node[anchor=west,font=\scriptsize\bfseries,text=solRed,fill=white,inner sep=0.8pt]
    at (4.55,2.20) {7.9$\times$};

  \fill[solGray] (4.82,3.28) rectangle ++(0.20,0.10);
  \node[anchor=west,font=\fontsize{5}{5.8}\selectfont,inner sep=1pt]
    at (5.12,3.33) {SVG2};
  \fill[solGreen] (4.82,3.03) rectangle ++(0.20,0.10);
  \node[anchor=west,font=\fontsize{5}{5.8}\selectfont,inner sep=1pt]
    at (5.12,3.08) {Sol-Attn};
  \fill[solLightGreen] (4.82,2.78) rectangle ++(0.20,0.10);
  \node[anchor=west,font=\fontsize{5}{5.8}\selectfont,inner sep=1pt]
    at (5.12,2.83) {Dense};

  \draw[line width=0.45pt] (\plotleft,\axisbaseline) -- (\plotright,\axisbaseline);
  \draw[line width=0.45pt] (\plotleft,\axisbaseline) -- (\plotleft,\plottop);

  \node[anchor=base,font=\scriptsize,inner sep=1pt] at (2.05,\xtickbaseline) {LTX 2.3};
  \node[anchor=base,font=\scriptsize,inner sep=1pt] at (4.45,\xtickbaseline) {Bernini};
  \node[anchor=base,font=\scriptsize\bfseries] at (3.40,\xlabelbaseline) {Video model};
  \node[font=\scriptsize\bfseries,rotate=90,inner sep=0pt] at (0.12,2.27) {Peak Memory (GB)};
  \end{pgfinterruptboundingbox}
\end{tikzpicture}
\endgroup

%% file: tables/main_results.tex
\begin{table*}[!t]
\centering
\scriptsize
\setlength{\tabcolsep}{3.5pt}
\caption{Quantitative comparison of text-to-video quality and efficiency. LTX 2.3 uses a coarse-to-fine two-stage pipeline, with \ourmethod{} applied only to stage-2; outside (inside) values denote end-to-end (stage-2) speedups.}
\label{tab:video_results}
\begin{adjustbox}{trim=0pt 0pt 0pt 0pt,clip}
\begin{tabular*}{\linewidth}{@{\hspace{4pt}\extracolsep{\fill}}lccccccccc ccc cc}
\toprule
\multirow{2}{*}{\textbf{Method}}
& \multicolumn{9}{c}{\textbf{VBench}$\uparrow$}
& \multicolumn{3}{c}{\textbf{Similarity}}
& \multicolumn{2}{c}{\textbf{Efficiency}} \\
\cmidrule(r{0.25em}){2-10} \cmidrule(l{0.25em}r{0.25em}){11-13} \cmidrule(l{0.25em}){14-15}
& \textbf{SC} & \textbf{BC} & \textbf{TF} & \textbf{MS} & \textbf{AQ} & \textbf{IQ} & \textbf{DD} & \textbf{OC} & \textbf{AVG}
& \textbf{PSNR}$\uparrow$ & \textbf{SSIM}$\uparrow$ & \textbf{LPIPS}$\downarrow$
& \textbf{Sparsity} & \textbf{Speedup}$\uparrow$ \\
\midrule

\multicolumn{15}{c}{\textit{\textbf{Wan 2.1} (14B, Text-to-Video, 720p, 81 Frames)}} \\
\midrule
FA3   & 95.59 & 96.75 & 98.75 & 97.86 & 63.17 & 68.88 & 61.11 & 25.08 & 75.90 & -- & -- & -- & \hspace{\widthof{0}}-- & 1.00$\times$ \\
XAttn & 93.71 & 95.43 & 97.94 & 97.12 & 60.56 & 66.12 & 62.50 & 25.37 & 74.84 & 18.62 & 0.5544 & 0.3226 & 85.60\% & 1.69$\times$ \\
SVG2  & 94.85 & 96.05 & 98.59 & 97.68 & 62.26 & 67.40 & 59.72 & 25.19 & 75.22 & 20.71 & 0.6543 & 0.2498 & 83.66\% & 1.85$\times$ \\
PISA  & 95.60 & 96.95 & 98.75 & 97.98 & 63.72 & 68.91 & 61.11 & 25.19 & 76.03 & 20.28 & 0.6775 & 0.2127 & 85.00\% & 1.86$\times$ \\
\rowcolor{nvgreen}[8pt][8pt]
\ourmethod{}
& 95.54 & 97.02 & 98.76 & 98.01 & 63.35 & 68.64 & 62.50 & 25.26 & 76.13
& 20.59 & 0.6825 & 0.2086
& 85.12\% & 2.02$\times$ \\

\midrule

\multicolumn{15}{c}{\textit{\textbf{HunyuanVideo} (13B, Text-to-Video, 720p, 129 Frames)}} \\
\midrule
FA3   & 93.87 & 96.60 & 98.97 & 99.11 & 61.54 & 67.87 & 72.22 & 26.28 & 77.06 & -- & -- & -- & \hspace{\widthof{0}}-- & 1.00$\times$ \\
XAttn & 93.76 & 95.57 & 99.05 & 98.76 & 60.10 & 66.38 & 51.39 & 26.16 & 73.90 & 22.90 & 0.6792 & 0.2511 & 85.60\% & 1.61$\times$ \\
SVG2  & 92.22 & 95.63 & 99.01 & 98.95 & 58.01 & 61.81 & 63.89 & 26.08 & 74.45 & 25.64 & 0.7853 & 0.2354 & 83.15\% & 2.01$\times$ \\
PISA  & 93.68 & 96.37 & 98.95 & 99.08 & 61.69 & 67.71 & 72.22 & 26.49 & 77.02 & 25.58 & 0.8164 & 0.1671 & 85.00\% & 1.88$\times$ \\
\rowcolor{nvgreen}[8pt][8pt]
\ourmethod{}
& 93.62 & 96.49 & 98.95 & 99.05 & 61.52 & 67.55 & 70.83 & 26.50 & 76.81
& 25.21 & 0.8048 & 0.1777
& 85.80\% & 2.12$\times$ \\

\midrule

\multicolumn{15}{c}{\textit{\textbf{LTX 2.3} (22B, Text-to-Video, 1080p, 361 Frames $\rightarrow$ 721 Frames)}} \\
\midrule
FA3   & 90.78 & 95.53 & 99.31 & 99.33 & 62.59 & 69.89 & 56.94 & 22.38 & 74.59 & -- & -- & -- & \hspace{\widthof{0}}-- & 1.0$\times$ \\
XAttn & 90.04 & 95.17 & 98.86 & 98.99 & 62.56 & 70.21 & 58.33 & 22.55 & 74.59 & 21.77 & 0.7621 & 0.2175 & 90.17\% & 1.6 \textcolor{gray}{(1.9)}$\times$ \\
SVG2  & 90.70 & 95.46 & 99.28 & 99.06 & 59.74 & 64.86 & 44.44 & 22.28 & 71.98 & 21.94 & 0.7806 & 0.2874 & 89.35\% & 1.7 \textcolor{gray}{(2.1)}$\times$ \\
PISA  & 90.89 & 95.64 & 99.37 & 99.39 & 62.76 & 70.10 & 56.94 & 22.27 & 74.67 & 25.26 & 0.8630 & 0.1288 & 90.00\% & 1.8 \textcolor{gray}{(2.3)}$\times$ \\
\rowcolor{nvgreen}[8pt][8pt]
\ourmethod{}
& 90.85 & 95.67 & 99.36 & 99.39 & 62.92 & 70.16 & 56.94 & 22.24 & 74.69
& 25.69 & 0.8704 & 0.1197
& 89.83\% & 1.9 \textcolor{gray}{(2.4)}$\times$ \\

\bottomrule
\end{tabular*}
\end{adjustbox}
\end{table*}

%% file: tables/sana_wm.tex
\centering
\scriptsize
\setlength{\tabcolsep}{1.5pt}
\renewcommand{\arraystretch}{1.0}
\begin{adjustbox}{trim=0pt 0pt 0pt 0pt,clip}
\begin{tabular*}{\linewidth}{@{\hspace{4pt}\extracolsep{\fill}}lccc ccc cc@{}}
\toprule
\multirow{2}{*}{\textbf{Method}}
& \multicolumn{3}{c}{\textbf{Pose Acc.}$\downarrow$}
& \multicolumn{3}{c}{\textbf{Similarity}}
& \multicolumn{2}{c}{\textbf{Efficiency}} \\
\cmidrule(r{0.25em}){2-4} \cmidrule(l{0.25em}r{0.25em}){5-7} \cmidrule(l{0.25em}){8-9}
& \textbf{R.Err.} & \textbf{T.Err.} & \textbf{CMC}
& \textbf{PSNR}$\uparrow$ & \textbf{SSIM}$\uparrow$ & \textbf{LPIPS}$\downarrow$
& \textbf{Sparsity} & \textbf{Speedup} \\
\midrule
FA3 & 7.133 & 1.167 & 1.216 & -- & -- & -- & -- & 1.00$\times$ \\
XAttn & 10.52 & 1.331 & 1.415 & 16.80 & 0.466 & 0.411 & 85.30\% & 2.17$\times$ \\
SVG2  & 9.109 & 1.413 & 1.477 & 16.89 & 0.473 & 0.413 & 85.08\% & 2.08$\times$ \\
PISA  & 8.822 & 1.237 & 1.299 & 17.65 & 0.506 & 0.351 & 84.96\% & 2.35$\times$ \\
\rowcolor{nvgreen}[4pt][4pt]
\ourmethod{}  & 8.781 & 1.233 & 1.296 & 17.72 & 0.507 & 0.343 & 85.11\% & 3.04$\times$ \\
\bottomrule
\end{tabular*}
\end{adjustbox}

%% file: tables/bernini_v2v.tex
\centering
\scriptsize
\setlength{\tabcolsep}{1.5pt}
\renewcommand{\arraystretch}{1.0}
\begin{adjustbox}{trim=0pt 0pt 0pt 0pt,clip}
\begin{tabular*}{\linewidth}{@{\hspace{4pt}\extracolsep{\fill}}lcccc ccc cc@{}}
\toprule
\multirow{2}{*}{\textbf{Method}}
& \multicolumn{4}{c}{\textbf{Bernini-Bench}$\uparrow$}
& \multicolumn{3}{c}{\textbf{Similarity}}
& \multicolumn{2}{c}{\textbf{Efficiency}} \\
\cmidrule(r{0.25em}){2-5} \cmidrule(l{0.25em}r{0.25em}){6-8} \cmidrule(l{0.25em}){9-10}
& \textbf{IF} & \textbf{VC} & \textbf{GQ} & \textbf{OS}
& \textbf{PSNR}$\uparrow$ & \textbf{SSIM}$\uparrow$ & \textbf{LPIPS}$\downarrow$
& \textbf{Sparsity} & \textbf{Speedup} \\
\midrule
FA3
& 3.32 & 3.80 & 3.70 & 3.41
& -- & -- & --
& -- & 1.00$\times$ \\
XAttn
& 3.46 & 3.65 & 3.80 & 3.46
& 27.42 & 0.864 & 0.079
& 84.91\% & 1.64$\times$ \\
SVG2
& 3.41 & 3.62 & 3.62 & 3.36
& 27.28 & 0.863 & 0.074
& 81.15\% & 2.04$\times$ \\
PISA
& 3.36 & 3.75 & 3.75 & 3.55
& 29.88 & 0.904 & 0.052
& 85.00\% & 2.17$\times$ \\
\rowcolor{nvgreen}[4pt][4pt]
\ourmethod{}
& 3.50 & 3.80 & 3.80 & 3.50
& 30.18 & 0.910 & 0.046
& 85.00\% & 2.34$\times$ \\
\bottomrule
\end{tabular*}
\end{adjustbox}

%% file: tables/ideogram.tex
\begin{minipage}[t]{0.65\linewidth}
\vspace{0pt}
\begin{minipage}[t][1.15cm][t]{\linewidth}
\captionof{table}{Comparison of attention methods on Ideogram 4 for 2K-resolution text-to-image generation. BSA denotes standard block sparse attention.}
\label{tab:mjhq_quality}
\end{minipage}
\centering
\scriptsize
\setlength{\tabcolsep}{1.4pt}
\renewcommand{\arraystretch}{1.0}
\newcommand{\tabfourhighlight}{%
  \makebox[0pt][l]{\hspace{-\tabcolsep}\color{nvgreen}%
    \rule[-\dp\strutbox]{1000pt}{\dimexpr\ht\strutbox+\dp\strutbox\relax}}}
\begin{adjustbox}{trim=0pt 0pt 0pt 0pt,clip}
\begin{tabular*}{\linewidth}{@{\hspace{3pt}\extracolsep{\fill}}lccccccccccc@{}}
\toprule
\multirow{2}{*}{\textbf{Method}}
& \multicolumn{6}{c}{\textbf{Qwen-Image-Bench}$\uparrow$}
& \multicolumn{3}{c}{\textbf{Similarity}}
& \multicolumn{2}{c}{\textbf{Efficiency}} \\
\cmidrule(r{0.25em}){2-7}
\cmidrule(l{0.25em}r{0.25em}){8-10}
\cmidrule(l{0.25em}){11-12}
& \textbf{Qual.} & \textbf{Aesth.} & \textbf{Align.}
& \textbf{Fidel.} & \textbf{Creat.} & \textbf{AVG}
& \textbf{PSNR} & \textbf{SSIM} & \textbf{LPIPS}
& \textbf{Sparsity} & \textbf{Speedup}\\
\midrule
FA3
& 56.10 & 60.88 & 59.74 & 56.40 & 64.95 & 59.24
& -- & -- & -- & -- & 1.00$\times$ \\
BSA
& 50.04 & 52.28 & 52.63 & 51.40 & 53.08 & 51.96
& 15.99 & 0.565 & 0.427 & 90.00\% & 1.54$\times$ \\
PISA
& 52.78 & 55.56 & 55.09 & 53.07 & 55.85 & 54.51
& 20.12 & 0.738 & 0.239 & 90.00\% & 1.47$\times$ \\
\tabfourhighlight \ourmethod{}
& 52.77 & 56.12 & 56.61 & 53.33 & 57.71 & 55.31
& 21.26 & 0.764 & 0.210 & 90.16\% & 1.56$\times$ \\
\bottomrule
\end{tabular*}
\end{adjustbox}
\end{minipage}

%% file: figures_tex/rtx5090_kernel_speedup_tikz.tex
\begingroup
\usetikzlibrary{plotmarks}
\definecolor{solGrid}{HTML}{D3D5D4}
\definecolor{solMuted}{HTML}{BCBCBC}
\definecolor{solRed}{HTML}{A31F34}
\definecolor{solDarkGreen}{HTML}{315F00}
\definecolor{solGreen}{HTML}{76B900}
\begin{tikzpicture}[x=0.84cm,y=0.912cm,baseline=0pt,line cap=round,line join=round]
  \def\axisbaseline{0.92}
  \def\xtickbaseline{0.54}
  \def\xlabelbaseline{0.12}
  \def\plotleft{0.82}
  \def\plotright{6.05}
  \def\plottop{3.62}
  \def\legendy{3.366875}
  \path[use as bounding box] (0,0) rectangle (6.25,4.20);
  \begin{pgfinterruptboundingbox}

  \foreach \value/\y in {4/1.08875,6/1.76375,8/2.43875,10/3.11375}{
    \draw[solGrid,densely dashed,line width=0.25pt] (\plotleft,\y) -- (\plotright,\y);
    \node[anchor=east,font=\scriptsize,inner sep=1pt] at (0.72,\y) {\value};
  }

  \foreach \value/\x in {70\%/0.82,75\%/2.1275,80\%/3.435,85\%/4.7425,90\%/6.05}{
    \node[anchor=base,font=\scriptsize,inner sep=1pt] at (\x,\xtickbaseline) {\value};
  }

  \draw[line width=0.45pt] (\plotleft,\axisbaseline) -- (\plotright,\axisbaseline);
  \draw[line width=0.45pt] (\plotleft,\axisbaseline) -- (\plotleft,\plottop);

  \draw[solMuted,solid,line width=1.20pt,mark=*,mark size=1.35pt]
    plot coordinates {(0.820000,1.022225) (2.127500,1.192052) (3.435000,1.415475) (4.742500,1.712704) (6.050000,2.129096)};
  \draw[solRed,solid,line width=1.20pt,mark=*,mark size=1.35pt]
    plot coordinates {(0.820000,1.104275) (2.127500,1.307112) (3.435000,1.628750) (4.742500,2.064463) (6.050000,2.787725)};
  \draw[nvgreen,solid,line width=1.20pt,mark=*,mark size=1.35pt]
    plot coordinates {(0.820000,1.123513) (2.127500,1.359425) (3.435000,1.659125) (4.742500,2.205538) (6.050000,3.075613)};
  \draw[nvidiagreen,solid,line width=1.20pt,mark=*,mark size=1.35pt]
    plot coordinates {(0.820000,1.148150) (2.127500,1.397225) (3.435000,1.762063) (4.742500,2.320950) (6.050000,3.307475)};
  \draw[solMuted,solid,line width=1.0pt,mark=*,mark size=1.15pt] (1.04,\legendy) -- (1.32,\legendy);
  \node[anchor=west,font=\fontsize{6.2}{6.8}\selectfont,inner sep=1pt] at (1.38,\legendy) {8K};
  \draw[solRed,solid,line width=1.0pt,mark=*,mark size=1.15pt] (1.92,\legendy) -- (2.20,\legendy);
  \node[anchor=west,font=\fontsize{6.2}{6.8}\selectfont,inner sep=1pt] at (2.26,\legendy) {16K};
  \draw[nvgreen,solid,line width=1.0pt,mark=*,mark size=1.15pt] (2.94,\legendy) -- (3.22,\legendy);
  \node[anchor=west,font=\fontsize{6.2}{6.8}\selectfont,inner sep=1pt] at (3.28,\legendy) {32K};
  \draw[nvidiagreen,solid,line width=1.0pt,mark=*,mark size=1.15pt] (3.96,\legendy) -- (4.24,\legendy);
  \node[anchor=west,font=\fontsize{6.2}{6.8}\selectfont,inner sep=1pt] at (4.30,\legendy) {64K};

  \node[anchor=base,font=\scriptsize\bfseries] at (3.435,\xlabelbaseline) {Sparsity};
  \node[font=\scriptsize\bfseries,rotate=90,inner sep=0pt] at (0.12,2.27) {Speedup over FA4};
  \end{pgfinterruptboundingbox}
\end{tikzpicture}
\endgroup

%% file: figures_tex/sol_engine_hunyuan_latency_tikz.tex
\begingroup
\definecolor{solGray}{HTML}{B9B9B9}
\definecolor{solMuted}{HTML}{626769}
\definecolor{solGrid}{HTML}{D3D5D4}
\definecolor{solRed}{HTML}{A31F34}
\definecolor{solGreen}{HTML}{76B900}
\definecolor{solGreenTwo}{HTML}{8CCB37}
\definecolor{solGreenThree}{HTML}{B7D77A}
\begin{tikzpicture}[x=0.84cm,y=0.912cm,baseline=0pt,line cap=round,line join=round,>=latex]
    \def\axisbaseline{0.92}
    \def\xtickbaseline{0.54}
    \def\xlabelbaseline{0.12}
    \def\plotleft{0.35}
    \def\plotright{6.12}
    \def\plottop{3.62}
    \path[use as bounding box] (0,0) rectangle (6.25,4.20);
    \begin{pgfinterruptboundingbox}
    \foreach \value/\x in {0/0.35,300/2.2733,600/4.1967,900/6.12}{
        \draw[solGrid,densely dashed,line width=0.25pt] (\x,\axisbaseline) -- (\x,\plottop);
        \node[anchor=base,font=\scriptsize,inner sep=1pt] at (\x,\xtickbaseline) {\value};
    }
    \draw[line width=0.45pt] (\plotleft,\axisbaseline) -- (\plotright,\axisbaseline);
    \foreach \label/\y in {{+ Kernel Fusion \& Optimization}/2.94,{+ Diffusion Step Cache}/2.24,{+ Sol-Attn}/1.54}{
        \node[anchor=west,font=\fontsize{6}{6.6}\selectfont,inner sep=1pt] at (\plotleft,\y) {\label};
    }
    \fill[solGray]       (\plotleft,3.15) rectangle (5.9080,3.49);
    \fill[solGreenThree] (\plotleft,2.45) rectangle (4.9500,2.79);
    \fill[solGreenTwo]   (\plotleft,1.75) rectangle (2.4552,2.09);
    \fill[solGreen]      (\plotleft,1.05) rectangle (1.4438,1.39);

    \node[anchor=west,font=\fontsize{6}{6.6}\selectfont\bfseries,text=white,inner sep=1pt]
        at (0.55,3.32) {HunyuanVideo};
    \node[anchor=east,font=\fontsize{5.3}{6}\selectfont\bfseries,inner sep=1pt] at (5.86,3.32) {866.9 s};
    \node[anchor=east,font=\fontsize{5.3}{6}\selectfont\bfseries,inner sep=1pt] at (4.90,2.62) {781.0 s};
    \node[anchor=east,font=\fontsize{5.3}{6}\selectfont\bfseries,inner sep=1pt] at (2.41,1.92) {328.4 s};
    \node[anchor=east,font=\fontsize{5.3}{6}\selectfont\bfseries,inner sep=1pt] at (1.40,1.22) {170.6 s};

    \draw[->,solRed,line width=0.85pt]
        (5.91,3.15) .. controls (5.91,2.70) and (5.30,2.62) .. (5.04,2.62);
    \draw[->,solRed,line width=0.85pt]
        (5.91,3.15) .. controls (5.91,1.98) and (3.62,1.92) .. (2.55,1.92);
    \draw[->,solRed,line width=0.85pt]
        (5.91,3.15) .. controls (5.91,1.30) and (3.10,1.22) .. (1.54,1.22);
    \node[font=\scriptsize\bfseries,text=solRed,inner sep=1pt] at (5.17,2.91) {1.11$\times$};
    \node[font=\scriptsize\bfseries,text=solRed,inner sep=1pt] at (3.50,2.22) {2.64$\times$};
    \node[font=\scriptsize\bfseries,text=solRed,inner sep=1pt] at (3.05,1.52) {5.08$\times$};
    \node[anchor=base,font=\scriptsize\bfseries] at (3.24,\xlabelbaseline) {End-to-end latency (s)};
    \end{pgfinterruptboundingbox}
\end{tikzpicture}
\endgroup

%% file: figures_tex/sol_engine_wan_latency_tikz.tex
\begingroup
\definecolor{solGray}{HTML}{B9B9B9}
\definecolor{solMuted}{HTML}{626769}
\definecolor{solGrid}{HTML}{D3D5D4}
\definecolor{solRed}{HTML}{A31F34}
\definecolor{solGreen}{HTML}{76B900}
\definecolor{solGreenTwo}{HTML}{8CCB37}
\definecolor{solGreenThree}{HTML}{B7D77A}
\begin{tikzpicture}[x=0.84cm,y=0.912cm,baseline=0pt,line cap=round,line join=round,>=latex]
    \def\axisbaseline{0.92}
    \def\xtickbaseline{0.54}
    \def\xlabelbaseline{0.12}
    \def\plotleft{0.35}
    \def\plotright{6.12}
    \def\plottop{3.62}
    \path[use as bounding box] (0,0) rectangle (6.25,4.20);
    \begin{pgfinterruptboundingbox}
    \foreach \value/\x in {0/0.35,200/2.2733,400/4.1967,600/6.12}{
        \draw[solGrid,densely dashed,line width=0.25pt] (\x,\axisbaseline) -- (\x,\plottop);
        \node[anchor=base,font=\scriptsize,inner sep=1pt] at (\x,\xtickbaseline) {\value};
    }
    \draw[line width=0.45pt] (\plotleft,\axisbaseline) -- (\plotright,\axisbaseline);
    \foreach \label/\y in {{+ Kernel Fusion \& Optimization}/2.94,{+ Diffusion Step Cache}/2.24,{+ Sol-Attn}/1.54}{
        \node[anchor=west,font=\fontsize{6}{6.6}\selectfont,inner sep=1pt] at (\plotleft,\y) {\label};
    }
    \fill[solGray]       (\plotleft,3.15) rectangle (5.7719,3.49);
    \fill[solGreenThree] (\plotleft,2.45) rectangle (4.8208,2.79);
    \fill[solGreenTwo]   (\plotleft,1.75) rectangle (2.4426,2.09);
    \fill[solGreen]      (\plotleft,1.05) rectangle (1.9060,1.39);

    \node[anchor=west,font=\fontsize{6}{6.6}\selectfont\bfseries,text=white,inner sep=1pt]
        at (0.55,3.32) {Wan2.1-14B};
    \node[anchor=east,font=\fontsize{5.3}{6}\selectfont\bfseries,inner sep=1pt] at (5.73,3.32) {563.8 s};
    \node[anchor=east,font=\fontsize{5.3}{6}\selectfont\bfseries,inner sep=1pt] at (4.78,2.62) {464.9 s};
    \node[anchor=east,font=\fontsize{5.3}{6}\selectfont\bfseries,inner sep=1pt] at (2.40,1.92) {217.6 s};
    \node[anchor=east,font=\fontsize{5.3}{6}\selectfont\bfseries,inner sep=1pt] at (1.86,1.22) {161.8 s};

    \draw[->,solRed,line width=0.85pt]
        (5.78,3.15) .. controls (5.78,2.70) and (5.22,2.62) .. (4.91,2.62);
    \draw[->,solRed,line width=0.85pt]
        (5.78,3.15) .. controls (5.78,1.98) and (3.35,1.92) .. (2.54,1.92);
    \draw[->,solRed,line width=0.85pt]
        (5.78,3.15) .. controls (5.78,1.30) and (3.42,1.22) .. (2.01,1.22);
    \node[font=\scriptsize\bfseries,text=solRed,inner sep=1pt] at (5.08,2.91) {1.21$\times$};
    \node[font=\scriptsize\bfseries,text=solRed,inner sep=1pt] at (3.48,2.22) {2.59$\times$};
    \node[font=\scriptsize\bfseries,text=solRed,inner sep=1pt] at (3.10,1.52) {3.48$\times$};
    \node[anchor=base,font=\scriptsize\bfseries] at (3.24,\xlabelbaseline) {End-to-end latency (s)};
    \end{pgfinterruptboundingbox}
\end{tikzpicture}
\endgroup

%% file: figures_tex/ablation_routing_budget_tikz.tex
\begingroup
\definecolor{solGrid}{HTML}{D3D5D4}
\definecolor{solMuted}{HTML}{626769}
\definecolor{solRed}{HTML}{C06C75}
\definecolor{solGreenLight}{HTML}{B8DD72}
\definecolor{solGrayLight}{HTML}{C5C9C9}
\begin{tikzpicture}[x=0.84cm,y=0.912cm,baseline=0pt,line cap=round,line join=round]
  \def\axisbaseline{0.92}
  \def\xtickbaseline{0.54}
  \def\xlabelbaseline{0.12}
  \def\plotleft{0.82}
  \def\plotright{6.05}
  \def\plottop{3.62}
  \path[use as bounding box] (0,0) rectangle (6.25,4.20);
  \begin{pgfinterruptboundingbox}

  \foreach \value/\y in {5/1.0378,10/1.6269,15/2.2160,20/2.8051,25/3.3942}{
    \draw[solGrid,densely dashed,line width=0.25pt] (\plotleft,\y) -- (\plotright,\y);
    \node[anchor=east,font=\scriptsize,inner sep=1pt] at (0.72,\y) {\value};
  }
  \draw[black,densely dashed,line width=0.40pt,opacity=0.35] (\plotleft,2.2160) -- (\plotright,2.2160);
  \draw[line width=0.45pt] (\plotleft,\axisbaseline) -- (\plotright,\axisbaseline);
  \draw[line width=0.45pt] (\plotleft,\axisbaseline) -- (\plotleft,\plottop);

  \node[anchor=base,font=\scriptsize,inner sep=1pt] at (1.8070,\xtickbaseline) {Wan};
  \node[anchor=base,font=\scriptsize,inner sep=1pt] at (3.4765,\xtickbaseline) {HY};
  \node[anchor=base,font=\scriptsize,inner sep=1pt] at (5.1460,\xtickbaseline) {LTX};
  \filldraw[fill=solGrayLight,draw=black,line width=0.50pt] (1.4380,2.2145) circle[radius=1.55pt];
  \draw[black,line width=0.90pt] (1.8070,1.0224) -- (1.8070,1.3940);
  \draw[black,line width=0.90pt] (1.8070,2.9951) -- (1.8070,3.4416);
  \draw[black,line width=0.90pt] (1.7378,1.0224) -- (1.8761,1.0224);
  \draw[black,line width=0.90pt] (1.7378,3.4416) -- (1.8761,3.4416);
  \filldraw[fill=solRed,draw=black,line width=0.60pt] (1.6340,1.3940) rectangle (1.9799,2.9951);
  \draw[black,line width=0.90pt,line cap=butt] (1.6548,2.1696) -- (1.9592,2.1696);
  \draw[black,line width=0.90pt] (2.3489,1.9277) -- (2.3489,2.0848);
  \draw[black,line width=0.90pt] (2.3489,2.3168) -- (2.3489,2.6185);
  \draw[black,line width=0.90pt] (2.2797,1.9277) -- (2.4181,1.9277);
  \draw[black,line width=0.90pt] (2.2797,2.6185) -- (2.4181,2.6185);
  \filldraw[fill=solGreenLight,draw=black,line width=0.60pt] (2.1759,2.0848) rectangle (2.5218,2.3168);
  \draw[black,line width=0.90pt,line cap=butt] (2.1967,2.1946) -- (2.5011,2.1946);
  \filldraw[fill=solGrayLight,draw=black,line width=0.50pt] (3.1075,2.2135) circle[radius=1.55pt];
  \draw[black,line width=0.90pt] (3.4765,0.9534) -- (3.4765,1.4517);
  \draw[black,line width=0.90pt] (3.4765,2.8610) -- (3.4765,3.3704);
  \draw[black,line width=0.90pt] (3.4073,0.9534) -- (3.5457,0.9534);
  \draw[black,line width=0.90pt] (3.4073,3.3704) -- (3.5457,3.3704);
  \filldraw[fill=solRed,draw=black,line width=0.60pt] (3.3036,1.4517) rectangle (3.6495,2.8610);
  \draw[black,line width=0.90pt,line cap=butt] (3.3243,2.4325) -- (3.6287,2.4325);
  \draw[black,line width=0.90pt] (4.0184,1.7961) -- (4.0184,2.0214);
  \draw[black,line width=0.90pt] (4.0184,2.3611) -- (4.0184,2.7991);
  \draw[black,line width=0.90pt] (3.9492,1.7961) -- (4.0876,1.7961);
  \draw[black,line width=0.90pt] (3.9492,2.7991) -- (4.0876,2.7991);
  \filldraw[fill=solGreenLight,draw=black,line width=0.60pt] (3.8455,2.0214) rectangle (4.1914,2.3611);
  \draw[black,line width=0.90pt,line cap=butt] (3.8662,2.1436) -- (4.1706,2.1436);
  \filldraw[fill=solGrayLight,draw=black,line width=0.50pt] (4.7771,2.2154) circle[radius=1.55pt];
  \draw[black,line width=0.90pt] (5.1460,1.1158) -- (5.1460,1.3930);
  \draw[black,line width=0.90pt] (5.1460,3.0398) -- (5.1460,3.5120);
  \draw[black,line width=0.90pt] (5.0769,1.1158) -- (5.2152,1.1158);
  \draw[black,line width=0.90pt] (5.0769,3.5120) -- (5.2152,3.5120);
  \filldraw[fill=solRed,draw=black,line width=0.60pt] (4.9731,1.3930) rectangle (5.3190,3.0398);
  \draw[black,line width=0.90pt,line cap=butt] (4.9939,2.1281) -- (5.2982,2.1281);
  \draw[black,line width=0.90pt] (5.6880,1.9839) -- (5.6880,2.1250);
  \draw[black,line width=0.90pt] (5.6880,2.2966) -- (5.6880,2.5626);
  \draw[black,line width=0.90pt] (5.6188,1.9839) -- (5.7571,1.9839);
  \draw[black,line width=0.90pt] (5.6188,2.5626) -- (5.7571,2.5626);
  \filldraw[fill=solGreenLight,draw=black,line width=0.60pt] (5.5150,2.1250) rectangle (5.8609,2.2966);
  \draw[black,line width=0.90pt,line cap=butt] (5.5358,2.2012) -- (5.8402,2.2012);

  \def\legendbase{3.74}
  \def\legendtop{3.88}
  \def\legendcenter{3.81}
  \filldraw[fill=solGrayLight,draw=black,line width=0.45pt] (1.38,\legendcenter) circle[radius=1.35pt];
  \node[anchor=base west,font=\scriptsize,inner sep=0pt] at (1.50,\legendbase) {Top-$k$};
  \filldraw[fill=solRed,draw=black,line width=0.45pt] (2.65,\legendbase) rectangle (2.81,\legendtop);
  \node[anchor=base west,font=\scriptsize,inner sep=0pt] at (2.90,\legendbase) {Top-$p$};
  \filldraw[fill=solGreenLight,draw=black,line width=0.45pt] (4.05,\legendbase) rectangle (4.21,\legendtop);
  \node[anchor=base west,font=\scriptsize,inner sep=0pt] at (4.30,\legendbase) {Ours};

  \node[anchor=base,font=\scriptsize\bfseries] at (3.435,\xlabelbaseline) {Video model};
  \node[font=\scriptsize\bfseries,rotate=90,inner sep=0pt] at (0.12,2.27) {Per-query density (\%)};
  \end{pgfinterruptboundingbox}
\end{tikzpicture}
\endgroup

%% file: figures_tex/ablation_error_sparsity_tikz.tex
\begingroup
\usetikzlibrary{plotmarks}
\definecolor{solGrid}{HTML}{D3D5D4}
\definecolor{solRed}{HTML}{A31F34}
\definecolor{solGreen}{HTML}{76B900}
\begin{tikzpicture}[x=0.84cm,y=0.912cm,baseline=0pt,line cap=round,line join=round,>=latex]
  \def\axisbaseline{0.92}
  \def\xtickbaseline{0.54}
  \def\xlabelbaseline{0.12}
  \def\plotleft{1.12}
  \def\plotright{6.05}
  \def\plottop{3.62}
  \def\legendy{3.81}
  \path[use as bounding box] (0,0) rectangle (6.25,4.20);
  \begin{pgfinterruptboundingbox}

  \foreach \value/\label in {0.04/{.04},0.08/{.08},0.12/{.12},0.16/{.16}}{
    \pgfmathsetmacro{\y}{\axisbaseline + (\value-0.03)/0.15*(\plottop-\axisbaseline)}
    \draw[solGrid,densely dashed,line width=0.25pt] (\plotleft,\y) -- (\plotright,\y);
    \node[anchor=east,font=\scriptsize,inner sep=1pt] at (1.02,\y) {0\label};
  }
  \draw[line width=0.45pt] (\plotleft,\axisbaseline) -- (\plotright,\axisbaseline);
  \draw[line width=0.45pt] (\plotleft,\axisbaseline) -- (\plotleft,\plottop);

  \node[anchor=base,font=\scriptsize,inner sep=1pt] at (1.1200,\xtickbaseline) {70\%};
  \node[anchor=base,font=\scriptsize,inner sep=1pt] at (2.3525,\xtickbaseline) {75\%};
  \node[anchor=base,font=\scriptsize,inner sep=1pt] at (3.5850,\xtickbaseline) {80\%};
  \node[anchor=base,font=\scriptsize,inner sep=1pt] at (4.8175,\xtickbaseline) {85\%};
  \node[anchor=base,font=\scriptsize,inner sep=1pt] at (6.0500,\xtickbaseline) {90\%};
  \draw[solRed,line width=1.20pt,mark=*,mark size=1.35pt] plot coordinates {(1.1200,1.786891) (2.3525,2.026909) (3.5850,2.355399) (4.8175,2.773550) (6.0500,3.347197)};
  \draw[solGreen,densely dashed,line width=1.20pt,mark=square*,mark size=1.35pt] plot coordinates {(1.1200,1.072104) (2.3525,1.166225) (3.5850,1.291073) (4.8175,1.444616) (6.0500,1.675588)};

  \draw[solRed,line width=1.0pt,mark=*,mark size=1.15pt] (1.35,\legendy) -- (1.63,\legendy);
  \node[anchor=west,font=\fontsize{6.2}{6.8}\selectfont,inner sep=1pt] at (1.69,\legendy) {Exact-only};
  \draw[solGreen,densely dashed,line width=1.0pt,mark=square*,mark size=1.15pt] (3.22,\legendy) -- (3.50,\legendy);
  \node[anchor=west,font=\fontsize{6.2}{6.8}\selectfont,inner sep=1pt] at (3.56,\legendy) {Exact-or-approx};

  \node[anchor=base,font=\scriptsize\bfseries] at (3.585,\xlabelbaseline) {Sparsity};
  \node[font=\scriptsize\bfseries,rotate=90,inner sep=0pt] at (0.12,2.27)
    {Relative $\ell_2$ error};
  \end{pgfinterruptboundingbox}
\end{tikzpicture}
\endgroup

%% file: figures_tex/ablation_error_length_tikz.tex
\begingroup
\usetikzlibrary{plotmarks}
\definecolor{solGrid}{HTML}{D3D5D4}
\definecolor{solRed}{HTML}{A31F34}
\definecolor{solRedLight}{HTML}{E7C0C4}
\definecolor{solGreen}{HTML}{76B900}
\definecolor{solGreenLight}{HTML}{D8ECB5}
\begin{tikzpicture}[x=0.84cm,y=0.912cm,baseline=0pt,line cap=round,line join=round]
  \def\axisbaseline{0.92}
  \def\xtickbaseline{0.54}
  \def\xlabelbaseline{0.12}
  \def\plotleft{1.12}
  \def\plotright{6.05}
  \def\plottop{3.62}
  \def\legendy{3.81}
  \path[use as bounding box] (0,0) rectangle (6.25,4.20);
  \begin{pgfinterruptboundingbox}

  \foreach \value/\label in {0.88/{0.88},0.92/{0.92},0.96/{0.96},1.00/{1.00}}{
    \pgfmathsetmacro{\y}{\axisbaseline + (\value-0.88)/0.14*(\plottop-\axisbaseline)}
    \draw[solGrid,densely dashed,line width=0.25pt] (\plotleft,\y) -- (\plotright,\y);
    \node[anchor=east,font=\scriptsize,inner sep=1pt] at (1.02,\y) {\label};
  }
  \draw[line width=0.45pt] (\plotleft,\axisbaseline) -- (\plotright,\axisbaseline);
  \draw[line width=0.45pt] (\plotleft,\axisbaseline) -- (\plotleft,\plottop);

  \node[anchor=base,font=\scriptsize,inner sep=1pt] at (1.1200,\xtickbaseline) {70\%};
  \node[anchor=base,font=\scriptsize,inner sep=1pt] at (2.3525,\xtickbaseline) {75\%};
  \node[anchor=base,font=\scriptsize,inner sep=1pt] at (3.5850,\xtickbaseline) {80\%};
  \node[anchor=base,font=\scriptsize,inner sep=1pt] at (4.8175,\xtickbaseline) {85\%};
  \node[anchor=base,font=\scriptsize,inner sep=1pt] at (6.0500,\xtickbaseline) {90\%};
  \draw[solRed,line width=1.20pt,mark=*,mark size=1.35pt] plot coordinates {(1.1200,2.520619) (2.3525,2.311874) (3.5850,2.009209) (4.8175,1.609503) (6.0500,1.055025)};
  \draw[solGreen,densely dashed,line width=1.20pt,mark=square*,mark size=1.35pt] plot coordinates {(1.1200,3.025292) (2.3525,2.968383) (3.5850,2.889105) (4.8175,2.789638) (6.0500,2.653429)};

  \draw[solRed,line width=1.0pt,mark=*,mark size=1.15pt] (1.35,\legendy) -- (1.63,\legendy);
  \node[anchor=west,font=\fontsize{6.2}{6.8}\selectfont,inner sep=1pt] at (1.69,\legendy) {Exact-only};
  \draw[solGreen,densely dashed,line width=1.0pt,mark=square*,mark size=1.15pt] (3.22,\legendy) -- (3.50,\legendy);
  \node[anchor=west,font=\fontsize{6.2}{6.8}\selectfont,inner sep=1pt] at (3.56,\legendy) {Exact-or-approx};

  \node[anchor=base,font=\scriptsize\bfseries] at (3.585,\xlabelbaseline) {Sparsity};
  \node[font=\scriptsize\bfseries,rotate=90,inner sep=0pt] at (0.12,2.27)
    {cosine similarity};
  \end{pgfinterruptboundingbox}
\end{tikzpicture}
\endgroup

%% file: figures_tex/ablation_sm90_oracle_gap_tikz.tex
\begingroup
\definecolor{solGrid}{HTML}{D3D5D4}
\definecolor{bsaGreen}{HTML}{B7D77A}
\definecolor{solGreen}{HTML}{76B901}
\definecolor{preAttnGray}{HTML}{969B99}
\definecolor{latencyIncrease}{HTML}{B64045}
\definecolor{latencyDecrease}{HTML}{4E8D16}
\ifdefined\smPanelDensity\else
  \def\smPanelDensity{15}
\fi
\ifnum\smPanelDensity=10
  \def\smYMax{8}
  \def\smTicks{0,2,4,6,8}
  \def\bsaKernelY{2.51423}
  \def\solKernelY{2.66535}
  \def\bsaEteY{2.86681}
  \def\solEteY{2.74532}
  \def\bsaKernelLabelY{2.53}
  \def\solKernelLabelY{2.68}
  \def\bsaEteLabelY{2.88}
  \def\solEteLabelY{2.76}
  \def\bsaKernelValue{4.72}
  \def\solKernelValue{5.17}
  \def\bsaEteValue{5.77}
  \def\solEteValue{5.41}
  \def\solKernelDelta{+9.4\%}
  \def\solEteDelta{-6.2\%}
\else\ifnum\smPanelDensity=20
  \def\smYMax{12}
  \def\smTicks{0,2,4,6,8,10,12}
  \def\bsaKernelY{2.97850}
  \def\solKernelY{3.01129}
  \def\bsaEteY{3.21130}
  \def\solEteY{3.06162}
  \def\bsaKernelLabelY{3.00}
  \def\solKernelLabelY{3.03}
  \def\bsaEteLabelY{3.23}
  \def\solEteLabelY{3.08}
  \def\bsaKernelValue{9.15}
  \def\solKernelValue{9.29}
  \def\bsaEteValue{10.18}
  \def\solEteValue{9.52}
  \def\solKernelDelta{+1.6\%}
  \def\solEteDelta{-6.6\%}
\else
  \def\smYMax{10}
  \def\smTicks{0,2,4,6,8,10}
  \def\bsaKernelY{2.79279}
  \def\solKernelY{2.86652}
  \def\bsaEteY{3.05657}
  \def\solEteY{2.92437}
  \def\bsaKernelLabelY{2.81}
  \def\solKernelLabelY{2.88}
  \def\bsaEteLabelY{3.08}
  \def\solEteLabelY{2.94}
  \def\bsaKernelValue{6.94}
  \def\solKernelValue{7.21}
  \def\bsaEteValue{7.91}
  \def\solEteValue{7.42}
  \def\solKernelDelta{+3.9\%}
  \def\solEteDelta{-6.2\%}
\fi\fi
\begin{tikzpicture}[x=0.84cm,y=0.912cm,baseline=0pt,line cap=round,line join=round]
  \def\plotleft{0.82}
  \def\plotright{6.05}
  \def\axisbaseline{0.92}
  \def\plottop{3.62}
  \path[use as bounding box] (0,0) rectangle (6.25,4.20);
  \begin{pgfinterruptboundingbox}

  \foreach \value in \smTicks{
    \pgfmathsetmacro{\y}{\axisbaseline + \value/\smYMax*(\plottop-\axisbaseline)}
    \draw[solGrid,densely dashed,line width=0.25pt] (\plotleft,\y) -- (\plotright,\y);
    \node[anchor=east,font=\scriptsize,inner sep=1pt] at (0.72,\y) {\value};
  }

  \path[fill=bsaGreen]
    (1.51,\axisbaseline) rectangle (2.07,\bsaKernelY);
  \path[fill=solGreen]
    (2.23,\axisbaseline) rectangle (2.79,\solKernelY);

  \path[fill=bsaGreen]
    (4.11,\axisbaseline) rectangle (4.67,\bsaKernelY);
  \path[fill=preAttnGray]
    (4.11,\bsaKernelY) rectangle (4.67,\bsaEteY);
  \path[fill=solGreen]
    (4.83,\axisbaseline) rectangle (5.39,\solKernelY);
  \path[fill=preAttnGray]
    (4.83,\solKernelY) rectangle (5.39,\solEteY);
  \draw[white,line width=0.5pt] (4.11,\bsaKernelY) -- (4.67,\bsaKernelY);
  \draw[white,line width=0.5pt] (4.83,\solKernelY) -- (5.39,\solKernelY);

  \draw[line width=0.45pt] (\plotleft,\axisbaseline) -- (\plotright,\axisbaseline);
  \draw[line width=0.45pt] (\plotleft,\axisbaseline) -- (\plotleft,\plottop);

  \node[anchor=south,font=\bfseries\fontsize{5.8}{6.2}\selectfont,inner sep=1pt] at (1.79,\bsaKernelLabelY) {\bsaKernelValue};
  \node[anchor=south,font=\bfseries\fontsize{5.8}{6.2}\selectfont,inner sep=1pt] at (2.51,\solKernelLabelY) {\solKernelValue};
  \node[anchor=south west,font=\bfseries\fontsize{5.8}{6.2}\selectfont,inner sep=0pt,text=latencyIncrease]
    at (2.82,\solKernelLabelY) {(\solKernelDelta)};
  \node[anchor=south,font=\bfseries\fontsize{5.8}{6.2}\selectfont,inner sep=1pt] at (4.39,\bsaEteLabelY) {\bsaEteValue};
  \node[anchor=south,font=\bfseries\fontsize{5.8}{6.2}\selectfont,inner sep=1pt] at (5.11,\solEteLabelY) {\solEteValue};
  \node[anchor=south west,font=\bfseries\fontsize{5.8}{6.2}\selectfont,inner sep=0pt,text=latencyDecrease]
    at (5.42,\solEteLabelY) {(\solEteDelta)};

  \node[anchor=base,font=\scriptsize,inner sep=1pt] at (2.15,0.54) {Attn kernel};
  \node[anchor=base,font=\scriptsize,inner sep=1pt] at (4.75,0.54) {End-to-End};
  \node[font=\scriptsize\bfseries,rotate=90,inner sep=0pt] at (0.12,2.27) {Latency (ms)};

  \def\legendbase{3.74}
  \def\legendtop{3.88}
  \path[fill=bsaGreen]
    (1.13,\legendbase) rectangle (1.29,\legendtop);
  \node[anchor=base west,font=\fontsize{5.8}{6.2}\selectfont,inner sep=0pt]
    at (1.38,\legendbase) {cuDNN BSA};
  \path[fill=solGreen]
    (2.93,\legendbase) rectangle (3.09,\legendtop);
  \node[anchor=base west,font=\fontsize{5.8}{6.2}\selectfont,inner sep=0pt]
    at (3.18,\legendbase) {Sol-Attn};
  \path[fill=preAttnGray]
    (4.43,\legendbase) rectangle (4.59,\legendtop);
  \node[anchor=base west,font=\fontsize{5.8}{6.2}\selectfont,inner sep=0pt]
    at (4.68,\legendbase) {Pre-attn.};

  \end{pgfinterruptboundingbox}
\end{tikzpicture}
\endgroup

%% file: sections/5_related.tex
\section{Related Work}
\label{sec:related}

\paragraph{Block Sparse Attention.}
Block sparse attention reduces quadratic attention cost by computing only
selected key-value blocks. Early methods reduce this cost through structured
layouts~\cite{child2019sparse,beltagy2020longformer,zaheer2020bigbird} or
content-based grouping~\cite{kitaev2020reformer,roy2020routing}. More recent
work learns block selection or combines compression, selection, and local
attention within a native sparse architecture~\cite{gao2025seerattention,
yuan2025nsa,lu2025moba}. In video generation, existing methods exploit the
spatial--temporal redundancy of DiTs in two broad ways. One line of work
constructs structured sparse masks from recurring spatial or temporal attention
patterns~\cite{xi2025sparse,li2025radial,zhang2025fast,chen2025sparsevdit};
another derives content-adaptive sparse patterns at runtime from lightweight
block-level proxy scores~\cite{zhang2025vsa,yang2025sparse,hu2026dfsattn,liu2025fpsattention}, achieving more flexible block
selection.

\paragraph{Routing Strategies.}
Dynamic sparse attention hinges on selecting informative key-value blocks
efficiently. The most direct strategies apply top-$k$ or top-$p$ to block-proxy
scores. SpargeAttn augments top-$k$ selection with a confidence threshold and
then applies an online-softmax-aware threshold to skip negligible
updates~\cite{zhang2025spargeattn}. SpargeAttn2~\cite{zhang2026spargeattn2} takes the union of top-$k$
and top-$p$ masks to avoid their complementary failure
modes~\cite{zhang2026spargeattn2}, whereas Twilight~\cite{lin2025twilight} first forms a conservative
top-$k$ candidate set and then prunes it with
top-$p$~\cite{lin2025twilight}. Other methods redesign the importance measure or
filtering primitive: XAttention scores blocks by antidiagonal
sums~\cite{xu2025xattention}, while VecAttention retains scores within a fixed
margin of the running row maximum and fuses the comparison into tiled score
computation~\cite{liu2026vecattention}. Related systems reduce the search cost
through query-aware page selection~\cite{tang2024quest}, hierarchical
pruning~\cite{lee2025hip}, context-dependent sparse patterns~\cite{lai2025flexprefill},
or learned semantic hashing~\cite{desai2025hashattention}. \ourmethod{} instead
computes a query-dependent threshold for each proxy row and applies it to scores
as they are generated, enabling routing to proceed online within the
online-softmax pass.

\paragraph{Sparse Attention with Correction.}
Correction-based methods approximate contributions omitted by the exact sparse branch. PISA applies block-wise Taylor expansion to unselected blocks and combines exact and approximate terms within the same softmax numerator and denominator~\cite{li2026pisa}. SVG-EAR uses centroid-based linear compensation with error-aware routing~\cite{zhou2026svgear}, and BA-Att introduces covariance-compensated block approximation~\cite{zhang2026baatt}. SLA assigns critical weights to sparse attention and marginal weights to linear attention~\cite{zhang2025sla}; SLA2 adds learned routing, weighted sparse--linear composition, and quantization-aware training~\cite{zhang2026sla2}. Existing corrections either follow a sparse pattern constructed in a separate routing stage or require model fine-tuning. \ourmethod{} instead derives routing and correction from the same token-to-block score tile in a single online-softmax pass.

\section{Conclusion}
We presented \ourmethod{}, a training-free sparse attention that combines threshold routing and approximate correction in a single online-softmax pass without materializing a proxy map. More broadly, our study explores a new form of sparse attention in which routing is no longer a separate prerequisite and unselected blocks are approximated rather than discarded. This unified streaming design substantially narrows the accuracy gap between sparse and dense attention with little additional overhead, yielding a favorable quality--efficiency trade-off across diverse visual-generation tasks.

\paragraph{Limitations and Future Work.}
Despite these encouraging results, \ourmethod{} remains limited in both its current implementation and application scope. The B200 kernel does not yet fully exploit Blackwell's performance potential and currently supports only forward inference, while our evaluation is confined to bidirectional diffusion-based visual generation and does not cover autoregressive video generation. Future work will address these limitations through further kernel optimization, extension to autoregressive models, and a trainable implementation with backward support.

%% file: sections/x_appendix.tex
\clearpage
\onecolumn
\normalsize
\section{Additional Experimental Details}
\label{sec:additional_details}

\paragraph{Generation Configurations.}
Wan2.1-14B and HunyuanVideo-13B use 50 denoising steps, with dense
attention for the first 10 steps and \ourmethod{} thereafter. Bernini-14B
follows the same policy over 40 steps with an eight-step dense warm-up.
Wan2.1 and Bernini keep cross-attention dense, while HunyuanVideo's MMDiT
treats text key-value tokens as sinks and retains them in the exact branch.
Ideogram 4 uses eight dense steps in its 48-step schedule and likewise retains
text key-value tokens in the exact branch. For LTX 2.3, the eight-step first
stage remains dense, and \ourmethod{} is applied throughout the remaining
three steps while cross-attention stays dense. The three-step SANA-WM refiner
uses \ourmethod{} throughout and retains dense cross-attention.

\section{Derivations and Error Analysis}

\paragraph{Threshold variance derivation.}
\label{sec:threshold_variance_derivation}
The exact variance in Eq.~\eqref{eq:threshold_stats_exact} follows directly
from the variance of the proxy row. Recall from
Eq.~\eqref{eq:block_proxy} that
$\widehat{s}_{ij}=\bar{\boldsymbol{Q}}_i
\bar{\boldsymbol{K}}_j^{\top}$. The corresponding row mean is
$\mu_i=\frac{1}{N}\sum\nolimits_{j=1}^{N}\widehat{s}_{ij}$, its variance
expands as
\begin{align}
    \sigma_i^2
    &=
    \frac{1}{N}\sum\nolimits_{j=1}^{N}
    (\widehat{s}_{ij}-\mu_i)^2
    \nonumber\\
    &=
    \frac{1}{N}\sum\nolimits_{j=1}^{N}
    \left(\widehat{s}_{ij}^{\,2}
    -2\mu_i\widehat{s}_{ij}+\mu_i^2\right)
    \nonumber\\
    &=
    \frac{1}{N}\sum\nolimits_{j=1}^{N}
    \widehat{s}_{ij}^{\,2}
    -2\mu_i
    \left(\frac{1}{N}\sum\nolimits_{j=1}^{N}\widehat{s}_{ij}\right)
    +\mu_i^2
    \nonumber\\
    &=
    \frac{1}{N}\sum\nolimits_{j=1}^{N}
    \widehat{s}_{ij}^{\,2}-\mu_i^2.
    \label{eq:appendix_proxy_variance}
\end{align}
Each squared proxy score can in
turn be expressed as
\begin{equation}
    \widehat{s}_{ij}^{\,2}
    =
    \left(\bar{\boldsymbol{Q}}_i
    \bar{\boldsymbol{K}}_j^{\top}\right)
    \left(\bar{\boldsymbol{Q}}_i
    \bar{\boldsymbol{K}}_j^{\top}\right)^{\top}
    =
    \bar{\boldsymbol{Q}}_i
    \bar{\boldsymbol{K}}_j^{\top}\bar{\boldsymbol{K}}_j
    \bar{\boldsymbol{Q}}_i^{\top}.
    \label{eq:appendix_proxy_square}
\end{equation}
Since $\widehat{s}_{ij}$ is scalar, the second factor can be replaced by its
transpose to expose the pooled-key second moment. Substituting
Eq.~\eqref{eq:appendix_proxy_square} into
Eq.~\eqref{eq:appendix_proxy_variance} and collecting the key-dependent terms
then gives
\begin{equation}
    \sigma_i^2
    =
    \bar{\boldsymbol{Q}}_i
    \left(\frac{1}{N}\sum\nolimits_{j=1}^{N}
    \bar{\boldsymbol{K}}_j^{\top}\bar{\boldsymbol{K}}_j\right)
    \bar{\boldsymbol{Q}}_i^{\top}
    -\mu_i^2.
    \label{eq:appendix_threshold_moments}
\end{equation}
This recovers Eq.~\eqref{eq:threshold_stats_exact}.

\vspace{-3pt}
\paragraph{Diagonal threshold estimator.}
\label{sec:diagonal_threshold}
We provide implementations of both the exact estimator in
Eq.~\eqref{eq:threshold_stats_exact} and a more efficient approximation that
retains only the diagonal of the pooled-key covariance. Let
$\boldsymbol{\mu}_{K}:=\mathbb{E}[\bar{\boldsymbol{K}}]$ and retain only the
diagonal
$\boldsymbol{v}_{K}:=\operatorname{diag}(\operatorname{Cov}[\bar{\boldsymbol{K}}])$.
The resulting estimator is
\begin{equation}
    \boldsymbol{v}_{K}
    =
    \frac{1}{N}\sum\nolimits_{j=1}^{N}
    \bar{\boldsymbol{K}}_j\odot\bar{\boldsymbol{K}}_j
    -\boldsymbol{\mu}_{K}\odot\boldsymbol{\mu}_{K},
    \qquad
    \sigma_{i,\mathrm{diag}}^2
    =
    (\bar{\boldsymbol{Q}}_i\odot\bar{\boldsymbol{Q}}_i)
    \boldsymbol{v}_{K}^{\top}.
    \label{eq:appendix_diagonal_threshold}
\end{equation}
It replaces the dense $d\times d$ covariance with element-wise second moments
and an $O(d)$ projection per query block.

\vspace{-3pt}
\paragraph{Approximation error.}
The zeroth-order rule also gives a direct error characterization. For one query row, write $a=\boldsymbol{q}\bar{\boldsymbol{K}}_j^{\top}$, $\delta_u=\boldsymbol{q}(\boldsymbol{k}_{j,u}-\bar{\boldsymbol{K}}_j)^{\top}$, and $\eta=\max_u|\delta_u|$. Let $\mathcal{D}_j,\widetilde{\mathcal{D}}_j$ and $\mathcal{N}_j,\widetilde{\mathcal{N}}_j$ denote the exact and zeroth-order block denominator and numerator contributions for this row. Since $\sum_u\delta_u=0$, Taylor's theorem yields
\begin{equation}
    0\leq \mathcal{D}_j-\widetilde{\mathcal{D}}_j
    \leq \frac{1}{2}\exp(a+\eta)\sum_{u=1}^{B}\delta_u^2,
    \qquad
    \lVert\mathcal{N}_j-\widetilde{\mathcal{N}}_j\rVert_2
    \leq \exp(a+\eta)\sum_{u=1}^{B}|\delta_u|\lVert\boldsymbol{v}_{j,u}\rVert_2.
    \label{eq:appendix_approximation_error}
\end{equation}
Thus, the denominator error is second order in the centered score spread,
whereas the numerator also depends on its alignment with the values. These
bounds explain the accuracy on smooth tail blocks and the residual error on
heterogeneous ones.
The derivation assumes dense, noncausal keys within each block, as in the
evaluated diffusion models; row-dependent semantic or causal masks instead
require centering and multiplicity over valid keys.

\clearpage
\normalsize
\section{Additional Qualitative Results}
\label{sec:additional_qualitative}
\vspace{-4pt}

\newcommand{\suppqualframe}[1]{%
  \includegraphics[width=0.1625\textwidth,trim=0 45 0 45,clip]{#1}%
}
\newcommand{\suppquallabeledframe}[2]{%
  \begin{tikzpicture}[baseline=(frame.south)]
    \node[inner sep=0pt] (frame) {\suppqualframe{#1}};
    \node[
      anchor=south west,
      fill=black,
      fill opacity=0.72,
      text opacity=1,
      text=white,
      font=\scriptsize,
      inner xsep=3pt,
      inner ysep=1.2pt
    ] at (frame.south west) {#2};
  \end{tikzpicture}%
}
\newcommand{\suppqualrow}[3]{%
  \suppquallabeledframe{figures/suppmm_#1_#2_01.jpg}{#3} &
  \suppqualframe{figures/suppmm_#1_#2_02.jpg} &
  \suppqualframe{figures/suppmm_#1_#2_03.jpg} &
  \suppqualframe{figures/suppmm_#1_#2_04.jpg} &
  \suppqualframe{figures/suppmm_#1_#2_05.jpg} &
  \suppqualframe{figures/suppmm_#1_#2_06.jpg}\\[-2.5pt]
}
\newcommand{\suppqualmodelcaption}[1]{%
  \noalign{\vskip 2pt}%
  \multicolumn{6}{c}{\small #1}\\[2pt]
}
\newcommand{\suppqualprompt}[2]{%
  \multicolumn{6}{@{}l@{}}{%
    \parbox[t]{0.982\textwidth}{%
      \fontsize{6}{6.6}\selectfont\raggedright
      \textit{#1:} #2
    }%
  }\\[-1pt]
}

{\centering
\enlargethispage{24pt}
\setlength{\tabcolsep}{0pt}
\renewcommand{\arraystretch}{1}
\begin{tabular}{@{}c@{\hspace{1.2pt}}c@{\hspace{1.2pt}}c@{\hspace{1.2pt}}c@{\hspace{1.2pt}}c@{\hspace{1.2pt}}c@{}}
\suppqualprompt{Prompt}{a motorcycle cruising along a coastal highway}
\suppqualrow{wan_motorcycle}{dense}{Dense}
\suppqualrow{wan_motorcycle}{sol}{Sol-Attn}
\suppqualprompt{Prompt}{a train speeding down the tracks}
\suppqualrow{wan_train}{dense}{Dense}
\suppqualrow{wan_train}{sol}{Sol-Attn}
\suppqualmodelcaption{(a) Wan 2.1}
\suppqualprompt{Prompt}{A Mars rover moving on Mars}
\suppqualrow{hy_rover}{dense}{Dense}
\suppqualrow{hy_rover}{sol}{Sol-Attn}
\suppqualprompt{Prompt}{A steam train moving on a mountainside}
\suppqualrow{hy_steamtrain}{dense}{Dense}
\suppqualrow{hy_steamtrain}{sol}{Sol-Attn}
\suppqualmodelcaption{(b) HunyuanVideo}
\suppqualprompt{Prompt}{INT. QUIET APARTMENT BY TALL WINDOW - LATE AFTERNOON. A bright cinematic 85mm close-up frames a beautiful woman in a cream sweater beside white curtains and wooden blinds. Her complete face fills the frame, with soft skin texture, clear eyes, and a calm natural expression; the background is a warm blurred room with plants and books, not futuristic. She starts looking down in partial shade. As wind moves the curtains, horizontal bands of golden sunlight travel slowly across her cheek, lips, and eyes. She raises her chin and turns from left profile into a clear three-quarter view, then gives a subtle thoughtful smile. The camera makes a slow push-in at eye level. Elegant natural portrait lighting, warm highlights, visible head motion, clean subject-background separation, no hands near face, no text, no logos. Audio: curtain fabric moving, distant city ambience, soft breath.}
\noalign{\vskip 2.5pt}
\suppqualrow{ltx_portrait}{dense}{Dense}
\suppqualrow{ltx_portrait}{sol}{Sol-Attn}
\suppqualprompt{Prompt}{A bright 65mm cockpit close-up frames a young seaplane pilot flying above a turquoise tropical archipelago in late morning. Sunlight fills the cream-colored cabin and clearly lights her face, amber aviator glasses, linen shirt, and weathered leather headset. She glances from the instruments to the horizon, smiles slightly, and banks the aircraft with one smooth movement. Through the softly blurred side windows, white beaches, coral lagoons, and small green islands drift below in vivid blue and jade layers; a few highlights move across brushed aluminum controls in the foreground. The camera remains close at passenger-seat height, with strong subject separation, natural reflections, shallow depth, bright airy color, and a relaxed adventure-film atmosphere. Audio: steady propeller, gentle cabin vibration, headset static, and wind around the floats.}
\noalign{\vskip 2.5pt}
\suppqualrow{ltx_seaplane}{dense}{Dense}
\suppqualrow{ltx_seaplane}{sol}{Sol-Attn}
\suppqualmodelcaption{(c) LTX 2.3}
\end{tabular}

\captionsetup{skip=1pt,font=small}

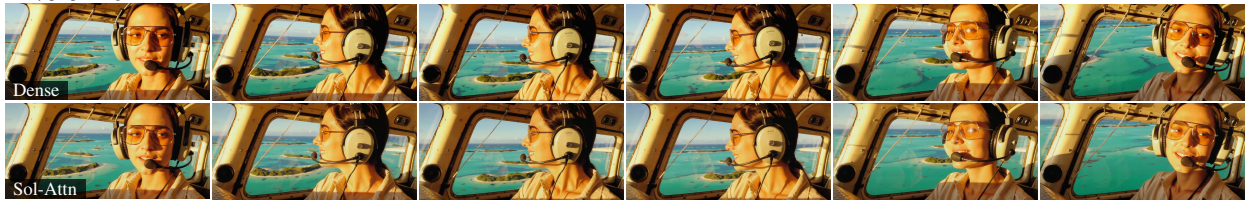
\captionof{figure}{Additional text-to-video comparisons of dense attention and \ourmethod{} across different models.}
\label{fig:supp_t2v_qualitative}
\par}

\clearpage
{\centering
\setlength{\tabcolsep}{0pt}
\renewcommand{\arraystretch}{1}
\begin{tabular}{@{}c@{\hspace{1.2pt}}c@{\hspace{1.2pt}}c@{\hspace{1.2pt}}c@{\hspace{1.2pt}}c@{\hspace{1.2pt}}c@{}}
\suppqualprompt{Edit prompt}{Turn the porcelain fox into transparent hand-blown glass containing a miniature thunderstorm with tiny clouds and lightning, while preserving its exact motion, silhouette, gallery, plinth, lighting, and camera path.}
\noalign{\vskip 2.5pt}
\suppqualrow{bernini_fox}{source}{Source}
\suppqualrow{bernini_fox}{dense}{Dense}
\suppqualrow{bernini_fox}{sol}{Sol-Attn}
\suppqualprompt{Edit prompt}{Turn the red raincoat into a living garment of emerald moss and small white flowers, and make the reflecting pool show a star-filled night sky, while preserving the woman, walk cycle, architecture, framing, and daylight.}
\noalign{\vskip 2.5pt}
\suppqualrow{bernini_coat}{source}{Source}
\suppqualrow{bernini_coat}{dense}{Dense}
\suppqualrow{bernini_coat}{sol}{Sol-Attn}
\suppqualmodelcaption{(a) Bernini}
\suppqualprompt{Prompt}{A first-person view from a strictly stationary observation point across an immense dry lakebed bordered by low mountain ranges. A black sports car occupies the central foreground on the pale, compacted surface, aligned toward the open horizon beneath a vast blue sky. The environment is broad and minimal, with flat beige desert crust, faint tire-worn texture, distant rocky ridgelines, and a few thin clouds stretching across the upper sky. Bright midday sunlight creates crisp shadows under the vehicle and a clean, high-visibility atmosphere of speed, openness, and isolation. The observer's perspective remains fixed, with no dynamic camera movement and no actions taken by the person recording. Autonomous motion belongs to the world itself: dust trails sweep low across the ground, heat haze shimmers near the horizon, clouds drift slowly, and the car's tires kick up fine desert grit.}
\noalign{\vskip 2.5pt}
\suppqualrow{sana_racer}{source}{Low-res.}
\suppqualrow{sana_racer}{dense}{Dense}
\suppqualrow{sana_racer}{sol}{Sol-Attn}
\suppqualprompt{Prompt}{A first-person view of a vast cobblestone plaza centered on an ornate equestrian statue atop a carved stone pedestal, flanked by symmetrical classical buildings with arched ground floors and uniform windows under slate roofs. The sky is pale blue with scattered clouds, casting soft daylight across the scene. Pigeons rest on the pavement in the foreground, while two pedestrians stroll near the statue's iron fence. Behind the monument, a grand neoclassical building with columns and sculptural reliefs anchors the background. Trees line the sides of the square, their green foliage contrasting with the beige stone architecture. The atmosphere is calm and open, with no visible motion beyond the stillness of the statues and the quiet presence of distant figures.}
\noalign{\vskip 2.5pt}
\suppqualrow{sana_plaza}{source}{Low-res.}
\suppqualrow{sana_plaza}{dense}{Dense}
\suppqualrow{sana_plaza}{sol}{Sol-Attn}
\suppqualmodelcaption{(b) SANA-WM}
\end{tabular}

\captionsetup{skip=2pt,font=small}

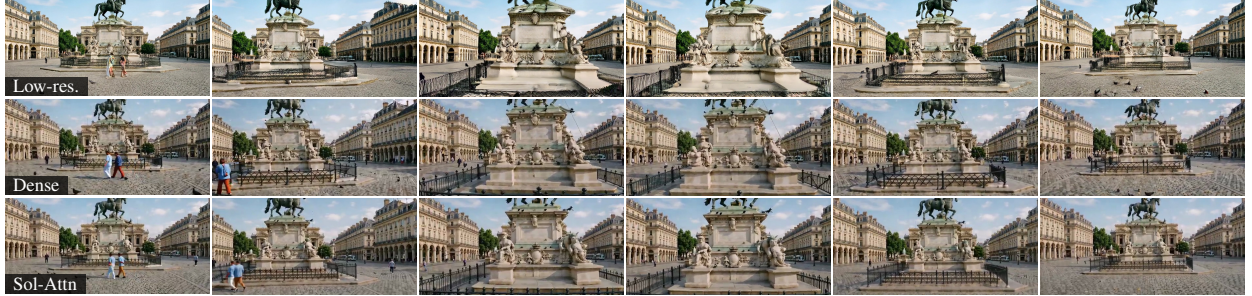
\captionof{figure}{Additional conditional-generation comparisons on Bernini and SANA-WM, with two cases per subfigure. Bernini rows show source, dense, and Sol-Attn-edited videos; SANA-WM rows show the low-resolution stage-1 video and both refined outputs.}
\label{fig:supp_conditional_qualitative}
\par}

\clearpage